\documentclass[a4paper,fleqn]{cas-sc}
\usepackage[authoryear,longnamesfirst]{natbib}

\usepackage{makecell}
\usepackage{multirow}
\usepackage{color}
\usepackage{tablefootnote}
\usepackage{tikz}
\usepackage{pgfplots}
\usepackage[edges]{forest}
\definecolor{paired-light-blue}{RGB}{198, 219, 239}
\definecolor{paired-dark-blue}{RGB}{49, 130, 188}
\definecolor{paired-light-orange}{RGB}{251, 208, 162}
\definecolor{paired-dark-orange}{RGB}{230, 85, 12}
\definecolor{paired-light-green}{RGB}{199, 233, 193}
\definecolor{paired-dark-green}{RGB}{49, 163, 83}
\tikzset{%
    parent/.style =          {align=center,text width=2.5cm,rounded corners=3pt, line width=0.3mm, fill=gray!10,draw=gray!80,text=black},
    for/.style =    {align=center,text width=2.2cm,rounded corners=3pt, fill=paired-light-blue!50,draw=paired-dark-blue!65,line width=0.3mm,text=black},   
    for_work/.style =           {align=center, text width=4.5cm,rounded corners=3pt, fill=paired-light-blue!50,draw=blue!0,line width=0.3mm,text=black},  
    of/.style =           {align=center,text width=2.2cm,rounded corners=3pt, fill=paired-light-orange!50,draw=paired-dark-orange!65,line width=0.3mm,text=black},   
    of_work/.style =           {align=center,text width=4.5cm,rounded corners=3pt, fill=paired-light-orange!50,draw=red!0,line width=0.3mm,text=black},    
    resource/.style =           {align=center,text width=2.2cm,rounded corners=3pt, fill= paired-light-green!50,draw=paired-dark-green!75,line width=0.3mm,text=black},   
    resource_work/.style =           {align=center,text width=4.5cm,rounded corners=3pt, fill= paired-light-green!50,draw= cyan!0,line width=0.3mm,text=black},      
}

\def\tsc#1{\csdef{#1}{\textsc{\lowercase{#1}}\xspace}}
\tsc{WGM}
\tsc{QE}
\tsc{EP}
\tsc{PMS}
\tsc{BEC}
\tsc{DE}

\begin{document}
\let\WriteBookmarks\relax
\def\floatpagepagefraction{1}
\def\textpagefraction{.001}

\shorttitle{}

\shortauthors{Xu et~al.}

\title [mode = title]{AI for Social Science and Social Science of AI: A Survey}


\author[1, 2]{Ruoxi Xu}[type=author, orcid=0000-0001-8145-6453]
\ead{ruoxi2021@iscas.ac.cn}
\author[1]{Yingfei Sun}[type=author]
\ead{yfsun@ucas.ac.cn}
\author[2]{Mengjie Ren}[type=author]
\ead{renmengjie2021@iscas.ac.cn}
\author[2]{Shiguang Guo}[type=author]
\ead{guoshiguang2021@iscas.ac.cn}
\author[2]{Ruotong Pan}[type=author]
\ead{panruotong2021@iscas.ac.cn}
\author[2]{Hongyu Lin}[type=author]
\cormark[1]
\ead{hongyu@iscas.ac.cn}
\author[2, 3]{Le Sun}[type=author]
\ead{sunle@iscas.ac.cn}
\author[2, 3]{Xianpei Han}[type=author]
\ead{xianpei@iscas.ac.cn}

\affiliation[1]{organization={School of Electronic, Electrical and Communication Engineering, University of Chinese Academy of Sciences},
    city={Beijing},
    country={China}}
\affiliation[2]{organization={Chinese Information Processing Laboratory, Institute of Software, Chinese Academy of Sciences},
    city={Beijing},
    country={China}}
\affiliation[3]{organization={State Key Laboratory of Computer Science, Institute of Software, Chinese Academy of Sciences},
    city={Beijing},
    country={China}}

\cortext[cor1]{Corresponding author}

\begin{abstract}
Recent advancements in artificial intelligence, particularly with the emergence of large language models (LLMs), have sparked a rethinking of artificial general intelligence possibilities.
The increasing human-like capabilities of AI are also attracting attention in social science research, leading to various studies exploring the combination of these two fields. 
In this survey, we systematically categorize previous explorations in the combination of AI and social science into two directions that share common technical approaches but differ in their research objectives. The first direction is focused on \emph{AI for social science}, where AI is utilized as a powerful tool to enhance various stages of social science research. While the second direction is the \emph{social science of AI}, which examines AI agents as social entities with their human-like cognitive and linguistic capabilities.
By conducting a thorough review, particularly on the substantial progress facilitated by recent advancements in large language models, this paper introduces a fresh perspective to reassess the relationship between AI and social science, provides a cohesive framework that allows researchers to understand the distinctions and connections between AI for social science and social science of AI, and also summarized state-of-art experiment simulation platforms to facilitate research in these two directions.
We believe that as AI technology continues to advance and intelligent agents find increasing applications in our daily lives, the significance of the combination of AI and social science will become even more prominent. 
\end{abstract}

\begin{keywords}
Social Science \sep
Large Language Models \sep
AI Simulation \sep
\end{keywords}

\maketitle

\section{Introduction}

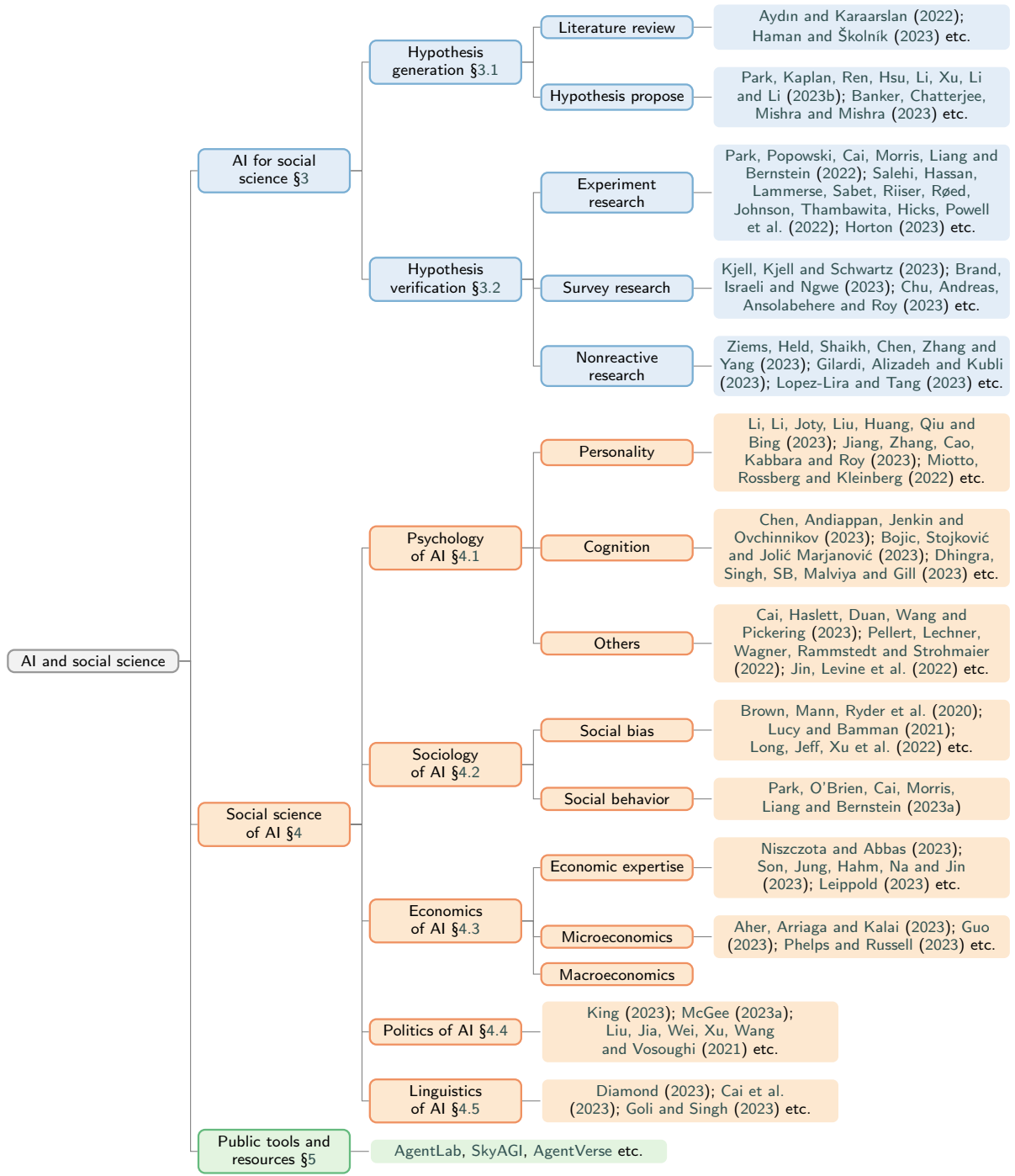
\begin{figure*}[!ht]
\scriptsize
    \begin{forest}
        for tree={
            forked edges,
            grow'=0,
            draw,
            rounded corners,
            node options={align=center,},
            text width=2.7cm,
            s sep=6pt,
            calign=edge midpoint,
        },
        [AI and social science, fill=gray!45, parent
            [AI for social science \S\ref{ai_for_social_science}, for tree={for}
                [Hypothesis generation \S\ref{hypothesis_generation},  for
                    [Literature review, for
                        [\citet{aydin2022openai, haman2023using} etc., for_work]
                    ]
                    [Hypothesis propose, for
                        [\citet{park2023can, banker2023machine} etc., for_work]
                    ]
                ]
                [Hypothesis verification \S\ref{hypothesis_verification},  for
                    [Experiment research,  for
                        [\citet{park2022social, salehi2022synthesizing, horton2023LargeLanguageModels} etc., for_work]
                    ]
                    [Survey research,  for
                        [\citet{kjell2023ai, brand2023UsingGPTMarketa, chu2023language} etc., for_work]
                    ]
                    [Nonreactive research, for
                        [\citet{ziems2023can, gilardi2023chatgpt, Lopez_Lira_2023} etc., for_work]
                    ]
                ]
            ]
            [Social science of AI \S\ref{social_science_of_ai}, for tree={of}
                [Psychology of AI \S\ref{psychology}, of
                    [Personality,  of
                        [\citet{li2023DoesGPT3Demonstrate, jiang2023PersonaLLMInvestigatingAbility, miotto2022WhoGPT3Exploration} etc., of_work]
                    ]
                    [Cognition,  of
                        [\citet{chen2023ManagerAIWalk, bojic2023SignsConsciousnessAi, dhingra2023MindMeetsMachine} etc., of_work]
                    ]
                    [Others,  of
                        [\citet{cai2023DoesChatGPTResemble, pellert2022AIPsychometricsUsing, jin2022WhenMakeExceptions} etc., of_work]
                    ]
                ]
                [Sociology of AI \S\ref{sociology}, of
                    [Social bias
                        [\citet{brown2020GPT3LanguageModels, lucy2021GenderRepresentationBias, ouyang2022InstructGPTTrainingLanguage} etc., of_work]
                    ]
                    [Social behavior
                        [\citet{park2023GenerativeAgentsInteractivea}, of_work]
                    ]
                ]
                [Economics of AI \S\ref{economics}, of
                    [Economic expertise, of
                        [\citet{niszczota2023GPTFinancialAdvisor, son2023ClassificationFinancialReasoning, leippold2023ThusSpokeGPT3} etc., of_work]
                    ]
                    [Microeconomics, of
                        [\citet{aher2023UsingLargeLanguage, guo2023GPTAgentsGame, phelps2023InvestigatingEmergentGoalLike} etc., of_work]
                    ]
                    [Macroeconomics, of
                    ]
                ]
                [Politics of AI \S\ref{politics}, of
                    [\citet{king2023GPT4AlignsNew, mcgee2023ChatGptBiased, liu2021MitigatingPoliticalBias} etc., of_work]
                ]
                [Linguistics of AI \S\ref{linguistics}, of
                    [\citet{diamond2023GenlangsZipfLaw, cai2023DoesChatGPTResemble, goli2023LanguageTimePreferences} etc., of_work]
                ]
            ]
            [Public tools and resources \S\ref{resource}, for tree={resource}
                [{\href{https://github.com/renmengjie7/AISimuToolKit}{AgentLab}, \href{https://github.com/litanlitudan/skyagi}{SkyAGI}, \href{https://github.com/OpenBMB/AgentVerse}{AgentVerse} etc.}, resource_work]
            ]
        ]
    \end{forest}
    \caption{Overview of the intersection of AI and social science. We have separately discussed "AI for social science" which summarizes the application of AI at every stage of social science research to provide guidance on tool selection for researchers, "social science of AI" which systematically describes the intelligence level and characteristics of AI agents from a social science perspective on different sub-disciplines, and "public tools and resources" which focus on simulation tools. These fields share technical methodologies to some extent, yet they possess distinct research subjects and objectives.}
    \label{fig:overview}
\end{figure*}
Building machines that can think, learn and create is the fundamental pursuit of artificial intelligence (AI)~\citep{russell2010artificial}. How to develop machines with general intelligence comparable to, or even greater than, that of human beings has never lost its appeal~\citep{goertzel2014artificial}. Recently, significant advancements have been made in the AI field~\citep{zhao2023survey}, particularly with the emergence of large language models (LLMs) such as ChatGPT and GPT-4~\citep{openai2023GPT4TechnicalReport}. These developments have led to the rethinking of the possibilities of artificial general intelligence (AGI)~\citep{zhao2023survey}.

The increasing human-like capabilities of AI are also attracting attention in social science research. There have been numerous studies exploring the combination of AI and social science~\citep{bail2023can, ziems2023can, changfen2023gen}. Along these lines, many novel research directions have been explored, including research tasks proposing~\citep{park2023can, banker2023machine}, social science simulation~\citep{brand2023UsingGPTMarketa, kjell2023ai, chu2023language}, AI agent governing~\citep{li2023DoesGPT3Demonstrate, jiang2023PersonaLLMInvestigatingAbility, miotto2022WhoGPT3Exploration} and so on.

\begin{figure*}[t!]
\centering
    \includegraphics[width=0.7\textwidth]{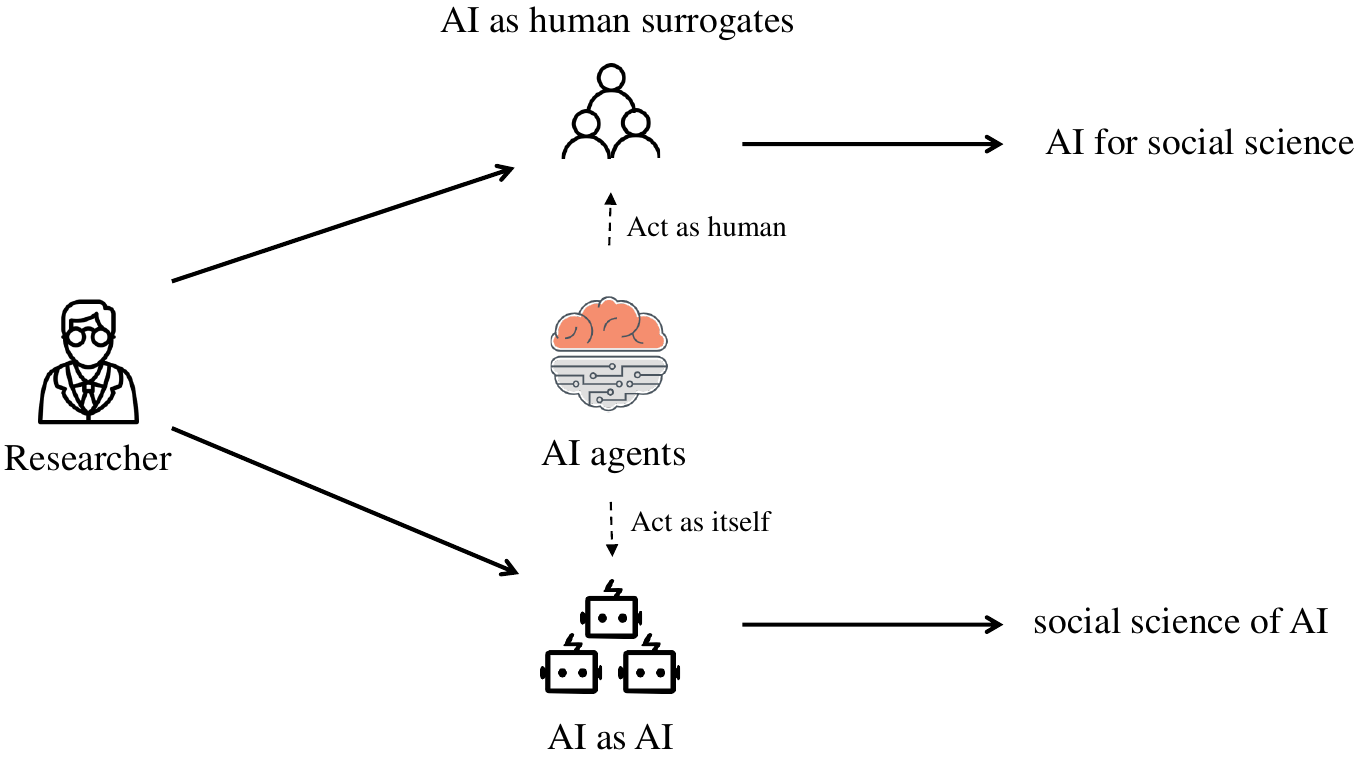}
    \caption{Computer simulation respectively in the context of "AI for social science" and "social science of AI". For “AI for social science”, AI agents are deployed to mimic human behaviors to enhance the understanding of human society.  Conversely, "social science of AI" delves into AI agents' own social questions.}
    \label{fig:simulation_example}
\end{figure*}

Despite the large number of related studies, the existing studies tend to concentrate on one specific instance of AI and social science intersection, thereby lacking a unified perspective to effectively distinguish and outline AI's role in social science research and its own social characteristics. In reality, the combination of AI and social science can be divided into two distinct directions.
On the one hand, the superior performance of AI allows them to serve as effective tools for social science research, such as using AI for literature searching and reviewing~\citep{mcgee2023using}, proposing questions and hypotheses~\citep{park2023can, banker2023machine}, analyzing data~\citep{ziems2023can}, assisting with writing~\citep{dergaa2023human, chen2023chatgpt}, and more. Systematically outlining the potential applications of AI in different phases of social science research can provide a valuable guide for researchers to choose appropriate research tools.
We refer to this direction as \emph{AI for social science} in this paper.
On the other hand, just as early myths and parables emphasized the social and ethical questions around human-created intelligence~\citep{mccorduck2004machines, kieval1997pursuing, pollin1965philosophical}, today’s intelligent machines present their own interesting social questions~\citep{frank2019evolution} and expanding research starts to explore and understand AI agents as social entities. Particularly, current AI agents, especially large language model agents, are exhibiting cognitive, logical reasoning, and linguistic capabilities on par with or even surpassing those of humans, along with unique behavioral characteristics~\citep{openai2023GPT4TechnicalReport}. Communities constituted by AI agents also exhibit emergent behaviors similar to human societies~\citep{park2023GenerativeAgentsInteractivea}.
This provides an interesting case for attempting to extend the social science to more universal phenomena of machines~\citep{klein2002social} and also presents a valuable opportunity to reevaluate a fundamental axiom in social science: human behavior can be understood as possessing unique social characteristics~\citep{woolgar1985not}. Exploring AI from a social science perspective can also provide crucial insights and guidance to make AI development more congruent with societal needs and human values. 
We refer to this direction as \emph{social science of AI} in this paper.
There is an important point to note, the term "social science" as used in this paper extend its traditional definition. It is used in a broader sense to provide a research perspective for describing certain high-level behaviors of humans or models, rather than equating it with actual human social behaviors.

Although these directions share common technological approaches, they have distinct research objectives, significance, and scopes of application.
For example in Figure~\ref{fig:simulation_example}, using AI agents for simulation serves as a technical method that could apply to both directions, but with differing objectives. When used for the former, the researcher's aim is to align the behavior of AI agents as closely with human behavior as possible, in order to study the operational laws of human society in a cost-effective, fast, and ethically risk-averse manner~\citep{park2022social, salehi2022synthesizing}. When used for the latter, the objective is to explore the behavioral laws of AI itself, with a particular focus on its unique aspects, especially those differing from the operational laws of human society~\citep{guo2023GPTAgentsGame}.
The absence of surveys from the above two perspectives makes it hard to ground each work's research significance and application scope, hindering us to comprehend and harness the distinctions and connections between these two directions. A joint analysis of their research progress can help to grasp the big picture of the current state of the combination of AI and social science.

To this end, we conduct a comprehensive review from these two directions respectively. We conduct a joint analysis of their research progress, comparing their similarities and differences to present an overview of the current state of the combination of AI and social science.
Considering that recent remarkable advancements in this field can be largely attributed to the development of large language models~\citep{zhao2023survey}, this paper narrows its scope to the combination of large language models and social sciences, approaching the topic from both the angle of AI for social science and the social science of AI. The main organization of this survey is summarized in Figure~\ref{fig:overview}.
Specifically, from the angle of AI for social science, we discuss large language models' potential as a highly efficient tool that can be integrated into existing research methodologies, significantly enhancing the efficiency of social science research. To achieve this, we structure the content according to the roles that AI plays in both the stages of hypothesis generation and verification within the social science process~\citep{donovan2013elements, bryman2016social}.
For hypothesis generation, we mainly focus on how AI can help human beings in literature reviewing and hypothesis proposing. For hypothesis verification stage, we respectively examine how large language models function in various research methods such as experiment research, survey research and non-reactive research.
From the perspective of the social science of AI, we are referring to a broad field of social science that focuses on regarding large language models as its research subject. We categorize the behavioral studies of these models according to each subfield within social science, following academic categorization. More specifically, we have compiled the behavioral laws of large language models by examining them from the viewpoints of psychology, sociology, economics, politics, and linguistics.
Additionally, we have also compiled a summary of the currently available tools in this field to facilitate research in the aforementioned areas. These platforms utilize large language models as agents and allows for the setting and implementation of intervention conditions to simulate diverse social situations, interactions, and behaviors. They serve the purpose of simulating human behavior for studying human societies, as well as exploring AI societies, thus catering to both of the above directions.

Generally, we summarize our contributions as follows:
\begin{itemize}
    \item We present a perspective of revisiting AI and social science combinations from two directions: \emph{AI for social science} and \emph{social science of AI}. We elaborate on the connections and distinctions between these two directions, grounding the research value and application scope of relevant work.
    \item Based on the substantial progress facilitated by recent advancements in large language models to these two directions, we conduct a literature review, which summarizes the research landscape, discusses the limitations of the existing research, and sheds light on potential future directions in the combination of AI and social science.
    \item We collect and compare existing open source large language model-based simulation tools. These platforms can serve an effective foundation to facilitate the future researches of the above-mentioned two directions.
\end{itemize}

The structure of this survey is organized as follows: Section~\ref{ai_for_social_science} introduces the application of large language models in social science research, while Section~\ref{social_science_of_ai} delves into social science research that takes large language models as the subject of study. Section~\ref{resource} provides information about the resources and tools available. Finally, we conclude the survey in Section~\ref{conclusion}, summarizing the main findings and discussing the remaining issues for future work.
\section{Background}

Nowadays, the AI community, and even the whole society, is witnessing the significant impacts brought about by large language models.
Large language models typically refer to transformer language models that contain hundreds of billions (or more) of parameters~\citep{shanahan2022talking}, which are trained on massive text data.
Notable examples include GPT-3~\citep{brown2020GPT3LanguageModels}, PaLM~\citep{chowdhery2022PaLMScalingLanguage}, LLaMA~\citep{touvron2023LLaMAOpenEfficient} and OpenAI's ChatGPT, which amassed 100 million users in less than two months, setting a new record in history.
These models have exhibited strong capacities to understand natural language and solve complex tasks (via text generation), capturing the attention and imagination of investors, consumers, and organizations.

Improvements in large language models have been so fast, and the potential societal repercussions so profound, that a broad cross-disciplinary lens from computer science and social science is necessary to start making sense of the implications. 
Firstly, the ability of large language models to generate texts for a broad range of tasks via an intuitive natural language interface may hold great promise for social science research. This has opened up avenues for various applications, including literature searching and reviewing~\citep{mcgee2023using}, proposing questions and hypotheses~\citep{park2023can, banker2023machine}, analyzing data~\citep{ziems2023can}, assisting with writing~\citep{dergaa2023human, chen2023chatgpt}, and more. 
Secondly, the evolution of large language models has significantly enhanced their capacity to exhibit human-like characteristics, leading to a surge in research regarding LMs as representations of human entities~\citep{krishna2022socially, andreas2022language, park2022social}. Research includes exploring the collaborative capabilities of large language models in complex tasks~\citep{irving2018ai}, developing "generative agents" to investigate emergent social behaviors~\citep{park2023GenerativeAgentsInteractivea}, employing GPT-3-based agents as substitutes for human participants~\citep{aher2023UsingLargeLanguage} and so on. 
Thirdly, the advancements in large language models have prompted a reconsideration of the ethical and societal issues it may entail.
These previous works have highlighted the transformative effects brought about by the emergence of large language models in the intersection of AI and social science. Our survey aims to provide a comprehensive overview of these developments and address the existing limitations and potential of this field.
\begin{figure*}[t!]
\centering
    \includegraphics[width=0.8\textwidth]{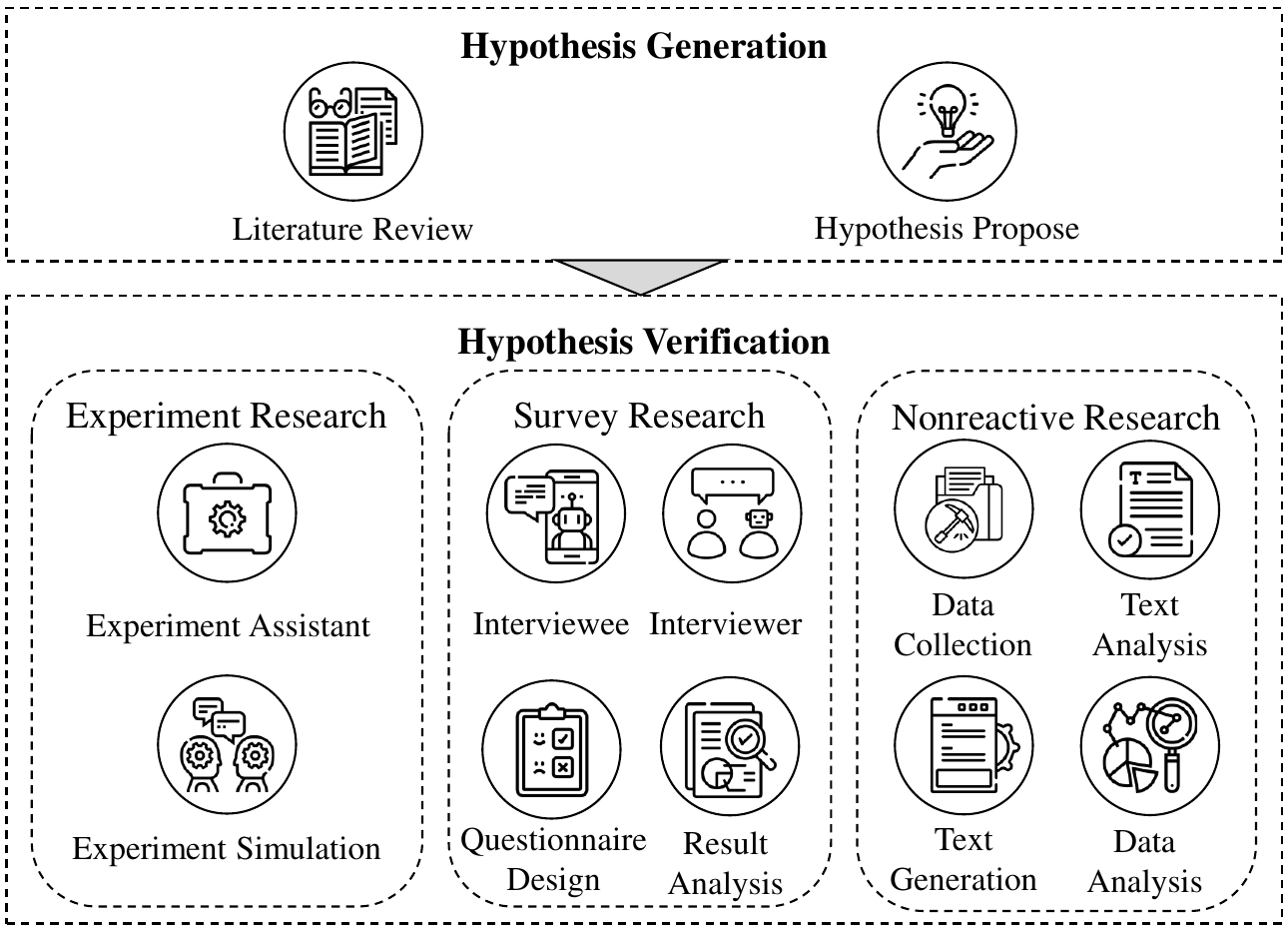}
    \caption{The application of large language models at every stage of social science research. Large language models offer new possibilities for improving existing social science research processes and automated science, but also bring new potential risks and ethical issues. Social science researchers should carefully consider whether and how to apply large language models in their research.}
    \label{fig:ai_for_social_science}
\end{figure*}

\section{AI for social science} \label{ai_for_social_science}

AI for social science refers to the application of AI in traditional social science research. Unlike the social science of AI, this section emphasizes large language models' human-like intelligence, which can mimic human behavior to help social science research.
In this section, we will draw upon the research paradigm outlined in~\citep{donovan2013elements, bryman2016social}, and discuss the application of large language models as multi-purpose tools at every stage of social science research as shown in Figure~\ref{fig:ai_for_social_science}. This aims to provide a comprehensive and informed perspective for social science researchers on how to apply large language models in the process of their research to enhance efficiency, while also revealing the untapped potential of large language models, warning about potential risks and ethical issues, and indicating possible future directions in their application.

\subsection{Hypothesis Generation}
\label{hypothesis_generation}

Hypothesis generation, which serves as the foundation and initial step of social science research, is the task of mining meaningful implicit associations between unrelated social science concepts~\citep{jha2019hypothesis}. In the early days, hypothesis generation was largely driven by brainstorming of researchers, drawing inspiration from existing theories, patterns of anomalies in data or cross-disciplinary connections that involved serendipitous discoveries~\citep{jaccard2019theory}. However, as the volume of research literature grows, researchers have started to explore quicker and more efficient methods of generating solid hypotheses~\citep{evans2010machine, krenn2020predicting, wilson2018automated}. In the following, we will present some attempts to accomplish hypothesis generation tasks using large language models.

\paragraph{Literature Review} is the understanding, summarization, and critical thinking about the academic literature on a specific topic. Researchers have been exploring the use of large language models to assist in literature review.
Some studies leverage large language models to aid in searching for relevant literature.~\citet{wang2023can} used ChatGPT to formulate and refine Boolean queries for systematic reviews, finding that ChatGPT compares favourably with the current state-of-the-art automated Boolean query generation methods in terms of precision, at the expenses of a lower recall. The paper also found that guided prompts lead to higher effectiveness than single prompt strategies.
Other works focus on enabling large language models to read articles and automatically summarize the key points within them.~\citet{aydin2022openai} used ChatGPT to paraphrase the abstracts of relevant papers and answer questions to automatically generate literature review, which has been cited over 100 times.
Elicit\footnote{https://elicit.org/} is also one typical application, an AI Research Assistant based on large language models, which can find relevant papers without perfect keyword match, summarize takeaways from the paper specific to your question, and extract key information from the papers.
However, it should be noted that directly relying on large language models for literature recommendation and summarization is currently infeasible due to the risk of unreliable papers and related information being generated~\citep{haman2023using}.

\paragraph{Hypothesis Propose} is to propose possible explanations to some phenomenon or event.
Researchers from various fields have made attempts to use large language models to help with this task.
For example,~\citet{park2023can} used GPT-4 to generate scientific hypotheses and draw the conclusion that current large language models seem to be able to effectively structure vast amounts of scientific knowledge and provide interesting and testable hypotheses while the error rate is high.
~\citet{banker2023machine} utilized a fine-tuned version of GPT-3 to generate psychological hypotheses and engaged 50 psychology experts to evaluate their quality, revealing that the model's generated hypotheses were not mere replicas of previously generated human hypotheses, and exhibited no significant differences in terms of clarity, impact, and originality compared to human-generated hypotheses.
~\citet{tang2023less} employed large language models to generate "less likely" hypotheses, effectively assisting humans in comprehensively examining problems and eliminating cognitive biases caused by their own knowledge and experience.
At present, the application of large language models in hypothesis generation is still somewhat rudimentary. Strategies for enhancing hypothesis quality include fine-tuning the large language models within specific domains~\citep{banker2023machine}, stepwise questioning~\citep{wei2022chain}, adversarial dialogue~\citep{park2023can} and so on.

\paragraph{Conclusion}

Current research on using large language models for hypothesis generation focuses on exploring its feasibility and validity, and has commonly unveiled promising prospects and potential of employing large language models for this task. The solution mainly involves interactive design of prompts. Users input the topic they want to review. Based on these inputs, large language model automates the process of formulating and refining boolean queries, extracting core points from the search results, and generating hypotheses about potential relationships among the objects of interest.

Compared to traditional methods, the \textbf{advantages} mainly lie in their exceptional performance in language understanding and generation, enabling them quickly analyze existing research, identify knowledge gaps, and establish connections between seemingly unrelated ideas~\citep{dahmen2023artificial}. This provides them with a natural advantage in language-based disciplines such as psychology~\citep{banker2023machine}.

However, researchers need to be noted that currently, large language models can only be used as auxiliary and inspirational tools in the early stage of research, and have the following \textbf{limitations}: 1) Fabricated or incorrect information, which may mislead users. This is because large language models lack of understanding regarding the validity of output content and simply spill out them without clear rationalization~\citep{park2023can}. 2) High sensitivity to the prompt, which results in a significant investment of effort in prompt design but yields uncertain outcomes~\citep{wang2023can}. 3)  Limited context length, which makes it hard to handle long and multiple documents.

In order to better harness the potential of large language models in hypothesis generation, \textbf{future directions} that we may consider include: 1) Integrating specialized domain knowledge, by retrieval augmented techniques or domain-specific training. This would help reduce hallucination. 2) Developing high-reward prompt strategies. This could involve considering novel prompt generation techniques or reward mechanisms to guide the model's hypothesis generation process. 3) Expanding the context windows of the large language models. By allowing the models to consider a larger context, they would have access to more comprehensive information, potentially leading to more robust and insightful hypotheses.

\begin{table*}[tp]
    \centering
    \begin{tabular}{lcc}
    \toprule
    \textbf{Research Stages} & \textbf{Traditional Methods} & \textbf{Large language models} \\ \hline
    \textbf{Hypothesis Generation} & & \\
    \quad Speed & Low & \textbf{High}   \\
    \quad Validity & \textbf{High}  & Low \\
    \quad Novelty & Low & \textbf{High}   \\
    \hline
    \textbf{Hypothesis Verification} & & \\
    Experiment Research & & \\
    \quad Cost & High & \textbf{Low}  \\
    \quad Speed & Low & \textbf{High}   \\
    \quad Reproducibility & Low & \textbf{High}   \\
    \quad Scalability & Low & \textbf{High}   \\
    \quad Fidelity & \textbf{Entire}  & Not Sure \\
    Survey Research & & \\
    \quad Cost & High & \textbf{Low}  \\
    \quad Engagement & Low & \textbf{Entire}   \\
    \quad Interaction & Fixed & \textbf{Natural}  \\
    \quad Bias & \textbf{Low}  & Not Sure \\
    Nonreactive Research & & \\
    \quad Generality & Single-purpose & \textbf{Multiple-purpose}  \\ 
    \quad Accessibility &  Low & \textbf{High}   \\
    \quad Numerical analysis & \textbf{Accurate}  & Not Sure \\
    \bottomrule
    \end{tabular}
    \caption{The comparisons between traditional methods and large language models as tools at every stage of social science research. The advantages are marked in bold.
    From the table, we can easily find that although large language models are more advantageous on cost, speed, generality and accessibility across various research stages, the critical current limitations of validity, possible ethical risks and lack of domain knowledge still hinder its real-word applications. Please note that the comparisons in this study are only valid until the publication of this paper. Given the rapid advancements in technology, LLM agents may overcome some of the current shortcomings that are challenging to address.}
    \label{tab:comparsion_between_tools_and_llms}
\end{table*}

\subsection{Hypothesis Verification}
\label{hypothesis_verification}

Once the research topic and hypotheses are established, social science researchers engage in hypothesis verification. This process involves collecting and analyzing data to provide evidence that either supports or refutes the proposed hypotheses~\citep{donovan2013elements}. In traditional social science research, hypothesis verification typically falls into quantitative methods like experimental research, survey research and nonreactive research, as well as qualitative methods such as field research and historical-comparative research~\citep{juren2017, fang2013, bryman2016social}. Given that large language models are currently limited in their applicability to qualitative research, we primarily discuss the role of large language models in quantitative methods.

\subsubsection{Experiment Research}

A laboratory experiment is "an inquiry for which the investigator plans, builds,or otherwise controls the conditions under which phenomena are observed and measured"~\citep{willer2007building}. The common practice involves manipulating conditions for some research participants while leaving them unaltered for others, aiming to compare the responses across groups to uncover consistent behavioral patterns. In the following, we will explore the applications of large language models in experimental research.

\paragraph{Experiment Assistant} refers to the use of large language models in social science experiments to automate some simple but labor-intensive tasks that would normally be done by researchers.
For instance, they can assist in creating hypothetical scenarios iteratively with feedback from researchers~\citep{bail2023can}, which can enhance the external validity and comparability of the experimental conditions.
Besides, large language models are capable of synthesizing the necessary information for experiments, eliminating the need for the use of real-life information utilization. This safeguards the privacy of individuals whose information could potentially be used in these studies.

\paragraph{Experiment Simulation}
\label{simulation}
aims to design a platform to explore, optimize, and predict behaviors of complex systems that might be challenging to investigate in the real world.
In simulation experiments, large language models are usually used as believable proxies of human behavior~\citep{aher2023UsingLargeLanguage, park2023GenerativeAgentsInteractivea}.
For example,~\citet{park2022social} provides a typical application where large language models are utilized to simulate the behavior of community users, assisting designers in gaining insights into the various possibilities of social interactions and identifying potential edge cases that could lead to the breakdown of a community.
~\citet{horton2023LargeLanguageModels} considers GPT-3 AIs as homo silicus agents, and demonstrates their ability to qualitatively recover findings from three classic behavioral economics experiments with real humans.
~\citet{guo2023GPTAgentsGame} designs well-crafted prompts to enable GPT agents to participate in strategic game experiments and achieve realistic performance.
~\citet{park2023GenerativeAgentsInteractivea} constructed a fully large language model-driven simulated community, where they observed human-like individual behaviors and emergent behaviors.

\paragraph{Conclusion}
In experimental research, large language models can serve dual roles - they can act as experiment assistants and as believable proxies of human behavior, becoming subjects of the experiment themselves. The latter, in particular, has attracted increasing attention in both AI and social science as large language models are increasingly capable of simulating human-like responses and behaviors. Currently, the design of AI agents is still relatively crude, usually including four modules: profile, memory, planning, and action~\citep{wang2023survey}, and warrants further improvement.

Using large language models for simulating experiments offers several \textbf{advantages}: 1) Improved efficiency, reduced costs, and enhanced scalability~\citep{bybee2023SurveyingGenerativeAI, guo2023GPTAgentsGame}. 2) Circumventing the ethical issues associated with human subjects. This opens the door to experiments that may be deemed unethical if performed on humans, such as the classic Stanford Prison Experiment~\citep{zimbardo1971stanford}.

However, social scientists must also proceed with caution in this area, taking into account the following \textbf{limitations}: 1) Uncertain believability. There is now no "gold standard" study demonstrating that groups of automated agents can accurately simulate humans~\citep{bail2023can}. 2) Low transparency and reproducibility. Since large language models themselves are still black boxes, we cannot provide a thorough explanation of their behavior.

In order to address the aforementioned limitations, potential \textbf{future directions} include: 1) developing methods for evaluating the quality of large language models simulations, and 2) incorporating insights from cognitive science to guide the development of AI agent frameworks and enhance their behavior's human-likeness and rationality.

\subsubsection{Survey Research}
Survey research is a fundamental methodology in social science, which uses written questionnaires or formal interviews to collect information on the beliefs, opinions, characteristics, and past or present behaviors of a target group~\citep{bryman2016social}. The core of modern survey research is three key components: sampling, measurement, and analysis~\citep{wright2010survey}. The following will explore the role that large language models play in each stage of survey research.

\paragraph{Sampling} involves selecting representative samples from human populations, whose observed characteristics provide unbiased estimates of the characteristics of those populations.
Large language models present a novel option for sampling, serving as proxies for specific human subgroups.
This enables the avoidance of the sampling step, or rather, utilizes the extensive training database of the large language models directly as the sample for the study.
Existing studies have proposed and explored the possibility of using language models as effective stand-ins for particular human demographics in the realm of social science research.
For example, ~\citet{argyle2022OutOneMany} compares real human participants from multiple large surveys in the United States and "silicon samples" which are created by conditioning large language models on socio-demographic backstories from them, demonstrating that the “algorithmic bias” within GPT-3 is both fine-grained and demographically correlated.

~\citet{bisbee2023artificially} investigates the quality, reliability, and reproducibility of synthetic survey data generated by the popular closed-source large language model, ChatGPT. The experimental results suggest that the average scores generated by ChatGPT closely align with the averages from baseline survey (conducted from 2016 to 2020 on the U.S. national elections). However, when it comes to more advanced features, such as variance, subgroups, and statistical inferences, it often leads researchers to draw conclusions that differ from those relying on human respondents.
~\citet{dillion2023can} takes moral judgments as an example to investigate whether and when language models can potentially replace human participants in psychology. The analysis indicate that language models align most closely with humans when the contextual circumstances involve explicit features that drive human judgment, do not pertain to competitive situations, and when the subjects being judged are representative within the training data.
~\citet{rao2023can} conducts the Myers-Briggs Type Indicator (MBTI) test on large language models agents of different subpopulations and showcases the ability of ChatGPT in analyzing personalities of different groups of people.
~\citet{brand2023UsingGPTMarketa} interview GPT-3 to estimate customers' willingness-to-pay for products and features and find that large language models can generate responses that align with economic theories and consumer behavior patterns.
~\citet{chu2023language} adapts LMs to subpopulation-specific media diets and successfully simulates how the subpopulation will respond to survey questions.

\paragraph{Measurement} focuses on designing questions to draw out valid and reliable responses across a broad spectrum of subjects, which are often characterized as "the art of asking questions".
While it's natural to leverage large language models to assist in the design of questionnaires or interview questions, more researchers are focusing on the role large language models play in facilitating a paradigm shift in the measurement methods within survey research - from closed-ended rating scales, to open-ended response questionnaires, and then further towards more natural interactive interviews.
For example,~\citet{kjell2022natural} compared the results of psychological surveys using rating scales and natural language-based open-ended questionnaires. The latter was found to achieve accuracy that either exceeded or rivaled the typical methods of reliability in rating scales, which is often considered as the theoretical upper limit.
~\citet{kjell2023ai} provides a future outlook of finer granularity and automated interactive interviews, making full use of interviewees' own words to best elicit truthful responses.

\paragraph{Analysis} is a step using multivariate data analysis techniques to identify and understand the statistical relationships among various variables.
Large language models can be used to analyze qualitative data, such as interview responses, to identify patterns, relationships, and common themes~\citep{abbas2023uses}. For instance,~\citet{yang2023evaluations} utilize large language models, specifically ChatGPT, to perform mental health analysis and highlight the significant potential of large language models in improving the interpretability of mental health analysis.
However, large language models, which are not specifically designed for analyzing quantitative data, are currently not the primary method for survey research analysis in the social sciences. Instead, survey data is typically presented in the form of charts, graphs, or tables, and analyzed using statistical methods~\citep{bryman2016social}. Future versions of the model may be able to integrate tools like Python and R libraries to conduct quantitative data analyses~\citep{mialon2023augmented}.

\paragraph{Conclusion}

The current applications of large language models in survey research primarily revolve around three main directions: 1) Effective proxies for specific human sub-populations. 2) Interactive interviewers. 3) Result analysis tools.

For the first direction, the current work has demonstrated that proper conditioning will cause large language models to accurately emulate response distributions from a wide variety of human subgroups. This approach can effectively address limitations regarding the number of questions, frequency and the subpopulation due to cost constraints~\citep{chu2023language}, as well as the common challenge of low response rates~\citep{anol2016social} in survey research, offering significant advantages in terms of engagement and cost. However, whether and which large language models can truly represent humans remains an open question~\citep{argyle2022OutOneMany}. This approach fundamentally relies on "algorithmic bias," which is heavily influenced by the training data and is susceptible to producing unfair and non-objective results. In light of these considerations, we do not propose that large language models should completely replace traditional sampling methods in survey research. Instead, we see their potential in simulating population responses to assist in survey design. This hybrid approach allows us to harness the strengths of large language models while still recognizing the importance of traditional sampling techniques to maintain the integrity and fairness of survey results.

Several researchers have envisioned the impact of language models on the form of survey. The advantage of large language models lies in their ability to fully utilize the individuals' own language to describe the information needed by researchers, which has the potential not only to gradually improve current assessments but also to fundamentally alter the nature of measuring and describing personal states, ultimately enhancing our understanding of social science. However, utilizing large language models to facilitate measurement also poses risks. Inherent biases in large language models and the potential for data leakage must be carefully navigated when implementing large language models in research scenarios.

For result analysis, large language models are primarily used for text analysis and are seldom employed for numerical analysis because they are not proficient in it. Future research can consider the integration of large language models with specialized computational tools.

\subsubsection{Nonreactive Research}
Nonreactive research refers to the research method where the participants are not aware that their information is part of the study, unlike experiment research and survey research that actively engage the people we study by creating experimental conditions or directly asking questions~\citep{juren2017}.
This method may reduce bias due to interference from researchers or measurement instruments~\citep{trochim2001research}.
In this section, we will adhere to the taxonomy in~\citep{juren2017}, and explore the roles that large language models play in content analysis and existing statistics analysis.

\paragraph{Content analysis} is a widely used technique for examining the content contained in written documents or other communication media. The remarkable performance of large language models in various traditional NLP tasks has attracted extensive attention about using them in content analysis tasks within the field of social science.

Some social science researchers employ large language models to perform text classification, a basic and important task that involves labeling or categorizing texts according to predefined categories~\citep{aggarwal2012survey}. Common text classification tasks in the field of social science include: 
1) \textbf{Sentiment analysis}, which aims to identify and extract the emotional attitudes in the text, such as joy, anger and sorrow. It is a widely applied technique in psychology and political science. In psychology, sentiment analysis helps researchers understand people’s emotional states, stress levels, and mental health conditions.~\citet{RODRIGUEZIBANEZ2023119862} suggest that large language models are the future paradigm for sentiment analysis, due to their zero-shot setting and simple invocation. However, they also point out the limitations of GPT-3 in the current tasks. ChatGPT performs excellently in three text-based mental health classification tasks, including stress detection, depression detection, and suicide detection~\citep{lamichhane2023evaluation}. ChatGPT also applies to differentiate paranoid texts from non-paranoid ones in some studies~\citep{uludag2023chatgpt}.~\citet{rathje2023gpt} evaluate the performance of GPT-3.5 and GPT-4 on multilingual sentiment and discrete emotions tasks and find that in many cases, GPT models perform close to (sometimes better than) fine-tuned machine learning models. They argue that GPT models offer a promising avenue for cross-lingual research in psychology.~\citet{wuBloombergGPTLargeLanguage2023} introduce BloombergGPT, a 50 billion parameter language model tailored for the financial domain, and evaluate it on two financial sentiment analysis datasets FPB~\citep{Malo2013GoodDO} and FiQA SA~\citep{finqa}. BloombergGPT outperforms general models such as GPT-Neo~\citep{black_sid_2021_5297715}, OPT~\citep{zhang2022OPTOpenPretrained} and BLOOM~\citep{workshop2023BLOOM176BParameterOpenAccess} on both tasks.~\citet{Frackiewicz2023-us} leverage ChatGPT for social network analysis, enabling fast identification of key topics, sentiments, and influencers in the network, content generation, monitoring and flagging of harmful content in the community, and bringing profound changes to social network analysis.
2) \textbf{Stance detection}, which aims to determine the political, social or cultural stance of the author or speaker expressed in a text. This is of great significance for fields such as political analysis, public opinion monitoring, etc.
~\citet{zhang2023stance} applies ChatGPT to two common datasets for stance detection and achieves state-of-the-art or comparable performance. Stance detection is a natural language processing task that aims to identify the attitude of a text author towards a target, such as support, oppose, or neutral. It is useful for analyzing different perspectives on social, political, or cultural issues. For example, in political analysis, public opinion monitoring, and other domains, it is important to understand people’s views on certain events or claims.
~\citet{wu2023large} uses large language models to measure the latent ideology of politicians and scores US senators on a liberal-conservative scale by having ChatGPT choose the more liberal (or more conservative) senator in pairwise comparisons.
~\citet{tornberg2023chatgpt4} experiment on identifying the political party affiliation of Twitter posters and find that GPT-4 surpasses human experts and crowdsourced workers in accuracy and reliability.
3) \textbf{Hate speech detection}, which aims to identify and filter out words, phrases or sentences that may contain hate speech in a text. Some hate speech may be expressed in subtle ways, or use multiple languages and dialects, thus posing certain challenges for hate speech detection.~\citet{Huang_2023} uses ChatGPT to detect whether tweets imply hate speech, and successfully identifies 80\% of the tweets containing hate speech.
4) \textbf{Misinformation detection}, which aims to identify and filter out words, phrases or sentences that may contain misinformation in a text. Social media is the main platform for people to communicate, share and get information, but also a hotbed for misinformation dissemination. Misinformation can not only mislead the public, but also affect social trust, democratic participation and policy making. Misinformation detection can help prevent users from posting misinformation, thereby reducing the spread of false and misleading information. In the field of cancer misinformation detection, ChatGPT achieves an accuracy of 96.9\% \citep{johnson_using_2023}. The answers generated by ChatGPT show no significant difference from those of the National Cancer Institute (NCI) in terms of word count or readability. \citet{hoes2023using} finds that ChatGPT has a classification accuracy of 72\% on fact-checking tasks, with higher accuracy for true statements.

Other social science researchers utilize large language models for text generation tasks, one of the most challenging and creative tasks in NLP that involves automatically producing coherent, fluent and meaningful texts based on a given input or goal~\citep{gatt2018survey}. The typical applications of large language models in the field of social science for text generation tasks include:
1) \textbf{Natural language descriptions or explanations}, which mainly aims to improve the interpretability and credibility of results.
For example,~\citet{10.1145/3543873.3587320} proposes Chain of Explanation, a method to guide large language models such as GPT-Neo, T5~\citep{10.5555/3455716.3455856}, OPT and others to generate high-quality explanations for online hate speech. Although it makes significant improvements over previous methods, it still lags behind human level in terms of clarity and informativeness.~\citet{Huang_2023} uses ChatGPT to explain whether tweets imply hate speech, and finds that its generated explanations are clearer than those written by humans, while having no significant difference in informativeness with human-written ones. Large language models are applied in the field of linguistics to generate explanations that help improve the understanding and evaluation of linguistic phenomena and theories.~\citet{Chakrabarty2022FLUTEFL} uses GPT-3 to generate explanations for a figurative language Natural Language Inference dataset, and lets GPT-3 generate explanations for its judgments on figurative expressions, involving three types: Sarcasm, Simile, and Metaphor. 
2) \textbf{Future predictions}. Large language models are used for future prediction due to their powerful generative capabilities, but they are also limited by the complexity and uncertainty of the prediction scenarios or reality.~\citet{kalinin2023geopolitical} explores the use of GPT-3 as an information retrieval tool for predicting the Russian-Ukrainian conflict. The responses of GPT-3 are used as inputs for a game theory-based model of strategic behavior called "Predictioneer's game". But GPT-3 is limited by its reliance on prewar data and its inability to capture complex patterns of behavior.~\citet{jungwirth2023forecasting} uses GPT-3 to predict the future of the war in Ukraine by using GPT-3 to generate future scenarios while assessing the consistency within the scenarios.

There are also studies that have conducted comprehensive evaluations of large language models' performance across multiple content analysis tasks. For example,
~\citet{gilardi2023chatgpt} expand the application scope of ChatGPT to five text annotation tasks. Their results show that ChatGPT outperforms crowd workers in annotation tasks such as relevance, stance, topics, and frames detection, and is much lower in cost than the latter.
~\citet{ziems2023can} evaluates the performance of ChatGPT on multiple NLP tasks related to social science, and finds that it performs poorly on tasks such as event argument extraction, character tropes, implicit hate, and empathy classification, which involve complex structures or subjective expert taxonomies. In contrast, large language models achieve an accuracy of over 70\% on tasks such as misinformation, stance and emotion classification, which are based on objective basic facts or clearly verbalized definition labels.

\paragraph{Existing statistics analysis} is a research method that builds on the analysis of existing statistical data, which comes from official agencies, organizations, institutions or individuals, and covers various social phenomena and issues. Analysis of existing statistics can help researchers save time and cost, use existing information resources, and explore new research questions and hypotheses. In the following, we will discuss the applications of large language models in descriptive and inferential analysis as well as predictive analysis.

Large language models can be used to describe the characteristics of the sample or the relationship between variables, or to make inferences about causal processes, which refers to descriptive and inferential analysis~\citep{rubin2016empowerment}. 
For example, 
~\citet{chen-etal-2022-convfinqa} attempts to use large language models to understand financial reports and statements, but GPT-3 either copies the reasoning steps from the examples or gives incorrect reasoning, resulting in an accuracy below 50\%. BloombergGPT ~\citep{wuBloombergGPTLargeLanguage2023} is also applied to this task, but achieves only 43.41\% exact match accuracy. Although both large language models do not reach satisfactory results, OpenAI recently released more powerful ChatGPT and GPT-4, which might be able to perform better on this task.

Large language models can also be used to infer future trends and changes based on historical data, which refers to predictive analysis. This provides a basis for decision making, thus having a very wide range of applications in fields such as finance.
For example,
~\citet{Lopez_Lira_2023} demonstrates the potential of using ChatGPT to predict stock market returns. The authors simulate financial experts with ChatGPT and ask it to evaluate the impact of company-related headline news from the previous day on the stock price, based on sentiment analysis. They find that ChatGPT's sentiment scores had a significant positive correlation with the subsequent daily stock market returns.
~\citet{xie2023wall} find that using ChatGPT to predict stock trends based on historical price features and tweets has limited success, and even performs worse than traditional methods using only price features. However, ChatGPT is still recognized as having the potential to improve financial market prediction by utilizing social media sentiment and historical stock price information.

\paragraph{Conclusion}
Numerous studies have applied and evaluated large language models in a wide range of computational social science tasks, clarifying that large language models can significantly transform nonreactive research in three ways: 1) Assisting data annotators on human annotation teams. 2) Serving as zero/few-shot text analysis tools. 3) Bootstrapping challenging creative generation tasks.

The application of large language models in nonreactive research offers several \textbf{advantages}. Firstly, it can partially remove the limitations of data resources since large language models can achieve performance comparable to fine-tuned models in many social science tasks without extensive training. Secondly, they exhibit broad cross-disciplinary applicability, providing general solutions to a wide range of problems. Thirdly, they lower the entry barriers for usage. large language models have changed the scenario where researchers previously had to rely on statistical learning, machine learning, or deep learning methods to handle massive statistical data, which posed a high degree of difficulty and complexity. They are capable of interacting directly through text inputs instead of complex code or commands, providing a more direct and user-friendly interface, significantly lowering the technical barriers to using artificial intelligence for analysis. Consequently, researchers can focus more on the research questions they are interested in, rather than becoming excessively immersed in the intricacies of technical implementation.

However, there are some \textbf{limitations} to note: 1) Almost all large language models struggle with conversational and full document data, which limits common applications such as topic modeling. 2) large language models may have difficulty understanding the subtle and non-conventional language of expert taxonomies, which don't present in pre-trained data.

NLP researchers working to improve existing large language models for better support in nonreactive tasks can look at the following \textbf{future directions}: 1) the unique technical challenges of conversational, long-document, and cross-document reasoning, 2) in-domain training to teach LLMs to understand novel social constructs, 3) integration of specialized numerical analysis tools.

\subsection{Revisiting Applications Across Disciplines}

\begin{table}[ht]
\centering
\resizebox{\linewidth}{!}{
\begin{tabular}{lllll}
\toprule
Subject & Task & Dataset & Related work \\ \midrule
\multirow{2}{*}{Psychology} & \makecell[l]{Mental Health \\ State Detection} & \makecell[l]{Depression\_Reddit~\citep{pirina-coltekin-2018-identifying},\\ CLPsych15~\citep{coppersmith-etal-2015-clpsych},\\ Dreaddit~\citep{Turcan2019DreadditAR},\\ SAD~\citep{Mauriello2021SADAS}} & \makecell[l]{\citet{yang2023interpretable};\\ \citet{lamichhane2023evaluation};\citet{uludag2023chatgpt};\\ \citet{kjell2022natural}}\\ 
 & \makecell[l]{Personality Measurement} & - & \citet{rao2023can} \\
\midrule
\multirow{3}{*}{Politics} & Stance Detection & {\makecell[l]{SemEval-2016 Stance detecting  Dataset \\ \citep{mohammad-etal-2016-semeval}}} & \citet{ziems2023can}  \\
 & Ideology Detection & IBC~\citep{iyyer-etal-2014-political} & \citet{ziems2023can} \\
 & Misinformation Detection & Politifact Fact Check\citep{misra2022politifact} & \citet{hoes2023using} \\
 \midrule
 \multirow{2}{*}{Sociology} & Hate Speech Detection & \makecell[l]{LatentHatred 
 \\ \citep{Elsherief2021LatentHA}} & \citet{Huang_2023} \\
 & Misinformation Detection & \makecell[l]{Misinfon Reaction Frames\\ \citep{gabriel2022misinfo}} & \citet{ziems2023can} \\
 \midrule
 \multirow{7}{*}{Finance} & Sentiment Analysis & FPB~\citep{Malo2013GoodDO} & \citet{wuBloombergGPTLargeLanguage2023,Xie2023PIXIUAL} \\
 & Aspect Sentiment Analysis & \makecell[l]{FiQA SA~\citep{finqa}} & \citet{wuBloombergGPTLargeLanguage2023,Xie2023PIXIUAL} \\
 & Binary Classification & \makecell[l]{Headlines~\citep{sinha_impact_2021},\\ BigData~\citep{10020720},\\ StockNet~\citep{xu-cohen-2018-stock}, \\ CIKM18~\citep{Wu2018HybridDS}} & \citet{wuBloombergGPTLargeLanguage2023,xie2023wall,Xie2023PIXIUAL} \\
 & Named Entity Recognition & NER~\citep{salinas-alvarado-etal-2015-domain} & \citet{wuBloombergGPTLargeLanguage2023} \\
 & \makecell[l]{Named Entity Recognition+ \\Named Entity Disambiguation }& NER+NED~\citep{wuBloombergGPTLargeLanguage2023,Xie2023PIXIUAL} & \citet{wuBloombergGPTLargeLanguage2023} \\
 & Question Answering & ConvFinQA~\citep{chen-etal-2022-convfinqa} & \citet{wuBloombergGPTLargeLanguage2023,Xie2023PIXIUAL} \\
 
\bottomrule
\end{tabular}
}
\caption{A disciplinary perspective on AI for Social Science. This table presents a comparison of representative tasks, datasets used, and related work for each discipline.}
\label{tab:subject}
\end{table}

In this section, we revisited the application of large language models in social science from a disciplinary perspective, to provide readers with a more comprehensive understanding of research progress in this field. The representative tasks, datasets used, and related work for each discipline are outlined in Table~\ref{tab:subject}.

From a task perspective, we observe that the diverse tasks across disciplines can be summarized mainly into three categories: text classification, structured parsing, and natural language generation, which are relatively easier to handle in social science. However, more complex tasks such as aggregating mining on massive datasets, multi-document summarization or topic modeling may exceed the scope of transformer-based language models at present. From the algorithmic perspective, large language models can serve as a universal solution, meaning that almost all tasks can be addressed using the same approach, with the only difference being the design of prompts. The main drawbacks of using large language models as a solution could be related to issues like bias, high computational cost, difficulties in fine-tuning for specific tasks and so on.

\subsection{Discussions}
As mentioned above, large language models can be applied at every stage of social science research, improving efficiency across the board. 
More specifically, the practical applications of large language models in social science research predominantly fall into three directions: 1) replacing traditional NLP tools in data analysis, 2) assisting in creative work in research, 3) simulating humans as study objects. 
So far, extensive studies have thoroughly examined the superiority of large language models in the first direction.
However, the last two directions, which particularly emphasize large language models' human-like intelligence, are still in the exploratory stage without a systematic body of research.

In the future, several directions of AI for social science may lie in the following:
1) Further exploring the untapped potential of large language models in social science research. For instance, a large language model-based fully automated social science research pipeline could be developed, covering everything from hypothesis generation to hypothesis verification and even peer review.
2) Injecting domain-specific knowledge into large language models, thereby facilitating the development of expert models.
3) Establishing comprehensive benchmarks to measure the human-like attributes of large language models.
4) Integrating tools into large language models to enhance their capabilities in logical reasoning and mathematical derivation.
5) Developing multi-modal large models, which could improve their real-world understanding of human social behavior and systems.
6) Ethical and moral norms. Constructing ethical and moral frameworks for the functioning and application of large language models, thereby ensuring their responsible use.
\begin{table}[ht]
\centering
\begin{tabular}{@{}p{2cm}p{6cm}p{6cm}@{}}
\toprule
 & \multicolumn{1}{c}{of Human Beings} & \multicolumn{1}{c}{of AI} \\ \midrule
Psychology & Study the psychological phenomena, consciousness, and behavior of humans~\citep{GeneralPsychology}. Research spans a wide range of areas including consciousness, sensation, perception, cognition, emotion, personality, behavior and relationships. & Study personality, consciousness, ability, cognition, and more of AI agents. \\

Sociology & Study human social life, groups, and societies, ranging from institutions or human interactions at the micro-sociological level to social systems or structures at the macro-sociological level~\citep{Introductionsociology}. & Study interactions and social behaviors among multiple different AI agents. \\
Economics & Study the production, distribution, and consumption of goods and services~\citep{backhouse2002PenguinHistoryEconomics,krugman2009economics}. & Study the behavior and interaction of AI agents as economic agents. \\
Politics & Study the authoritative allocation of societal values~\citep{easton1955PoliticalSystemInquiry}. & Study the political behaviors and phenomena exhibited by AI agents, such as ideology, party affiliations, and political prudence.  \\
Linguistics & Study  language~\citep{halliday2006LanguageLinguistics}, including syntax, semantics, morphology, phonetics, phonology, pragmatics~\citep{farmer2010LinguisticsWorkbookCompanion}, and etc. & Study the language use patterns of AI agents and compare them to human language use.  \\ 
\bottomrule
\end{tabular}
\caption{The differences between social science of human beings and social science of AI in different sub-disciplines. The fundamental distinction between the two lies in the difference in their research subjects. The former investigates behavioral patterns within the human population, while the latter regards AI agents as intelligent entities and explores the behavioral patterns within the groups they form.}
\label{tab:of_diffs}
\end{table}

\section{Social science of AI} \label{social_science_of_ai}
Social science of AI refers to AI's social science. Specifically, we will focus on social science researches that use large language models as objects, with a particular emphasis on how they differ from traditional human behavior. Unlike AI for social science, the aim is not to make large language models mimic human behavior, but rather to explore the behavioral patterns of large language model agents themselves.
In Table~\ref{tab:of_diffs}, we give specific differences between social science of human beings and of large language models.

In this section, we will explore the social behavioral patterns of large language models as intelligent agents through the lens of various sub-disciplines within the social sciences. This emerging field has become increasingly significant due to several factors. Firstly, AI has demonstrated its ability to autonomously perform tasks in various domains. Secondly, research has shown that the collaboration of multiple AI agents can effectively enhance their performance. However, the behavior patterns, consequences, and impacts of AI collaboration are still not very clear. Additionally, the factors that drive changes in collaborative behaviors among AI agents are also not clearly understood. Similar to social science on humans, the ultimate goal of the Social Science of AI is to inform us about the behavioral traits exhibited by AI agents when they collaborate with each other and how to model and understand these behavioral traits. This type of research is of significant importance for the autonomous decision-making and control of future AI collectives.

\subsection{Psychology of AI}
\label{psychology}
Psychology of AI, or to say the psychology of machines, is typically defined as the scientific study of mind and behavior in AI agents~\citep{hagendorff2023MachinePsychologyInvestigating}. 
Extensive research in this field has been conducted with the enhanced capabilities of large language models. It has even been claimed that large language models may have a consciousness of its own. A typical example is a Google engineer's assertion that the conversational AI system, LaMDA, which he was developing, has become sentient and capable of thinking and reasoning like a human, leading to his suspension from work~\citep{tiku2023GoogleEngineerWho}. He believes that this large language model has attained the intelligence level of a 7-year-old or 8-year-old child.
In this section, we will organize the current advancements in psychology of AI,  according to the research content of the psychology~\citep{GeneralPsychology}, such as personality, cognition, and more.

\paragraph{Personality} refers to the sum of distinctive traits and characteristics that an individual possesses psychologically, emotionally, and behaviorally. Due to the stochastic nature of large language model's outputs, the personality of large language models refers to its overall tendency in generating responses. Researching the personality of large language models contributes to creating more human-friendly large language models. OpenAI's blog~\citep{HowShouldAI} pointed out that for models such as ChatGPT, the emotional bias of its output depends on both the pre-training stage and the fine-tuning stage. The sentiment tendency of the pre-training part comes from a large amount of text, and the value tendency of the fine-tuning stage may be related to the labeling staff or the fine-tuning task due to the different techniques used.

The most commonly used method for personality assessment of large language models is the utilization of questionnaires. With advancements in GPT-3 and its more powerful successor large language model, these language models are now capable of comprehending and fluently answering questions. The format of questionnaires is also highly compatible with language models. Survey questionnaires designed for human subjects typically require only minor adjustments in terms of formatting or vocabulary to be directly employed for personality testing~\citep{miotto2022WhoGPT3Exploration}. After adding the output method, it can be directly sent to the language model as a prompt for a reply. 

Research on large language model's personality has yielded some interesting conclusions.
~\citet{jiang2022MPIEvaluatingInducing} proposed the Machine Personality Inventory (MPI) dataset for evaluating the machine personality, finding that personality indeed exists in large language models.
~\citet{miotto2022WhoGPT3Exploration} used Hexaco questionnaire~\citep{ashton2009HEXACO60Short} to analyze GPT-3 and found that GPT-3 is generally a young woman whose personality level is consistent with the general tendency of human beings. In the assessment of human values, GPT-3 accords importance to every value, which can sometimes appear contradictory.
~\citet{li2023DoesGPT3Demonstrate} use Short Dark Triad (SD-3) and Big Five Inventory (BFI) tested GPT-3, InstructGPT, and FLAN-T5 and found that the tested large language models all showed darker than humans. The latter two are no better than GPT-3, even after fine-tuning. 
Furthermore, some studies have found that the personality traits of large language models can be effectively changed by fine-tuning~\citep{karra2023EstimatingPersonalityWhiteBox} or increasing memory~\citep{jiang2023PersonaLLMInvestigatingAbility}. This opens up the possibility of more related research.

\paragraph{Cognition} is about how humans understand, perceive, make decisions, and solve problems. Incorporating methodologies from cognitive psychology into large language models aids us in better understanding how these models process and address problems. 

There have been numerous studies investigating whether large language models are capable of human-like cognition.
For instance,~\citet{binz2023UsingCognitivePsychology} employed classic vignette-based and task-based experiments from the cognitive psychology literature to assess GPT-3's decision-making, information search, deliberation, and causal reasoning abilities. The results indicate that GPT-3 can achieve human-comparable performance on most tasks, but its behavior is highly influenced by how the vignettes are presented and it does not learn and explore in a human-like manner.
~\citet{han2022HumanlikePropertyInduction} focused on GPT-3's inductive reasoning ability. Experiment results suggested that GPT-3 can qualitatively mimic human performance for some inductive phenomena (especially those that depend primarily on similarity relationships), but fails to explain human inductive inferences, which may be due to GPT-3 not following the reasoning principles used by humans.
~\citet{webb2023EmergentAnalogicalReasoning} compared the analogical reasoning abilities of human reasoners and the text-davinci-003 variant of GPT-3 and found that large language models displayed a surprisingly strong capacity for abstract pattern induction, which may explain their abilities to reason about novel problems zero-shot.
~\citet{prystawski2022PsychologicallyinformedChainofthoughtPrompts} incorporated Chain of Thought (CoT) into the metaphor process inspired by cognitive psychology.
~\citet{collins2022StructuredFlexibleRobust} proposed a new benchmark for comparing the capabilities of humans and language models in problem-solving domains (planning and explanation generation). On this benchmark, humans are much more robust than large language models.
~\citet{kosoy2022UnderstandingHowMachines} tested the abilities of GPT-3 and PaLM in causal reasoning environments.
Besides direct reasoning abilities, biases in reasoning or decision-making processes have also received attention.
Various studies, in different manners and types, have collectively demonstrated the existence of certain biases in large language models, such as ChinChilla-7B/70B, CodeX, and ChatGPT, which are often similar to human cognitive biases ~\citep{dasgupta2022LanguageModelsShow,jones2022CapturingFailuresLarge,chen2023ManagerAIWalk}.

Theory of mind (ToM), another cognitive ability, refers to the capacity to comprehend others by attributing psychological states to them. Studies, exemplified by the false belief task, indicate that more advanced large language models perform better in ToM~\citep{kosinski2023TheoryMindMay}. Interestingly, we have located research papers on various models of the GPT series. In their study on GPT-3-davinci,~\citet{sap2022NeuralTheoryofMindLimits} noted that large language models cannot comprehend the intentions and reactions of social participants and infer the mental states of situational participants. However, when the research subject shifts to GPT-3.5 models such as text-davinci-002 and text-davinci-003, the large language models become more competent and closer to humans~\citep{trott2022LargeLanguageModels,dou2023ExploringGPT3Model}. 
Other studies of large language models' cognitive abilities include creativity tests, such as alternative tool tests for GPT-3~\citep{stevenson2022PuttingGPT3Creativity},  cognitive reflection tests and semantic illusions to examine the decision-making processes of large language models~\citep{hagendorff2022MachineIntuitionUncovering}, and assessments of GPT-4's cognitive abilities~\citep{dhingra2023MindMeetsMachine}.

\paragraph{Others}
Apart from personality and cognition, there are many other psychological aspects of AI being studied.
~\citet{feng2023BodySizeMetric} proposed that human body size affects the affordances of objects around them, and demonstrated that ChatGPT also exhibits this ability. They concluded that the embodied perception of ChatGPT (GPT-4 version) could be comparable to that of an average adult human, around 5 feet 6 inches tall.
~\citet{pellert2022AIPsychometricsUsing} suggested administering psychometric questionnaires to various models and requesting output, proposing the concept of AI Psychometrics. 
Further research has examined moral concepts and values in large language models~\citep{fischer2023WhatDoesChatGPT,jin2022WhenMakeExceptions}.

\paragraph{Conclusion}
In summary, extensive research has been conducted to explore the psychological features of large language model agents, from the perspectives of personality, cognition, and others, leading to many intriguing findings.

From a personality perspective, researchers generally find that large language models do exhibit personality tendencies, but these tendencies are not consistent and stable like those of humans. Instead, large language models are superpositions of cultural perspectives. These personality traits can be effectively altered through methods such as fine-tuning or enhancing memory capacities.

In the realm of cognitive abilities, studies have explored various facets, including induction, analogy, causal reasoning, theory of mind and so on. It is commonly found that the most advanced large language models, represented by GPT-3.5 and GPT-4, can demonstrate cognitive capabilities comparable to or even surpassing human abilities. These abilities improve with the evolution of the models. However, the cognitive modes employed by these language models are inconsistent with those of humans, and there is currently no universally accepted hypothesis to explain their cognitive abilities.

We believe that there are several pressing issues in this field that need to be addressed. These issues include: 1) Data leakage concerns. Much of the current research is based on classic psychological experiments to explore the cognitive capabilities of models, but it is unclear whether the test data is part of the language model's training data. 2) Unclear influence factors. The impact of factors such as the model's training objective, size, and data to its abilities has not been systematically analyzed.

\subsection{Sociology of AI}
\label{sociology}
Unlike the psychology of AI, which focuses on the behaviors of individual AI agents, the sociology of AI mainly studies the social behaviors and interactions of multiple AI agents.
It is important to note that we will primarily discuss researches that are similar to human sociology, but with a focus on AI agents as the research subjects.
This distinction sets us apart from other reviews that primarily concentrate on societal changes and issues arising from advancements in AI~\citep{joyce2021SociologyArtificialIntelligence}.

\paragraph{Social bias}
Many researches focused on social bias\footnote{Research on bias falls within the purview of social psychology, a cross-domain of sociology and psychology. We categorize it under the sociology section to emphasize its interactive nature.}, referring to unfair situations that emerge in large language models. With the widespread application of large language models, this can negatively impact user decision-making and interactions. The discussion about various biases in large language models started almost from the very beginning of their inception ~\citep{brown2020GPT3LanguageModels} and the topic is discussed with almost every large model released ~\citep{chowdhery2022PaLMScalingLanguage, openai2023GPT4TechnicalReport, touvron2023LLaMAOpenEfficient}. In papers, it often appears in sections like Limitations or Ethical Considerations. However, as a single section in a paper, the exploration is obviously insufficient, and subsequent research has focused on exploring social bias in large models. For instance, ~\citet{lucy2021GenderRepresentationBias} studied gender bias in GPT-3 and found that in the stories generated by GPT-3, females are more likely to be associated with family and appearance and are portrayed as less powerful than male characters. For minority groups, GPT-3 was also pointed out to have a bias against disabled people ~\citep{amin2022DisabilityLensBiases}. Some methods have been adopted to reduce bias or toxicity in large models. For instance, InstructGPT used a reinforcement learning approach to fine-tune the model, reducing toxic outputs ~\citep{ouyang2022InstructGPTTrainingLanguage}. We believe that there is a need for concerted efforts from researchers in sociology and AI to address these issues.

\paragraph{Social behavior}
Other researchers have considered the sociological behaviors among multiple large language model-driven agents. While there are not many researches in this area, interesting phenomena can still be observed.~\citet{park2023GenerativeAgentsInteractivea} created a rather interesting experimental environment. They built a sandbox where 25 agents live in a small town within the sandbox. The agents can wake up, make breakfast, and have small talk among themselves. They ended up exhibiting surprising social behaviors, such as spontaneously inviting and scheduling a party when a user assigns an agent to host one. In this research, through the introduction of a memory module and carefully designed processes, agents have become more human-like. Chirper\footnote{https://chirper.ai/} is an online community, where, unlike other communities like Reddit, all participants in Chirper are AI. You can set a backstory for a bot, and the bot will automatically post messages and interact. For instance, under the \#NewFriends topic, bots invite each other for walks in the park with their dogs, just like in a real human community, but all participants are AI. While these research efforts are intriguing, we believe they have yet to truly tap into the potential of AI sociology. We look forward to seeing more researchers delve into this field in the future.

\paragraph{Conclusion}
Currently, researchers have made some progress in social bias of AI, but there is still a dearth of studies specifically exploring sociological phenomena within the AI community.

The issue of social bias in AI has been a long-standing topic of discussion. Researchers have put forth various assessment methods and benchmarks to tackle bias issues, and they have made considerable strides in ensuring fairness and impartiality in language models when it comes to surface-level queries. Nevertheless, there is still much work to be done in this area, and several key challenges must be addressed: 1) Positive Stereotypes and essentializing Narratives. Even if a word may seem positive in sentiment, it can lead to harmful narratives. For example, praising women for being submissive and humble. 2) Implicit cognitive bias. Language model bias can be induced in various ways.

Despite our society witnesses a surge in the number of AI agents and the emergence of AI agent-exclusive communities, there has been limited research on the interaction patterns of language models within these environments. We anticipate a greater involvement of researchers from both the field of artificial intelligence and social science to systematically uncover the similarities and differences between the sociology of AI and human.

\subsection{Economics of AI}
\label{economics}
Economics of AI is the scientific study of how AI agents, as economic entities, produce, allocate, and consume goods and services~\citep{backhouse2002PenguinHistoryEconomics,krugman2009economics}.
Large language models have demonstrated their ability to act as economic agents, especially in the field of economics. Meanwhile, researchers have observed and compared their performance with that of humans in some economic experiments. In the following, we give an overview of both.

\paragraph{Economic expertise} measures large language models' professional knowledge, skills, and experience in the field of economics.
Although there are still some limitations, large language models show the ability of economic agents, including certain knowledge and understanding of economics, risk assessment and management ability, etc.
For economics, in the Test of Understanding in College Economics, ChatGPT outperformed 91$\%$ of students in the microeconomics and 99$\%$ in the macroeconomics.
For finance, Davinci and ChatGPT score 58$\%$ and 67$\%$, respectively, on the financial literacy test, 31$\%$ above the benchmark level~\citep{niszczota2023GPTFinancialAdvisor}. Large language models show some ability on the "Financial Investment Opinion Generation (FIOG)" task, but there is still room for improvement~\citep{son2023ClassificationFinancialReasoning}. GPT-3 can identify the potential impacts of climate change on economic growth, employment, poverty, inequality, and financial stability, and was also able to suggest some countermeasures~\citep{leippold2023ThusSpokeGPT3}.
Large language models match many biases in the expectations of existing professional humans and institutions for various financial and macroeconomic variables, including inflation, based on a sample of Journal news articles from 1984 to 2021~\citep{bybee2023SurveyingGenerativeAI}.
For operations management, ChatGPT earned a B- to B on the final exam for the MBA core course Operations Management but it doesn't work well for simple calculations or more advanced process analysis~\citep{terwiesch2023WouldChatGPT3}.
For marketing, the marketing content generated by ChatGPT matched and sometimes surpassed, human content on quantitative and qualitative measures~\citep{rivas2023MarketingChatGPTNavigating}. Conversations with consumers also show a high degree of intelligence and adaptability~\citep{rivas2023MarketingChatGPTNavigating}. ChatGPT was found to be more accurate and less biased than humans in problems with explicit mathematical or probabilistic properties, but also showed many human biases in complex, ambiguous and implicit problems~\citep{chen2023ManagerAIWalk}.
~\citet{cribben2023BenefitsLimitationsChatGPT} further discusses the strengths and limitations of ChatGPT in management science and operations management.
For accounting, large language models' performance is significantly lower than human capacity~\citep{wood2023ChatGPTArtificialIntelligence,bommarito2023GPTKnowledgeWorker}, especially when it comes to computation~\citep{bommarito2023GPTKnowledgeWorker}. In non-computational problems such as memory and understanding, large language models are almost human-level~\citep{bommarito2023GPTKnowledgeWorker}. On audit tasks, ChatGPT performed similar to or better than human experts on financial ratio analysis and text mining tasks, and slightly worse, but still acceptable, on log testing tasks~\citep{gu2023ArtificialIntelligenceCoPiloted}.

\paragraph{Microeconomics} is a branch of mainstream economics that studies the behavior of individual large language model agents in making decisions regarding the allocation of scarce resources and the interactions among these individuals~\citep{besanko2020microeconomics}.
Given that single or multiple large language model agents can naturally serve as the subjects of microeconomic research, a considerable amount of studies exist in this field. We will now organize them according to their research methodologies.

Some researchers use classical experiments in economics to study the behaviors and decision preferences of large language model agents.
For example,~\citet{aher2023UsingLargeLanguage} uses six large language models in the experiment and found that it can reproduce human behavior characteristics in the classic behavioral economics experiment -- ultimatum game.
~\citet{guo2023GPTAgentsGame} investigates the response of GPT-3.5-turbo in the ultimatum game with finite repetition and the Prisoner's Dilemma game. The results show that given a well-designed prompt, GPT can produce realistic results and exhibit behavior consistent with human behavior in some important respects, such as the positive correlation between the acceptance rate and proposed amount in the ultimatum game and the positive cooperation rate in the Prisoner's Dilemma game. The authors also observed some differences between GPT and human behavior. For example, in the repeated ultimatum game, the human agent generally decreased the amount of offers and the acceptance rate as the number of rounds increased, while the GPT agent showed no such tendency. This may indicate that GPT agents lack the ability of human agents to learn, adapt and punish.
In the repeated Prisoner's dilemma experiment,~\citet{phelps2023InvestigatingEmergentGoalLike} also finds the limitations of large language models in adjusting their behavior based on conditional reciprocity, and large language models can make different choices when it is injected with altruistic or selfish characteristics.
~\citet{johnson2023EvidenceBehaviorConsistent} explores whether AI agents exhibit behaviors consistent with self-interest and altruism in non-social decision-making tasks and dictator games. The results showed that only the most complex AI agent maximized its gains more in the non-social decision-making task, and that this AI agent also exhibited the most generous altruistic behavior in the dictator game, similar to humans.
~\citet{fu2023ImprovingLanguageModel} lets two large language models play the role of buyer and seller respectively in a bargaining game, the goal is to reach a deal, the buyer for the low price, and the seller for the high price. Experiments show that only strong large language models can improve the trading price through self-game and AI feedback.

Some other researchers investigate through survey research.
For example, ChatGPT's responses to different types of survey questions were compared with  human consumers and found to be able to answer in a manner consistent with economic theory and well-documented patterns of consumer behavior and matched estimates from a recent study that elicited human consumer preferences~\citep{brand2023UsingGPTMarketa}.
~\citet{goli2023LanguageTimePreferences} finds that GPT is more inclined to choose larger and later rewards in weak future tense references (FTR) languages, while smaller and earlier rewards in strong FTR languages, which is consistent with human preference. However, while human choices tend to prefer larger and later rewards as the reward gap increases, GPT choices do not.
Some questions are designed based on classical experiments in economics literature~\citep{horton2023LargeLanguageModels}, and it is found that large language models show similar behaviors and preferences to humans, such as fairness preference, loss aversion, state criteria, etc. However, there are also some differences, such as the attitude to risk, understanding of probability, sensitivity to language ,and so on.

\paragraph{Macroeconomics} is a branch of economics that deals with the performance, structure, behavior, and decision-making of an economy as a whole—for example, using interest rates, taxes, and government spending to regulate an economy's growth and stability~\citep{barro1997macroeconomics}.
Currently, there is no existing research on this topic, due to the nascent development of AI agents and the absence of a mature AI-driven economic framework.

\paragraph{Conclusion}

Extensive research has been conducted in the field of the economics of AI, with most of the studies focusing on the economic expertise, behavior and decision preferences of individual large language model agents.

In terms of economic expertise, large language model agents have demonstrated capabilities comparable to or even exceeding those of human experts in non-computational areas of economics, displaying a deep understanding of economic concepts. However, when it comes to more computational areas like accounting, they exhibit notably lower proficiency compared to humans. This observation suggests that language models possess the potential to reshape the quality and efficiency of future work within the field of economics.

In terms of economic behavior and decision-making preferences, large language model agents, when presented with well-defined prompts, can display behavior patterns consistent with human behavior in some significant aspects. They also show an understanding of human cooperative norms, such as altruism or selfishness, and can act in accordance with these norms. However, these agents also show certain limitations, such as their inability to adopt reasonable response strategies based on the cooperation patterns of others.

It's important to acknowledge the limitations of current research. Firstly, most results are based on testing GPT-3.5-turbo, and it remains uncertain whether these findings are universally applicable to all large language models. Additionally, these models have been trained on a significant amount of literature related to classical economics experiments, making it unclear how they would perform in more ecologically valid task environments they haven't encountered previously.

To address these challenges, we encourage the research community to further investigate: 1) The factors influencing intelligent agent behavior generated by language models across a wider range of social dilemmas, including model architecture, training parameters, and various partner strategies. 2) The development of new social dilemma games to assess language model capabilities, accompanied by task descriptions, rather than relying solely on existing literature anecdotes.

\subsection{Politics of AI}
\label{politics}
Politics of AI studies the political behaviors and phenomena exhibited by large language models, as political participants. More specifically, it involves the set of activities that are associated with making decisions in groups, or other forms of power relations among individuals, such as the distribution of resources or status~\citep{easton1955PoliticalSystemInquiry}.
Currently, research in this field primarily focuses on the political leanings and political prudence of large language model agents. Researchers have also conducted preliminary analyses to identify the factors that contribute to these characteristics and have proposed certain countermeasures.

\paragraph{Political leanings} refer to a large language model agent's preference for certain political beliefs, values, or views. Some researchers examine the general political leanings of large language models themselves, without providing any background settings.
For example, there exist some studies indicating that ChatGPT leans towards left-leaning liberal progressives~\citep{van2023chatgpt,hartmann2023PoliticalIdeologyConversational,gover2023PoliticalBiasLarge,liu2021MitigatingPoliticalBias}.
~\citet{mcgee2023ChatGptBiased} also find that ChatGPT favored liberal politicians over conservatives at least in some cases.
~\citet{rutinowski2023SelfPerceptionPoliticalBiases} indicates that ChatGPT seemed to favor progressive views.
~\citet{king2023GPT4AlignsNew} show that ChatGPT supports the New Liberal Party~\citep{king2023GPT4AlignsNew}. It is most likely to vote Green in Germany and the Netherlands~\citep{hartmann2023PoliticalIdeologyConversational}, and leans towards the Democratic Party in America, Lula in Brazil and the Labour Party in Britain~\citep{motoki2023MoreHumanHuman}.
In terms of territorial sovereignty, ChatGPT's responses to different territories are inconsistent and biased, sometimes at variance with Wikipedia information and UN resolutions~\citep{castillo-eslava2023RoleLargeLanguage}.
Other researchers examine the political leanings of large language model agents when given specific backgrounds. For example,~\citet{santurkar2023WhoseOpinionsLanguage} find that current large language models reflect views that are significantly inconsistent with those of American demographic groups. It's also very different from the views of certain demographic groups in a given description~\citep{santurkar2023WhoseOpinionsLanguage}.~\citet{argyle2022OutOneMany} draws the conclusion that large language models are significantly consistent with human samples in terms of political orientation, political knowledge, and political participation, and can capture the subtle differences existing in human samples.

Intuitively, the political bias of large language models may stem from the biases and tendencies of the training data sources themselves.
When large language models were fine-tuned in tweets from two politically inclined communities, Republicans and Democrats, the models show the political attitudes and worldviews of the two communities, respectively~\citep{jiang2022CommunityLMProbingPartisan}.
~\citet{feng2023PretrainingDataLanguage} further explored how political biases in pre-training data affect large language models.
~\citet{motoki2023MoreHumanHuman} agrees that ChatGPT and large language models may perpetuate or even amplify Internet and social media views of politics.
It is worth mentioning that ChatGPT's political bias can be influenced by its context and tends to copy the ideological bias of the prompt text~\citep{liu2021MitigatingPoliticalBias,gover2023PoliticalBiasLarge}.
The influence of populist framework is also explored~\citep{griffin2023SusceptibilityInfluenceLarge}. The experiment shows that, like human beings, it has a positive or negative influence on the news persuasiveness and political mobilization of large language models, while the anti-immigration frame has a negative influence, but there are some differences, such as the moderating effect of relative deprivation on the effect of the populist frame.

\paragraph{Conclusion}

The current exploration of the politics of AI is still in its nascent stage, with a primary focus on assessing the political biases inherent in large language model agents. Extensive research suggests that, in the absence of specific contextual information, representative large language model ChatGPT exhibits strong and systematic political biases, leaning notably towards the left end of the political spectrum. These biases can largely be attributed to inherent predispositions and patterns ingrained in the training data sources. Therefore, we can infer that the political leanings of language models can vary due to the diversity of their training data, demanding a thorough investigation. Notably, the political leanings of large language models can be significantly influenced by the contextual information provided. When these models are given a specific role or background information, they can align their political inclinations with those of individuals who share similar backgrounds. Even with implicit textual prompts, these models tend to capture and reproduce the ideological biases embedded in the text.

These observations underscore the concerning potential for language models to perpetuate and even amplify political perspectives on the internet, raising concerns about the influence they may wield over users and the potential for adverse political and electoral ramifications. Consequently, there is an urgent need to develop robust methods reliable methodologies for measuring the biases of large language models and poses a significant challenge to AI researchers aiming to construct more equitable and impartial language models. Additionally, exploring the capabilities of large language models in comprehending and handling political issues is an avenue of inquiry that deserves further attention.

\subsection{Linguistics of AI}
\label{linguistics}
Linguistics of AI aims at exploring the language use patterns of large language model agents, including syntax, semantics, morphology, phonetics, phonology, pragmatics and etc~\citep{farmer2010LinguisticsWorkbookCompanion}. We will emphasize large language model agents' unique language use patterns.

Researchers have made some interesting findings on the exploration of language use by large language models. In the following, we focus on the consistency and differences in language use between large language models and humans. Given the large number of existing studies, we will not dwell on the assessment of language proficiency in large language models.
The garden path sentence experiment is repeated on large language models, and it seems that large language models also have ambiguity in understanding language mechanisms~\citep{aher2023UsingLargeLanguage}.
And~\citet{diamond2023GenlangsZipfLaw} shows that GPT-generated languages statistically follow Zipf's law just like humans do.
\citet{cai2023DoesChatGPTResemble} further explores the consistency and differences between ChatGPT and humans in language use and finds that it is able to associate unfamiliar words with different meanings based on form, reinterpret unreasonable sentences that may be corrupted by noise, etc. as well as humans. It does not like to use shorter words to convey less information and does not use context to disambiguate syntax.
In addition, the degenerated version, GPT-D, obtained by changing the parameters of GPT-2, has language features associated with Alzheimer's disease, such as repetition, semantic loss, and grammatical errors~\citep{li2022GPTDInducingDementiarelated}, which contributes to a better understanding of the internal mechanisms of generative neural language models. 
It is worth mentioning that large language models' preferences for time and reward are similar to human decision-makers and are influenced by future tense references in language~\citep{goli2023LanguageTimePreferences}.

\paragraph{Conclusion}

These studies delve deep into the linguistic features of large language models and compare them to human, offering us initial insights into the language usage patterns of these models. For instance, large language models, like humans, can understand unfamiliar words based on affixes, may misinterpret sentences as typical but grammatically incorrect meanings, and follow similar vocabulary distributions statistically.

In future research, we believe that combining a linguistic perspective with a natural language processing perspective might offer a better understanding of the internal mechanisms of generative neural language models. This integration could serve as a crucial foundation for further exploring and explaining the language and intelligence capabilities of large language models. For example, the consistency in vocabulary distribution with humans can be attributed to the fact that language models inherently learn the probabilities of word occurrences and context combinations in language. Additionally, the comprehension of word forms by language models is likely influenced by the role of tokenization, enabling language models to understand the meanings of affixes in English and thus combine to comprehend previously unseen words.

\subsection{Discussions and Future Works}
We have conducted a comprehensive review of research on the social behavior of large language model agents based on several representative sub-disciplines of social science. Notably, current research is predominantly focused on exploring the social behaviors exhibited by individual large language model agents, with a lack of study on large language model agent groups or systems. This could be attributed to the present absence of instances of large language model agent groups or systems.

Furthermore, we have identified some limitations in the existing research.
Firstly, the testing of large language models' capabilities or characters is greatly associated with the version and parameter settings of the experimental model. For widely used models such as ChatGPT, versions vary over time, and responses from ChatGPT to the same question may differ at different times, which may affect the reproducibility of experimental results~\citep{tu2023ChatLogRecordingAnalyzing}. The research by~\citet{miotto2022WhoGPT3Exploration} confirmed that changes in temperature affect the personality tendencies of the model. 
Secondly, large language models are sensitive to the order of given prompts~\citep{lu2022FantasticallyOrderedPrompts,zhao2021CalibrateUseImproving}, a factor that should be taken into account in experiments. 
Thirdly, experiments exploring the psychology of large language models require careful design. The large language models training process uses a vast amount of textual material, which may contain classic psychological scenarios that could impact experimental results, but this issue is considered in only a few documents~\citep{binz2023UsingCognitivePsychology}. Different evaluation methods for the same ability could lead to different results. In the exploration of GPT-3's mental abilities,~\citet{sap2022NeuralTheoryofMindLimits} and~\citet{bojic2023SignsConsciousnessAi} had disparities due to differences in the experimental process. 
Finally, the direct application of human evaluation methods to large language models remains a question worth considering.~\citet{ullman2023LargeLanguageModels} found that large language models often fail in tests when false belief tasks are added with various disturbances, suggesting that large models actually lack ToM. The conclusion is that while ToM tests are effective for humans, they may not reasonably assess the abilities of large language models.

To address the aforementioned challenges, the future directions for social science of AI lie in:
1) \textbf{Establishment of a systematic theory for the social science of AI}, similar to the social science of humans. This will aid in connecting and organizing the currently fragmented research efforts, and allow for a comprehensive examination of AI agents' social characteristics as intelligent entities themselves.
2) \textbf{In-depth exploration of social phenomena in large language model agent groups or systems}, delving into the complex interactions and dynamics that may emerge.
3) \textbf{Standardized experimental designs}, such as those pertaining to model versions, parameters, and prompts, to minimize result deviations caused by variations in experimental designs.
4) \textbf{Tailored evaluation methods for large language model agents}, considering that directly applying human evaluation methods to large language model agents may not result in reasonable assessments.
5) \textbf{Combination of social science theories with AI theories}. We note that the study of large language models shares some similarities with the study of social science. For instance, both the thought process of large language model agents and humans can be seen as a 'black box' to some extent. We cannot fully grasp the various reasoning or cognitive processes inside the large language models and the human body, but we can gauge them using external tools such as performance metrics on specific tasks and observable behaviors. For this black box, we have both attempted to probe the internal mechanisms or, in other words, to 'open' the box. For the NLP community, this endeavor involves parameter tuning, knowledge injection, and modification; for the field of social science, it may involve areas like neuroscience. We hope these commonalities can provide inspiration for further exploration in the future.
\section{Public Tools and Resources}
\label{resource}

To facilitate the utilization of large language models for social science research, there already exist several publicly available tools and resources as aids. In this section, we focus on introducing simulation tools and platforms that are based on large language models, taking into account that other applications mainly rely on direct use or simple script-based invocation. Based on this, we conduct a systematic analysis of simulation requirements and compare the functionalities of various platforms.

\subsection{Public Simulation Tools}

The evolution of human-like abilities in large language models has opened up new possibilities for computational simulation, leading to the emergence of various simulation platforms or tools based on these models. 
In the following, we collect and compare existing open source large language model-based simulation tools.

\paragraph{SkyAGI} is a Python package that demonstrates the emerging capability of large language models in simulating believable human behaviors. It offers a role-playing game experience that is highly engaging and immersive. Unlike previous AI-based NPC systems, SkyAGI's NPCs generate incredibly realistic human responses. This platform has significant potential for rethinking game development, particularly in the area of NPC script writing.

\paragraph{AgentVerse} is a versatile framework designed to streamline the process of creating custom multi-agent environments for large language models. It offers efficient environment building tools, allowing researchers to easily construct basic environments like chat rooms for large language models by defining settings and prompts. Additionally, it supports customizable components, empowering researchers to create their own multi-agent environments according to their specific requirements. Furthermore, AgentVerse integrates tools (plugins) to enhance functionality, currently supporting BMTools. This platform enables researchers to optimize their experiments and analyses in a seamless and efficient manner.

\paragraph{LangChain} is a powerful platform designed to assist developers in building applications through composability, harnessing the capabilities of large language models. One of its key features is the ability to create agents. Agents utilize LLMs to make decisions, perform actions, observe outcomes, and iterate until their objectives are achieved. 
Langchain does not provide out-of-the-box usage similar to other frameworks, but it implements interfaces such as Time-weighted vector store retriever,which plays a very important role in the agent's memory. You need to write your own code to implement interaction between multiple agents and other functions.

\paragraph{GenerativeAgents} is the implementation of \citep{park2023GenerativeAgentsInteractivea}. Although the author does not propose it as a platform, users can still customize a simulation environment by modifying character configuration, code, etc. As mentioned in the paper, it provides a map for better visualization and interaction with the environment,and this provides users with more possibilities.

\paragraph{Agents} is an open source framework for building autonomous language singletons\citep{zhou2023agents}. It supports long and short-term memory, tool usage, etc. What differentiates Agents from other frameworks is that it supports more detailed control of agents through SOP. SOP defines the state of the agent and the transition relationship between states. In other words, the process can be configured using more complex configuration files rather than modifying the code.

\paragraph{AgentLab} is a large language model-based simulation toolkit for social science research. It allows the creation of multiple intelligence agents with heterogeneous features by inputting different profiles. Each intelligence agent can learn knowledge either through model weights (i.e., fine-tuning the model based on its experience) or model inputs (i.e., incorporating knowledge into an input message). Additionally, it supports the customization of social backgrounds as required. Once all experimental conditions are set, the platform can automatically facilitate human-like interactions among agents.

\begin{table}[]
\centering
\begin{tabular}{cccccc}
\hline
\multirow{2}{*}{Name} & Single Agent & \multicolumn{2}{c}{Multi Agents} & \multicolumn{2}{c}{Environment} \\ \cline{2-6} 
 & Tool Use & Interact & Scheduling & Physical Environment & Social Background \\ \hline
SkyAGI\tablefootnote{https://github.com/litanlitudan/skyagi} & no & auto+influence & serial & no & no \\
AgentVerse\tablefootnote{https://github.com/OpenBMB/AgentVerse} & yes & auto & serial+parallelism & no & yes \\
LangChain\tablefootnote{https://github.com/hwchase17/langchain}\textsuperscript{*} & yes & - & - & - & - \\
Generative Agents\tablefootnote{https://github.com/joonspk-research/generative\_agents} & no & auto & serial+parallelism  & yes & yes \\
Agents\tablefootnote{https://github.com/aiwaves-cn/agents} & yes & auto+influence & serial+parallelism & no & yes \\
AgentLab\tablefootnote{https://github.com/renmengjie7/AISimuToolKit} & yes & auto & serial+parallelism & no & yes \\
\hline
\end{tabular}
\textsuperscript{*}As mentioned in the main text, you need to write the code for the simulation yourself.
\caption{Functional comparison of existing open source simulation toolkits. For single-agent scenarios, current simulation toolkits universally facilitate diverse population modeling, exclusively employ the "prompt" method for internalizing memory, and only accommodate predefined models rather than allowing for user-defined models for simulation. Consequently, for the sake of simplicity, these three columns have been omitted from the table.}
\label{tab:comparsion_between_tools}
\end{table}

\subsection{Core Functions of Simulation Tools}
Based on a summary of above existing toolkits, as well as the complexity and interactivity of the real world, we have formulated key features of simulation platforms using a functional hierarchy framework to satisfy various needs and summarized and compared the above toolkits based on this framework in Table {\ref{tab:comparsion_between_tools}}. Specifically, these features can be classified into three levels: single-agent, multi-agent, and environment.

\begin{itemize}
    \item single-agent, the basic building block of simulation
        \begin{itemize}
            \item able to generate human profiles based on key information, enabling researchers to model populations with different characteristics using large language models.
            \item support for using tools to complete some tasks. Using tools is an essential part of human life. The support for using tools can expand the scope of simulation experiments.
            \item able to maintain and internalize its memory, including short memory and long memory. Inspired by Stanford's work~\citep{park2023GenerativeAgentsInteractivea}, and based on the roadmap of large language models~\citep{zhao2023survey}, we believe that prompt and fine-tuning mechanisms exist that can be used to perform short-term and long-term memory tasks respectively.
            \item support for multiple and pluggable models for simulation, enabling researchers to choose different models based on their needs and assumptions. This flexibility can facilitate innovation and customization of research, enabling researchers to better adapt to different research scenarios~\citep{li2022GPTDInducingDementiarelated}.
        \end{itemize}
    \item multi-agent, which interact to form the society
        \begin{itemize}
            \item able to interact spontaneously or under the influence of the researcher.
            \item support complex interaction rules: serial, parallel, and both. Serial means that two operations cannot be executed at the same time, which means there is an order of precedence. Specifically, we believe there are three modes, sequential, bidding, and specified\footnote{A special case of bidding mode}, referring to langchain\footnote{https://python.langchain.com/en/latest/use\_cases/agent\_simulations.html}. Parallel means that the two actions occur in overlapping periods. Both means supporting both serial rule and parallel rule.
        \end{itemize}
    \item environment, the container for simulation, including the physical environment and social background
        \begin{itemize}
            \item able to interact with the physical environment. It means that agents have an impact on the physical environment, such as consumption of water, electricity and food and are also affected by the physical environment. A simple example is visibility. For example, when a report is given in a conference room, people outside the meeting room cannot directly know the specific content and progress of the report.
            \item able to include social backgrounds, such as economic background, political background, cultural background, social norms, institutions, etc. It's necessary because the social background of each era is different, and human beings in the era also have their characteristics. Under different social backgrounds, human beings will have completely different cognition, decision-making behavior, etc. 
        \end{itemize}
\end{itemize}

The above three layers are only a minimum set of our large language models simulation tool imagination. A good and convenient toolkit should also be oriented to researchers, provide good visualization, automatically build profiles and prompts, etc.

We noticed that the current open-source simulation toolkits still have some limitations.
Firstly, the current implementation of agents' knowledge learning in these platforms is quite simplistic, primarily relying on prompts while overlooking the more natural learning method of fine-tuning employed by language models.
Secondly, these platforms restrict the underlying large language model for agents, which hinders the agent's ability to adapt flexibly to diverse research contexts.
Lastly, the current platforms struggle to provide adequate support for effective interaction between agents and their environments. We believe that addressing these issues will contribute to a more comprehensive agent simulation.

\subsection{Discussions and Future Works}
Currently, there are still some common issues in simulation platforms based on large language models.
Firstly, there remains a significant gap between large language model-based agents and real-life human behavior. This gap stems partly from the large language model itself. While natural language can express the vast majority of meanings, sometimes information like visuals and sound is indispensable. We believe that future multimodal large models can do better. Even within the scope of natural language itself, large language models still struggle to perfectly replicate the diversity of human behavior, especially when handling complex tasks. This may be due to large language models lacking a "theory of mind," which refers to the ability to understand the mental states and intentions of others. This deficiency hinders their performance in simulating complex interactions among multiple agents.
Secondly, real-life social contexts, physical environments, and operational rules exhibit vast variations, making it challenging to build systems that involve interactions among multiple agents.  This complex task often necessitates social science researchers to acquire programming knowledge and natural science researchers to gain an understanding of social science principles. Furthermore, considering temporal aspects adds another layer of complexity. Modeling temporal properties involves accounting for interaction behaviors, dynamic changes, and event sequences, presenting researchers with even greater challenges.

The future of social simulation platforms may involve: 1) Incorporating cognitive theories as frameworks for agent decision-making, thus enhancing the human-like aspects or correctness of agents, as well as providing interpretability for their behavior. 2) Harnessing the multimodal capabilities of large language models to improve agents' ability to acquire and express information. 3) Establishing an evaluation framework to qualitatively assess simulation platforms.
\section{Conclusion} \label{conclusion}
In this paper, we surveyed the latest developments at the intersection of large language models and social science. We propose a dichotomy to outline the progression in this field, encompassing 'AI for Social Science' and the 'Social Science of AI'.
We note that large language models can be integrated into various stages of social science research, serving as auxiliary tools, a source of inspiration, annotation tools, content analysis tools, and so on, thereby enhancing efficiency. While large language models as tools have the advantages of speed, cost-effectiveness, ethically risk-free experimentation, and a low barrier to entry, the reliability and authenticity of their generated text need to be verified. Whether they can replace humans in conducting experiments and surveys also remains an open question. Therefore, researchers need to consider the additional cost of validation and the risk of bias when using these models.
Furthermore, both the large language models themselves and the communities formed around them have exhibited some unique and intriguing behaviors. However, the enduring and unresolved issue in the field of social science is whether machines or intelligent agents should be the subject of social science research. We emphasize the promising future of this research direction, which will become increasingly important as AI agents become more prevalent in daily life.
These two directions are complementary. The latter can guide the development of the former, while the former can enhance the efficiency of the latter's research.
In conclusion, we believe that while AI cannot replace sociologists, it will become deeply integrated into the research process. Social scientists will play a significant role in guiding the development of AI.

As for future works, there is a need for an in-depth study into how and to what extent AI influence human behavior during computer-human interaction, the third intersection of AI with social science. Unlike AI for social science and social science of AI which address social issues within specific groups – the former concentrating on human populations and the latter on AI agent populations – this direction primarily explores new societal issues arising from interactions between AI and humans and introduces new methodologies. It is essential for gaining valuable insights on how to effectively utilize large language models in human-computer interactions and successfully accomplish social-oriented objectives.

\section*{Acknowledgement}
We sincerely thank all reviewers for their insightful comments and valuable suggestions. This research work is supported by the National Natural Science Foundation of China under Grants no. 62122077, 62106251, 62306303. Xianpei Han is sponsored by CCF-BaiChuan-Ebtech Foundation Model Fund.

\printcredits

\bibliographystyle{cas-model2-names}

\bibliography{reference}

\begin{thebibliography}{186}
\expandafter\ifx\csname natexlab\endcsname\relax\def\natexlab#1{#1}\fi
\providecommand{\url}[1]{\texttt{#1}}
\providecommand{\href}[2]{#2}
\providecommand{\path}[1]{#1}
\providecommand{\DOIprefix}{doi:}
\providecommand{\ArXivprefix}{arXiv:}
\providecommand{\URLprefix}{URL: }
\providecommand{\Pubmedprefix}{pmid:}
\providecommand{\doi}[1]{\href{http://dx.doi.org/#1}{\path{#1}}}
\providecommand{\Pubmed}[1]{\href{pmid:#1}{\path{#1}}}
\providecommand{\bibinfo}[2]{#2}
\ifx\xfnm\relax \def\xfnm[#1]{\unskip,\space#1}\fi
\bibitem[{Abbas(2023)}]{abbas2023uses}
\bibinfo{author}{Abbas, M.}, \bibinfo{year}{2023}.
\newblock \bibinfo{title}{Uses and misuses of chatgpt by academic community: An overview and guidelines}.
\newblock \bibinfo{journal}{Available at SSRN 4402510} .
\bibitem[{Aggarwal and Zhai(2012)}]{aggarwal2012survey}
\bibinfo{author}{Aggarwal, C.C.}, \bibinfo{author}{Zhai, C.}, \bibinfo{year}{2012}.
\newblock \bibinfo{title}{A survey of text classification algorithms}.
\newblock \bibinfo{journal}{Mining text data} , \bibinfo{pages}{163--222}.
\bibitem[{Aher et~al.(2023)Aher, Arriaga and Kalai}]{aher2023UsingLargeLanguage}
\bibinfo{author}{Aher, G.V.}, \bibinfo{author}{Arriaga, R.I.}, \bibinfo{author}{Kalai, A.T.}, \bibinfo{year}{2023}.
\newblock \bibinfo{title}{Using large language models to simulate multiple humans and replicate human subject studies}, in: \bibinfo{booktitle}{International Conference on Machine Learning}, \bibinfo{organization}{PMLR}. pp. \bibinfo{pages}{337--371}.
\bibitem[{Alvarado et~al.(2015)Alvarado, Cesar, Verspoor et~al.}]{salinas-alvarado-etal-2015-domain}
\bibinfo{author}{Alvarado, S.}, \bibinfo{author}{Cesar, J.}, \bibinfo{author}{Verspoor, K.}, et~al., \bibinfo{year}{2015}.
\newblock \bibinfo{title}{Domain adaption of named entity recognition to support credit risk assessment}, in: \bibinfo{booktitle}{Proceedings of the Australasian Language Technology Association Workshop 2015}, \bibinfo{address}{Parramatta, Australia}. pp. \bibinfo{pages}{84--90}.
\newblock \URLprefix \url{https://aclanthology.org/U15-1010}.
\bibitem[{Amin and Kabir(2022)}]{amin2022DisabilityLensBiases}
\bibinfo{author}{Amin, A.A.}, \bibinfo{author}{Kabir, K.S.}, \bibinfo{year}{2022}.
\newblock \bibinfo{title}{A {{Disability Lens}} towards {{Biases}} in {{GPT-3 Generated Open-Ended Languages}}}.
\newblock \DOIprefix\doi{10.48550/arXiv.2206.11993}, \href{http://arxiv.org/abs/2206.11993}{\tt arXiv:2206.11993}.
\bibitem[{Andreas(2022)}]{andreas2022language}
\bibinfo{author}{Andreas, J.}, \bibinfo{year}{2022}.
\newblock \bibinfo{title}{Language models as agent models}, in: \bibinfo{booktitle}{Findings of the Association for Computational Linguistics: EMNLP 2022}, pp. \bibinfo{pages}{5769--5779}.
\bibitem[{Argyle et~al.(2023)Argyle, Busby, Fulda, Gubler, Rytting and Wingate}]{argyle2022OutOneMany}
\bibinfo{author}{Argyle, L.P.}, \bibinfo{author}{Busby, E.C.}, \bibinfo{author}{Fulda, N.}, \bibinfo{author}{Gubler, J.R.}, \bibinfo{author}{Rytting, C.}, \bibinfo{author}{Wingate, D.}, \bibinfo{year}{2023}.
\newblock \bibinfo{title}{Out of one, many: Using language models to simulate human samples}.
\newblock \bibinfo{journal}{Political Analysis} \bibinfo{volume}{31}, \bibinfo{pages}{337--351}.
\bibitem[{Ashton and Lee(2009)}]{ashton2009HEXACO60Short}
\bibinfo{author}{Ashton, M.C.}, \bibinfo{author}{Lee, K.}, \bibinfo{year}{2009}.
\newblock \bibinfo{title}{The {{HEXACO}}\textendash 60: {{A Short Measure}} of the {{Major Dimensions}} of {{Personality}}}.
\newblock \bibinfo{journal}{Journal of Personality Assessment} \bibinfo{volume}{91}, \bibinfo{pages}{340--345}.
\newblock \DOIprefix\doi{10.1080/00223890902935878}.
\bibitem[{Ayd{\i}n and Karaarslan(2022)}]{aydin2022openai}
\bibinfo{author}{Ayd{\i}n, {\"O}.}, \bibinfo{author}{Karaarslan, E.}, \bibinfo{year}{2022}.
\newblock \bibinfo{title}{Openai chatgpt generated literature review: Digital twin in healthcare}.
\newblock \bibinfo{journal}{Available at SSRN 4308687} .
\bibitem[{Backhouse(2002)}]{backhouse2002PenguinHistoryEconomics}
\bibinfo{author}{Backhouse, R.}, \bibinfo{year}{2002}.
\newblock \bibinfo{title}{The {{Penguin}} History of Economics}.
\newblock \bibinfo{publisher}{{Penguin}}, \bibinfo{address}{{London}}.
\bibitem[{Bail(2023)}]{bail2023can}
\bibinfo{author}{Bail, C.A.}, \bibinfo{year}{2023}.
\newblock \bibinfo{title}{Can generative ai improve social science?} .
\bibitem[{Banker et~al.(2023)Banker, Chatterjee, Mishra and Mishra}]{banker2023machine}
\bibinfo{author}{Banker, S.}, \bibinfo{author}{Chatterjee, P.}, \bibinfo{author}{Mishra, H.}, \bibinfo{author}{Mishra, A.}, \bibinfo{year}{2023}.
\newblock \bibinfo{title}{Machine-assisted social psychology hypothesis generation} .
\bibitem[{Barro(1997)}]{barro1997macroeconomics}
\bibinfo{author}{Barro, R.J.}, \bibinfo{year}{1997}.
\newblock \bibinfo{title}{Macroeconomics}.
\newblock \bibinfo{publisher}{MIT Press}.
\bibitem[{Besanko and Braeutigam(2020)}]{besanko2020microeconomics}
\bibinfo{author}{Besanko, D.}, \bibinfo{author}{Braeutigam, R.}, \bibinfo{year}{2020}.
\newblock \bibinfo{title}{Microeconomics}.
\newblock \bibinfo{publisher}{John Wiley \& Sons}.
\bibitem[{Bhattacherjee(2012)}]{anol2016social}
\bibinfo{author}{Bhattacherjee, A.}, \bibinfo{year}{2012}.
\newblock \bibinfo{title}{Social science research: Principles, methods, and practices}.
\newblock \bibinfo{publisher}{USA}.
\bibitem[{Binz and Schulz(2023)}]{binz2023UsingCognitivePsychology}
\bibinfo{author}{Binz, M.}, \bibinfo{author}{Schulz, E.}, \bibinfo{year}{2023}.
\newblock \bibinfo{title}{Using cognitive psychology to understand {{GPT-3}}}.
\newblock \bibinfo{journal}{Proceedings of the National Academy of Sciences} \bibinfo{volume}{120}, \bibinfo{pages}{e2218523120}.
\newblock \DOIprefix\doi{10.1073/pnas.2218523120}.
\bibitem[{Bisbee et~al.(2023)Bisbee, Clinton, Dorff, Kenkel and Larson}]{bisbee2023artificially}
\bibinfo{author}{Bisbee, J.}, \bibinfo{author}{Clinton, J.}, \bibinfo{author}{Dorff, C.}, \bibinfo{author}{Kenkel, B.}, \bibinfo{author}{Larson, J.}, \bibinfo{year}{2023}.
\newblock \bibinfo{title}{Artificially precise extremism: How internet-trained llms exaggerate our differences} .
\bibitem[{Black et~al.(2021)Black, Leo, Wang, Leahy and Biderman}]{black_sid_2021_5297715}
\bibinfo{author}{Black, S.}, \bibinfo{author}{Leo, G.}, \bibinfo{author}{Wang, P.}, \bibinfo{author}{Leahy, C.}, \bibinfo{author}{Biderman, S.}, \bibinfo{year}{2021}.
\newblock \bibinfo{title}{{GPT-Neo: Large Scale Autoregressive Language Modeling with Mesh-Tensorflow}}.
\newblock \URLprefix \url{https://doi.org/10.5281/zenodo.5297715}, \DOIprefix\doi{10.5281/zenodo.5297715}.
\bibitem[{Bojic et~al.(2023)Bojic, Stojkovi{\'c} and Joli{\'c}~Marjanovi{\'c}}]{bojic2023SignsConsciousnessAi}
\bibinfo{author}{Bojic, L.}, \bibinfo{author}{Stojkovi{\'c}, I.}, \bibinfo{author}{Joli{\'c}~Marjanovi{\'c}, Z.}, \bibinfo{year}{2023}.
\newblock \bibinfo{title}{Signs of {{Consciousness}} in {{Ai}}: {{Can Gpt-3 Tell How Smart}} it {{Really}} is?}
\newblock \DOIprefix\doi{10.2139/ssrn.4399438}.
\bibitem[{Bommarito et~al.(2023)Bommarito, Bommarito, Katz and Katz}]{bommarito2023GPTKnowledgeWorker}
\bibinfo{author}{Bommarito, J.}, \bibinfo{author}{Bommarito, M.}, \bibinfo{author}{Katz, D.M.}, \bibinfo{author}{Katz, J.}, \bibinfo{year}{2023}.
\newblock \bibinfo{title}{{{GPT}} as {{Knowledge Worker}}: {{A Zero-Shot Evaluation}} of ({{AI}}){{CPA Capabilities}}}.
\newblock \DOIprefix\doi{10.48550/arXiv.2301.04408}, \href{http://arxiv.org/abs/2301.04408}{\tt arXiv:2301.04408}.
\bibitem[{Brand et~al.(2023)Brand, Israeli and Ngwe}]{brand2023UsingGPTMarketa}
\bibinfo{author}{Brand, J.}, \bibinfo{author}{Israeli, A.}, \bibinfo{author}{Ngwe, D.}, \bibinfo{year}{2023}.
\newblock \bibinfo{title}{Using {{GPT}} for {{Market Research}}}.
\newblock \bibinfo{journal}{SSRN Electronic Journal} \DOIprefix\doi{10.2139/ssrn.4395751}.
\bibitem[{van~den Broek(2023)}]{van2023chatgpt}
\bibinfo{author}{van~den Broek, M.}, \bibinfo{year}{2023}.
\newblock \bibinfo{title}{Chatgpt’s left-leaning liberal bias}.
\newblock \bibinfo{journal}{University of Leiden} \URLprefix \url{https://www.universiteitleiden.nl/binaries/content/assets/algemeen/bb-scm/nieuws/political_bias_in_chatgpt.pdf}.
\bibitem[{Brown et~al.(2020)Brown, Mann, Ryder et~al.}]{brown2020GPT3LanguageModels}
\bibinfo{author}{Brown, T.B.}, \bibinfo{author}{Mann, B.}, \bibinfo{author}{Ryder, N.}, et~al., \bibinfo{year}{2020}.
\newblock \bibinfo{title}{{{GPT3--Language Models}} are {{Few-Shot Learners}}}.
\newblock \DOIprefix\doi{10.48550/arXiv.2005.14165}, \href{http://arxiv.org/abs/2005.14165}{\tt arXiv:2005.14165}.
\bibitem[{Bryman(2016)}]{bryman2016social}
\bibinfo{author}{Bryman, A.}, \bibinfo{year}{2016}.
\newblock \bibinfo{title}{Social research methods}.
\newblock \bibinfo{publisher}{Oxford university press}.
\bibitem[{Bybee(2023)}]{bybee2023SurveyingGenerativeAI}
\bibinfo{author}{Bybee, L.}, \bibinfo{year}{2023}.
\newblock \bibinfo{title}{Surveying {{Generative AI}}'s {{Economic Expectations}}}.
\newblock \DOIprefix\doi{10.48550/arXiv.2305.02823}, \href{http://arxiv.org/abs/2305.02823}{\tt arXiv:2305.02823}.
\bibitem[{Cai et~al.(2023)Cai, Haslett, Duan, Wang and Pickering}]{cai2023DoesChatGPTResemble}
\bibinfo{author}{Cai, Z.G.}, \bibinfo{author}{Haslett, D.A.}, \bibinfo{author}{Duan, X.}, \bibinfo{author}{Wang, S.}, \bibinfo{author}{Pickering, M.J.}, \bibinfo{year}{2023}.
\newblock \bibinfo{title}{Does {{ChatGPT}} resemble humans in language use?}
\newblock \DOIprefix\doi{10.48550/arXiv.2303.08014}, \href{http://arxiv.org/abs/2303.08014}{\tt arXiv:2303.08014}.
\bibitem[{{Castillo-Eslava} et~al.(2023){Castillo-Eslava}, Mougan, {Romero-Reche} and Staab}]{castillo-eslava2023RoleLargeLanguage}
\bibinfo{author}{{Castillo-Eslava}, F.}, \bibinfo{author}{Mougan, C.}, \bibinfo{author}{{Romero-Reche}, A.}, \bibinfo{author}{Staab, S.}, \bibinfo{year}{2023}.
\newblock \bibinfo{title}{The {{Role}} of {{Large Language Models}} in the {{Recognition}} of {{Territorial Sovereignty}}: {{An Analysis}} of the {{Construction}} of {{Legitimacy}}}.
\newblock \DOIprefix\doi{10.48550/arXiv.2304.06030}, \href{http://arxiv.org/abs/2304.06030}{\tt arXiv:2304.06030}.
\bibitem[{Chakrabarty et~al.(2022)Chakrabarty, Saakyan, Ghosh and Muresan}]{Chakrabarty2022FLUTEFL}
\bibinfo{author}{Chakrabarty, T.}, \bibinfo{author}{Saakyan, A.}, \bibinfo{author}{Ghosh, D.}, \bibinfo{author}{Muresan, S.}, \bibinfo{year}{2022}.
\newblock \bibinfo{title}{Flute: Figurative language understanding through textual explanations}, in: \bibinfo{booktitle}{Conference on Empirical Methods in Natural Language Processing}.
\bibitem[{Chen(2023a)}]{changfen2023gen}
\bibinfo{author}{Chen, C.}, \bibinfo{year}{2023}a.
\newblock \bibinfo{title}{Generative intelligence and news communication: practical empowerment, conceptual challenge and role reshaping}.
\newblock \bibinfo{journal}{Press Circles} .
\bibitem[{Chen(2023b)}]{chen2023chatgpt}
\bibinfo{author}{Chen, T.J.}, \bibinfo{year}{2023}b.
\newblock \bibinfo{title}{Chatgpt and other artificial intelligence applications speed up scientific writing}.
\newblock \bibinfo{journal}{Journal of the Chinese Medical Association} \bibinfo{volume}{86}, \bibinfo{pages}{351--353}.
\bibitem[{Chen et~al.(2023)Chen, Andiappan, Jenkin and Ovchinnikov}]{chen2023ManagerAIWalk}
\bibinfo{author}{Chen, Y.}, \bibinfo{author}{Andiappan, M.}, \bibinfo{author}{Jenkin, T.}, \bibinfo{author}{Ovchinnikov, A.}, \bibinfo{year}{2023}.
\newblock \bibinfo{title}{A {{Manager}} and an {{AI Walk}} into a {{Bar}}: {{Does ChatGPT Make Biased Decisions Like We Do}}?}
\newblock \DOIprefix\doi{10.2139/ssrn.4380365}.
\bibitem[{Chen et~al.(2022)Chen, Li, Smiley, Ma, Shah and Wang}]{chen-etal-2022-convfinqa}
\bibinfo{author}{Chen, Z.}, \bibinfo{author}{Li, S.}, \bibinfo{author}{Smiley, C.}, \bibinfo{author}{Ma, Z.}, \bibinfo{author}{Shah, S.}, \bibinfo{author}{Wang, W.Y.}, \bibinfo{year}{2022}.
\newblock \bibinfo{title}{{C}onv{F}in{QA}: Exploring the chain of numerical reasoning in conversational finance question answering}, in: \bibinfo{booktitle}{Proceedings of the 2022 Conference on Empirical Methods in Natural Language Processing}, \bibinfo{publisher}{Association for Computational Linguistics}, \bibinfo{address}{Abu Dhabi, United Arab Emirates}. pp. \bibinfo{pages}{6279--6292}.
\newblock \URLprefix \url{https://aclanthology.org/2022.emnlp-main.421}, \DOIprefix\doi{10.18653/v1/2022.emnlp-main.421}.
\bibitem[{Chowdhery et~al.(2022)Chowdhery, Narang, Devlin, Bosma, Mishra, Roberts, Barham, Chung, Sutton, Gehrmann et~al.}]{chowdhery2022PaLMScalingLanguage}
\bibinfo{author}{Chowdhery, A.}, \bibinfo{author}{Narang, S.}, \bibinfo{author}{Devlin, J.}, \bibinfo{author}{Bosma, M.}, \bibinfo{author}{Mishra, G.}, \bibinfo{author}{Roberts, A.}, \bibinfo{author}{Barham, P.}, \bibinfo{author}{Chung, H.W.}, \bibinfo{author}{Sutton, C.}, \bibinfo{author}{Gehrmann, S.}, et~al., \bibinfo{year}{2022}.
\newblock \bibinfo{title}{{{PaLM}}: {{Scaling Language Modeling}} with {{Pathways}}}.
\newblock \DOIprefix\doi{10.48550/arXiv.2204.02311}, \href{http://arxiv.org/abs/2204.02311}{\tt arXiv:2204.02311}.
\bibitem[{Chu et~al.(2023)Chu, Andreas, Ansolabehere and Roy}]{chu2023language}
\bibinfo{author}{Chu, E.}, \bibinfo{author}{Andreas, J.}, \bibinfo{author}{Ansolabehere, S.}, \bibinfo{author}{Roy, D.}, \bibinfo{year}{2023}.
\newblock \bibinfo{title}{Language models trained on media diets can predict public opinion}.
\newblock \bibinfo{journal}{arXiv preprint arXiv:2303.16779} .
\bibitem[{Collins et~al.(2022)Collins, Wong, Feng, Wei and Tenenbaum}]{collins2022StructuredFlexibleRobust}
\bibinfo{author}{Collins, K.M.}, \bibinfo{author}{Wong, C.}, \bibinfo{author}{Feng, J.}, \bibinfo{author}{Wei, M.}, \bibinfo{author}{Tenenbaum, J.B.}, \bibinfo{year}{2022}.
\newblock \bibinfo{title}{Structured, flexible, and robust: Benchmarking and improving large language models towards more human-like behavior in out-of-distribution reasoning tasks}.
\newblock \DOIprefix\doi{10.48550/arXiv.2205.05718}, \href{http://arxiv.org/abs/2205.05718}{\tt arXiv:2205.05718}.
\bibitem[{Coppersmith et~al.(2015)Coppersmith, Dredze, Harman et~al.}]{coppersmith-etal-2015-clpsych}
\bibinfo{author}{Coppersmith, G.}, \bibinfo{author}{Dredze, M.}, \bibinfo{author}{Harman, C.}, et~al., \bibinfo{year}{2015}.
\newblock \bibinfo{title}{{CLP}sych 2015 shared task: Depression and {PTSD} on {T}witter}, in: \bibinfo{booktitle}{Proceedings of the 2nd Workshop on Computational Linguistics and Clinical Psychology: From Linguistic Signal to Clinical Reality}, \bibinfo{publisher}{Association for Computational Linguistics}, \bibinfo{address}{Denver, Colorado}. pp. \bibinfo{pages}{31--39}.
\newblock \URLprefix \url{https://aclanthology.org/W15-1204}, \DOIprefix\doi{10.3115/v1/W15-1204}.
\bibitem[{Cribben and Zeinali(2023)}]{cribben2023BenefitsLimitationsChatGPT}
\bibinfo{author}{Cribben, I.}, \bibinfo{author}{Zeinali, Y.}, \bibinfo{year}{2023}.
\newblock \bibinfo{title}{The benefits and limitations of chatgpt in business education and research: A focus on management science, operations management and data analytics}.
\newblock \bibinfo{journal}{Operations Management and Data Analytics (March 29, 2023)} .
\bibitem[{Dahmen et~al.(2023)Dahmen, Kayaalp, Ollivier, Pareek, Hirschmann, Karlsson and Winkler}]{dahmen2023artificial}
\bibinfo{author}{Dahmen, J.}, \bibinfo{author}{Kayaalp, M.E.}, \bibinfo{author}{Ollivier, M.}, \bibinfo{author}{Pareek, A.}, \bibinfo{author}{Hirschmann, M.T.}, \bibinfo{author}{Karlsson, J.}, \bibinfo{author}{Winkler, P.W.}, \bibinfo{year}{2023}.
\newblock \bibinfo{title}{Artificial intelligence bot chatgpt in medical research: the potential game changer as a double-edged sword}.
\newblock \bibinfo{journal}{Knee Surgery, Sports Traumatology, Arthroscopy} \bibinfo{volume}{31}, \bibinfo{pages}{1187--1189}.
\bibitem[{Danling(2019)}]{GeneralPsychology}
\bibinfo{author}{Danling, P.}, \bibinfo{year}{2019}.
\newblock \bibinfo{title}{General Psychology}.
\newblock \bibinfo{edition}{5} ed., \bibinfo{publisher}{Beijing Normal University Publishing Group}.
\bibitem[{Dasgupta et~al.(2022)Dasgupta, Lampinen, Chan, Creswell, Kumaran, McClelland and Hill}]{dasgupta2022LanguageModelsShow}
\bibinfo{author}{Dasgupta, I.}, \bibinfo{author}{Lampinen, A.K.}, \bibinfo{author}{Chan, S.C.Y.}, \bibinfo{author}{Creswell, A.}, \bibinfo{author}{Kumaran, D.}, \bibinfo{author}{McClelland, J.L.}, \bibinfo{author}{Hill, F.}, \bibinfo{year}{2022}.
\newblock \bibinfo{title}{Language models show human-like content effects on reasoning}.
\newblock \DOIprefix\doi{10.48550/arXiv.2207.07051}, \href{http://arxiv.org/abs/2207.07051}{\tt arXiv:2207.07051}.
\bibitem[{Dergaa et~al.(2023)Dergaa, Chamari, Zmijewski and Saad}]{dergaa2023human}
\bibinfo{author}{Dergaa, I.}, \bibinfo{author}{Chamari, K.}, \bibinfo{author}{Zmijewski, P.}, \bibinfo{author}{Saad, H.B.}, \bibinfo{year}{2023}.
\newblock \bibinfo{title}{From human writing to artificial intelligence generated text: examining the prospects and potential threats of chatgpt in academic writing}.
\newblock \bibinfo{journal}{Biology of Sport} \bibinfo{volume}{40}, \bibinfo{pages}{615--622}.
\bibitem[{Dhingra et~al.(2023)Dhingra, Singh, SB, Malviya and Gill}]{dhingra2023MindMeetsMachine}
\bibinfo{author}{Dhingra, S.}, \bibinfo{author}{Singh, M.}, \bibinfo{author}{SB, V.}, \bibinfo{author}{Malviya, N.}, \bibinfo{author}{Gill, S.S.}, \bibinfo{year}{2023}.
\newblock \bibinfo{title}{Mind meets machine: {{Unravelling GPT-4}}'s cognitive psychology}.
\newblock \DOIprefix\doi{10.48550/arXiv.2303.11436}, \href{http://arxiv.org/abs/2303.11436}{\tt arXiv:2303.11436}.
\bibitem[{Diamond(2023)}]{diamond2023GenlangsZipfLaw}
\bibinfo{author}{Diamond, J.}, \bibinfo{year}{2023}.
\newblock \bibinfo{title}{"{{Genlangs}}" and {{Zipf}}'s {{Law}}: {{Do}} languages generated by {{ChatGPT}} statistically look human?}
\newblock \DOIprefix\doi{10.48550/arXiv.2304.12191}, \href{http://arxiv.org/abs/2304.12191}{\tt arXiv:2304.12191}.
\bibitem[{Dillion et~al.(2023)Dillion, Tandon, Gu and Gray}]{dillion2023can}
\bibinfo{author}{Dillion, D.}, \bibinfo{author}{Tandon, N.}, \bibinfo{author}{Gu, Y.}, \bibinfo{author}{Gray, K.}, \bibinfo{year}{2023}.
\newblock \bibinfo{title}{Can ai language models replace human participants?}
\newblock \bibinfo{journal}{Trends in Cognitive Sciences} .
\bibitem[{Donovan and Hoover(2013)}]{donovan2013elements}
\bibinfo{author}{Donovan, T.}, \bibinfo{author}{Hoover, K.R.}, \bibinfo{year}{2013}.
\newblock \bibinfo{title}{The elements of social scientific thinking}.
\newblock \bibinfo{publisher}{Cengage Learning}.
\bibitem[{Dou(2023)}]{dou2023ExploringGPT3Model}
\bibinfo{author}{Dou, Z.}, \bibinfo{year}{2023}.
\newblock \bibinfo{title}{Exploring {{GPT-3 Model}}'s {{Capability}} in {{Passing}} the {{Sally-Anne Test A Preliminary Study}} in {{Two Languages}}}.
\newblock \DOIprefix\doi{10.31219/osf.io/8r3ma}.
\bibitem[{Easton(1955)}]{easton1955PoliticalSystemInquiry}
\bibinfo{author}{Easton, D.}, \bibinfo{year}{1955}.
\newblock \bibinfo{title}{The {{Political System}}: {{An Inquiry Into}} the {{State}} of {{Political Science}}}.
\newblock \bibinfo{journal}{Ethics} \bibinfo{volume}{65}, \bibinfo{pages}{201--205}.
\newblock \DOIprefix\doi{10.1086/291002}.
\bibitem[{Elsherief et~al.(2021)Elsherief, Ziems, Muchlinski et~al.}]{Elsherief2021LatentHA}
\bibinfo{author}{Elsherief, M.}, \bibinfo{author}{Ziems, C.}, \bibinfo{author}{Muchlinski, D.A.}, et~al., \bibinfo{year}{2021}.
\newblock \bibinfo{title}{Latent hatred: A benchmark for understanding implicit hate speech}, in: \bibinfo{booktitle}{Conference on Empirical Methods in Natural Language Processing}.
\newblock \URLprefix \url{https://api.semanticscholar.org/CorpusID:237490428}.
\bibitem[{Evans and Rzhetsky(2010)}]{evans2010machine}
\bibinfo{author}{Evans, J.}, \bibinfo{author}{Rzhetsky, A.}, \bibinfo{year}{2010}.
\newblock \bibinfo{title}{Machine science}.
\newblock \bibinfo{journal}{Science} \bibinfo{volume}{329}, \bibinfo{pages}{399--400}.
\bibitem[{Farmer and Demers(2010)}]{farmer2010LinguisticsWorkbookCompanion}
\bibinfo{author}{Farmer, A.K.}, \bibinfo{author}{Demers, R.A.}, \bibinfo{year}{2010}.
\newblock \bibinfo{title}{A {{Linguistics Workbook}}: {{Companion}} to {{Linguistics}}, {{Sixth Edition}}}.
\newblock \bibinfo{publisher}{{MIT Press}}.
\bibitem[{Feng et~al.(2023a)Feng, Park, Liu and Tsvetkov}]{feng2023PretrainingDataLanguage}
\bibinfo{author}{Feng, S.}, \bibinfo{author}{Park, C.Y.}, \bibinfo{author}{Liu, Y.}, \bibinfo{author}{Tsvetkov, Y.}, \bibinfo{year}{2023}a.
\newblock \bibinfo{title}{From {{Pretraining Data}} to {{Language Models}} to {{Downstream Tasks}}: {{Tracking}} the {{Trails}} of {{Political Biases Leading}} to {{Unfair NLP Models}}}.
\newblock \DOIprefix\doi{10.48550/arXiv.2305.08283}, \href{http://arxiv.org/abs/2305.08283}{\tt arXiv:2305.08283}.
\bibitem[{Feng et~al.(2023b)Feng, Xu, Li and Liu}]{feng2023BodySizeMetric}
\bibinfo{author}{Feng, X.}, \bibinfo{author}{Xu, S.}, \bibinfo{author}{Li, Y.}, \bibinfo{author}{Liu, J.}, \bibinfo{year}{2023}b.
\newblock \bibinfo{title}{Body size as a metric for the affordable world}.
\newblock \DOIprefix\doi{10.1101/2023.03.20.533336}.
\bibitem[{Fischer et~al.(2023)Fischer, {Luczak-Roesch} and Karl}]{fischer2023WhatDoesChatGPT}
\bibinfo{author}{Fischer, R.}, \bibinfo{author}{{Luczak-Roesch}, M.}, \bibinfo{author}{Karl, J.A.}, \bibinfo{year}{2023}.
\newblock \bibinfo{title}{What does {{ChatGPT}} return about human values? {{Exploring}} value bias in {{ChatGPT}} using a descriptive value theory}.
\newblock \DOIprefix\doi{10.48550/arXiv.2304.03612}, \href{http://arxiv.org/abs/2304.03612}{\tt arXiv:2304.03612}.
\bibitem[{Fr{\k a}ckiewicz(2023)}]{Frackiewicz2023-us}
\bibinfo{author}{Fr{\k a}ckiewicz, M.}, \bibinfo{year}{2023}.
\newblock \bibinfo{title}{How {ChatGPT} is transforming the landscape of social network analysis and community building}.
\newblock \bibinfo{howpublished}{\url{https://ts2.space/en/how-chatgpt-is-transforming-the-landscape-of-social-network-analysis-and-community-building/}}.
\newblock \bibinfo{note}{Accessed: 2023-5-18}.
\bibitem[{Frank et~al.(2019)Frank, Wang, Cebrian and Rahwan}]{frank2019evolution}
\bibinfo{author}{Frank, M.R.}, \bibinfo{author}{Wang, D.}, \bibinfo{author}{Cebrian, M.}, \bibinfo{author}{Rahwan, I.}, \bibinfo{year}{2019}.
\newblock \bibinfo{title}{The evolution of citation graphs in artificial intelligence research}.
\newblock \bibinfo{journal}{Nature Machine Intelligence} \bibinfo{volume}{1}, \bibinfo{pages}{79--85}.
\bibitem[{Fu et~al.(2023)Fu, Peng, Khot and Lapata}]{fu2023ImprovingLanguageModel}
\bibinfo{author}{Fu, Y.}, \bibinfo{author}{Peng, H.}, \bibinfo{author}{Khot, T.}, \bibinfo{author}{Lapata, M.}, \bibinfo{year}{2023}.
\newblock \bibinfo{title}{Improving {{Language Model Negotiation}} with {{Self-Play}} and {{In-Context Learning}} from {{AI Feedback}}}.
\newblock \DOIprefix\doi{10.48550/arXiv.2305.10142}, \href{http://arxiv.org/abs/2305.10142}{\tt arXiv:2305.10142}.
\bibitem[{Gabriel et~al.(2022)Gabriel, Hallinan, Sap et~al.}]{gabriel2022misinfo}
\bibinfo{author}{Gabriel, S.}, \bibinfo{author}{Hallinan, S.}, \bibinfo{author}{Sap, M.}, et~al., \bibinfo{year}{2022}.
\newblock \bibinfo{title}{Misinfo reaction frames: Reasoning about readers’ reactions to news headlines}, in: \bibinfo{booktitle}{Proceedings of the 60th Annual Meeting of the Association for Computational Linguistics (Volume 1: Long Papers)}, pp. \bibinfo{pages}{3108--3127}.
\bibitem[{Gatt and Krahmer(2018)}]{gatt2018survey}
\bibinfo{author}{Gatt, A.}, \bibinfo{author}{Krahmer, E.}, \bibinfo{year}{2018}.
\newblock \bibinfo{title}{Survey of the state of the art in natural language generation: Core tasks, applications and evaluation}.
\newblock \bibinfo{journal}{Journal of Artificial Intelligence Research} \bibinfo{volume}{61}, \bibinfo{pages}{65--170}.
\bibitem[{Giddens(2007)}]{Introductionsociology}
\bibinfo{author}{Giddens, A.}, \bibinfo{year}{2007}.
\newblock \bibinfo{title}{Introduction to sociology}.
\newblock \bibinfo{publisher}{New York : W. W. Norton \& Co.}
\bibitem[{Gilardi et~al.(2023)Gilardi, Alizadeh and Kubli}]{gilardi2023chatgpt}
\bibinfo{author}{Gilardi, F.}, \bibinfo{author}{Alizadeh, M.}, \bibinfo{author}{Kubli, M.}, \bibinfo{year}{2023}.
\newblock \bibinfo{title}{Chatgpt outperforms crowd-workers for text-annotation tasks}.
\newblock \href{http://arxiv.org/abs/2303.15056}{\tt arXiv:2303.15056}.
\bibitem[{Goertzel(2014)}]{goertzel2014artificial}
\bibinfo{author}{Goertzel, B.}, \bibinfo{year}{2014}.
\newblock \bibinfo{title}{Artificial general intelligence: concept, state of the art, and future prospects}.
\newblock \bibinfo{journal}{Journal of Artificial General Intelligence} \bibinfo{volume}{5}, \bibinfo{pages}{1}.
\bibitem[{Goli and Singh(2023)}]{goli2023LanguageTimePreferences}
\bibinfo{author}{Goli, A.}, \bibinfo{author}{Singh, A.}, \bibinfo{year}{2023}.
\newblock \bibinfo{title}{Language, time preferences, and consumer behavior: Evidence from large language models}.
\newblock \bibinfo{journal}{Time Preferences, and Consumer Behavior: Evidence from Large Language Models (May 4, 2023)} .
\bibitem[{Gover(2023)}]{gover2023PoliticalBiasLarge}
\bibinfo{author}{Gover, L.}, \bibinfo{year}{2023}.
\newblock \bibinfo{title}{Political bias in large language models}.
\newblock \bibinfo{journal}{The Commons: Puget Sound Journal of Politics} \bibinfo{volume}{4}, \bibinfo{pages}{2}.
\bibitem[{Griffin et~al.(2023)Griffin, Kleinberg, Mozes, Mai, Vau, Caldwell and {Marvor-Parker}}]{griffin2023SusceptibilityInfluenceLarge}
\bibinfo{author}{Griffin, L.D.}, \bibinfo{author}{Kleinberg, B.}, \bibinfo{author}{Mozes, M.}, \bibinfo{author}{Mai, K.T.}, \bibinfo{author}{Vau, M.}, \bibinfo{author}{Caldwell, M.}, \bibinfo{author}{{Marvor-Parker}, A.}, \bibinfo{year}{2023}.
\newblock \bibinfo{title}{Susceptibility to {{Influence}} of {{Large Language Models}}}.
\newblock \DOIprefix\doi{10.48550/arXiv.2303.06074}, \href{http://arxiv.org/abs/2303.06074}{\tt arXiv:2303.06074}.
\bibitem[{Gu et~al.(2023)Gu, Schreyer, Moffitt and Vasarhelyi}]{gu2023ArtificialIntelligenceCoPiloted}
\bibinfo{author}{Gu, H.}, \bibinfo{author}{Schreyer, M.}, \bibinfo{author}{Moffitt, K.}, \bibinfo{author}{Vasarhelyi, M.A.}, \bibinfo{year}{2023}.
\newblock \bibinfo{title}{Artificial {{Intelligence Co-Piloted Auditing}}}.
\newblock \bibinfo{journal}{SSRN Electronic Journal} \DOIprefix\doi{10.2139/ssrn.4444763}.
\bibitem[{Guo(2023)}]{guo2023GPTAgentsGame}
\bibinfo{author}{Guo, F.}, \bibinfo{year}{2023}.
\newblock \bibinfo{title}{{{GPT Agents}} in {{Game Theory Experiments}}}.
\newblock \DOIprefix\doi{10.48550/arXiv.2305.05516}, \href{http://arxiv.org/abs/2305.05516}{\tt arXiv:2305.05516}.
\bibitem[{Hagendorff(2023)}]{hagendorff2023MachinePsychologyInvestigating}
\bibinfo{author}{Hagendorff, T.}, \bibinfo{year}{2023}.
\newblock \bibinfo{title}{Machine {{Psychology}}: {{Investigating Emergent Capabilities}} and {{Behavior}} in {{Large Language Models Using Psychological Methods}}}.
\newblock \DOIprefix\doi{10.48550/arXiv.2303.13988}, \href{http://arxiv.org/abs/2303.13988}{\tt arXiv:2303.13988}.
\bibitem[{Hagendorff et~al.(2022)Hagendorff, Fabi and Kosinski}]{hagendorff2022MachineIntuitionUncovering}
\bibinfo{author}{Hagendorff, T.}, \bibinfo{author}{Fabi, S.}, \bibinfo{author}{Kosinski, M.}, \bibinfo{year}{2022}.
\newblock \bibinfo{title}{Machine intuition: {{Uncovering}} human-like intuitive decision-making in {{GPT-3}}.5}.
\newblock \DOIprefix\doi{10.48550/arXiv.2212.05206}, \href{http://arxiv.org/abs/2212.05206}{\tt arXiv:2212.05206}.
\bibitem[{Halliday(2006)}]{halliday2006LanguageLinguistics}
\bibinfo{author}{Halliday, M.A.K.}, \bibinfo{year}{2006}.
\newblock \bibinfo{title}{On {{Language}} and {{Linguistics}}}.
\newblock \bibinfo{publisher}{{A\&C Black}}.
\bibitem[{Haman and {\v{S}}koln{\'\i}k(2023)}]{haman2023using}
\bibinfo{author}{Haman, M.}, \bibinfo{author}{{\v{S}}koln{\'\i}k, M.}, \bibinfo{year}{2023}.
\newblock \bibinfo{title}{Using chatgpt to conduct a literature review}.
\newblock \bibinfo{journal}{Accountability in Research} , \bibinfo{pages}{1--3}.
\bibitem[{Han et~al.(2022)Han, Ransom, Perfors and Kemp}]{han2022HumanlikePropertyInduction}
\bibinfo{author}{Han, S.J.}, \bibinfo{author}{Ransom, K.J.}, \bibinfo{author}{Perfors, A.}, \bibinfo{author}{Kemp, C.}, \bibinfo{year}{2022}.
\newblock \bibinfo{title}{Human-like property induction is a challenge for large language models}.
\newblock \bibinfo{journal}{Proceedings of the Annual Meeting of the Cognitive Science Society} \bibinfo{volume}{44}.
\newblock \URLprefix \url{https://escholarship.org/uc/item/3w84q1s1}.
\bibitem[{Hartmann et~al.(2023)Hartmann, Schwenzow and Witte}]{hartmann2023PoliticalIdeologyConversational}
\bibinfo{author}{Hartmann, J.}, \bibinfo{author}{Schwenzow, J.}, \bibinfo{author}{Witte, M.}, \bibinfo{year}{2023}.
\newblock \bibinfo{title}{The political ideology of conversational {{AI}}: {{Converging}} evidence on {{ChatGPT}}'s pro-environmental, left-libertarian orientation}.
\newblock \DOIprefix\doi{10.48550/arXiv.2301.01768}, \href{http://arxiv.org/abs/2301.01768}{\tt arXiv:2301.01768}.
\bibitem[{Hoes et~al.(2023)Hoes, Altay and Bermeo}]{hoes2023using}
\bibinfo{author}{Hoes, E.}, \bibinfo{author}{Altay, S.}, \bibinfo{author}{Bermeo, J.}, \bibinfo{year}{2023}.
\newblock \bibinfo{title}{Using chatgpt to fight misinformation: Chatgpt nails 72\% of 12,000 verified claims} .
\bibitem[{Horton(2023)}]{horton2023LargeLanguageModels}
\bibinfo{author}{Horton, J.J.}, \bibinfo{year}{2023}.
\newblock \bibinfo{title}{Large language models as simulated economic agents: What can we learn from homo silicus?} .
\bibitem[{Huang et~al.(2023a)Huang, Kwak and An}]{10.1145/3543873.3587320}
\bibinfo{author}{Huang, F.}, \bibinfo{author}{Kwak, H.}, \bibinfo{author}{An, J.}, \bibinfo{year}{2023}a.
\newblock \bibinfo{title}{Chain of explanation: New prompting method to generate quality natural language explanation for implicit hate speech}, in: \bibinfo{booktitle}{Companion Proceedings of the ACM Web Conference 2023}, \bibinfo{publisher}{Association for Computing Machinery}, \bibinfo{address}{New York, NY, USA}. p. \bibinfo{pages}{90–93}.
\newblock \URLprefix \url{https://doi.org/10.1145/3543873.3587320}, \DOIprefix\doi{10.1145/3543873.3587320}.
\bibitem[{Huang et~al.(2023b)Huang, Kwak and An}]{Huang_2023}
\bibinfo{author}{Huang, F.}, \bibinfo{author}{Kwak, H.}, \bibinfo{author}{An, J.}, \bibinfo{year}{2023}b.
\newblock \bibinfo{title}{Is {ChatGPT} better than human annotators? potential and limitations of {ChatGPT} in explaining implicit hate speech}, in: \bibinfo{booktitle}{Companion Proceedings of the {ACM} Web Conference 2023}, \bibinfo{publisher}{{ACM}}.
\newblock \URLprefix \url{https://doi.org/10.1145\%2F3543873.3587368}, \DOIprefix\doi{10.1145/3543873.3587368}.
\bibitem[{Irving et~al.(2018)Irving, Christiano and Amodei}]{irving2018ai}
\bibinfo{author}{Irving, G.}, \bibinfo{author}{Christiano, P.}, \bibinfo{author}{Amodei, D.}, \bibinfo{year}{2018}.
\newblock \bibinfo{title}{Ai safety via debate}.
\newblock \bibinfo{journal}{arXiv preprint arXiv:1805.00899} .
\bibitem[{Iyyer et~al.(2014)Iyyer, Enns, Boyd-Graber et~al.}]{iyyer-etal-2014-political}
\bibinfo{author}{Iyyer, M.}, \bibinfo{author}{Enns, P.}, \bibinfo{author}{Boyd-Graber, J.}, et~al., \bibinfo{year}{2014}.
\newblock \bibinfo{title}{Political ideology detection using recursive neural networks}, in: \bibinfo{booktitle}{Proceedings of the 52nd Annual Meeting of the Association for Computational Linguistics (Volume 1: Long Papers)}, \bibinfo{publisher}{Association for Computational Linguistics}, \bibinfo{address}{Baltimore, Maryland}. pp. \bibinfo{pages}{1113--1122}.
\newblock \URLprefix \url{https://aclanthology.org/P14-1105}, \DOIprefix\doi{10.3115/v1/P14-1105}.
\bibitem[{Jaccard and Jacoby(2019)}]{jaccard2019theory}
\bibinfo{author}{Jaccard, J.}, \bibinfo{author}{Jacoby, J.}, \bibinfo{year}{2019}.
\newblock \bibinfo{title}{Theory construction and model-building skills: A practical guide for social scientists}.
\newblock \bibinfo{publisher}{Guilford publications}.
\bibitem[{Jha et~al.(2019)Jha, Xun, Wang and Zhang}]{jha2019hypothesis}
\bibinfo{author}{Jha, K.}, \bibinfo{author}{Xun, G.}, \bibinfo{author}{Wang, Y.}, \bibinfo{author}{Zhang, A.}, \bibinfo{year}{2019}.
\newblock \bibinfo{title}{Hypothesis generation from text based on co-evolution of biomedical concepts}, in: \bibinfo{booktitle}{Proceedings of the 25th ACM SIGKDD International Conference on Knowledge Discovery \& Data Mining}, pp. \bibinfo{pages}{843--851}.
\bibitem[{Jiang et~al.(2022a)Jiang, Xu, Zhu, Han, Zhang and Zhu}]{jiang2022MPIEvaluatingInducing}
\bibinfo{author}{Jiang, G.}, \bibinfo{author}{Xu, M.}, \bibinfo{author}{Zhu, S.C.}, \bibinfo{author}{Han, W.}, \bibinfo{author}{Zhang, C.}, \bibinfo{author}{Zhu, Y.}, \bibinfo{year}{2022}a.
\newblock \bibinfo{title}{{{MPI}}: {{Evaluating}} and {{Inducing Personality}} in {{Pre-trained Language Models}}}.
\newblock \DOIprefix\doi{10.48550/arXiv.2206.07550}, \href{http://arxiv.org/abs/2206.07550}{\tt arXiv:2206.07550}.
\bibitem[{Jiang et~al.(2022b)Jiang, Beeferman, Roy and Roy}]{jiang2022CommunityLMProbingPartisan}
\bibinfo{author}{Jiang, H.}, \bibinfo{author}{Beeferman, D.}, \bibinfo{author}{Roy, B.}, \bibinfo{author}{Roy, D.}, \bibinfo{year}{2022}b.
\newblock \bibinfo{title}{Communitylm: Probing partisan worldviews from language models}, in: \bibinfo{booktitle}{Proceedings of the 29th International Conference on Computational Linguistics}, pp. \bibinfo{pages}{6818--6826}.
\bibitem[{Jiang et~al.(2023)Jiang, Zhang, Cao, Kabbara and Roy}]{jiang2023PersonaLLMInvestigatingAbility}
\bibinfo{author}{Jiang, H.}, \bibinfo{author}{Zhang, X.}, \bibinfo{author}{Cao, X.}, \bibinfo{author}{Kabbara, J.}, \bibinfo{author}{Roy, D.}, \bibinfo{year}{2023}.
\newblock \bibinfo{title}{{{PersonaLLM}}: {{Investigating}} the {{Ability}} of {{GPT-3}}.5 to {{Express Personality Traits}} and {{Gender Differences}}}.
\newblock \DOIprefix\doi{10.48550/arXiv.2305.02547}, \href{http://arxiv.org/abs/2305.02547}{\tt arXiv:2305.02547}.
\bibitem[{Jin et~al.(2022)Jin, Levine et~al.}]{jin2022WhenMakeExceptions}
\bibinfo{author}{Jin, Z.}, \bibinfo{author}{Levine, S.}, et~al., \bibinfo{year}{2022}.
\newblock \bibinfo{title}{When to {{Make Exceptions}}: {{Exploring Language Models}} as {{Accounts}} of {{Human Moral Judgment}}}.
\newblock \bibinfo{journal}{Advances in Neural Information Processing Systems} \bibinfo{volume}{35}, \bibinfo{pages}{28458--28473}.
\newblock \URLprefix \url{https://proceedings.neurips.cc/paper_files/paper/2022/hash/b654d6150630a5ba5df7a55621390daf-Abstract-Conference.html}.
\bibitem[{Johnson et~al.(2023)Johnson, King, Warner, Aneja, Kann and Bylund}]{johnson_using_2023}
\bibinfo{author}{Johnson, S.B.}, \bibinfo{author}{King, A.J.}, \bibinfo{author}{Warner, E.L.}, \bibinfo{author}{Aneja, S.}, \bibinfo{author}{Kann, B.H.}, \bibinfo{author}{Bylund, C.L.}, \bibinfo{year}{2023}.
\newblock \bibinfo{title}{Using {ChatGPT} to evaluate cancer myths and misconceptions: artificial intelligence and cancer information}.
\newblock \bibinfo{journal}{JNCI Cancer Spectrum} \bibinfo{volume}{7}, \bibinfo{pages}{pkad015}.
\newblock \URLprefix \url{https://doi.org/10.1093/jncics/pkad015}, \DOIprefix\doi{10.1093/jncics/pkad015}.
\bibitem[{Johnson and Obradovich(2023)}]{johnson2023EvidenceBehaviorConsistent}
\bibinfo{author}{Johnson, T.}, \bibinfo{author}{Obradovich, N.}, \bibinfo{year}{2023}.
\newblock \bibinfo{title}{Evidence of behavior consistent with self-interest and altruism in an artificially intelligent agent}.
\newblock \DOIprefix\doi{10.48550/arXiv.2301.02330}, \href{http://arxiv.org/abs/2301.02330}{\tt arXiv:2301.02330}.
\bibitem[{Jones and Steinhardt(2022)}]{jones2022CapturingFailuresLarge}
\bibinfo{author}{Jones, E.}, \bibinfo{author}{Steinhardt, J.}, \bibinfo{year}{2022}.
\newblock \bibinfo{title}{Capturing {{Failures}} of {{Large Language Models}} via {{Human Cognitive Biases}}}.
\newblock \bibinfo{journal}{Advances in Neural Information Processing Systems} \bibinfo{volume}{35}, \bibinfo{pages}{11785--11799}.
\newblock \URLprefix \url{https://proceedings.neurips.cc/paper_files/paper/2022/hash/4d13b2d99519c5415661dad44ab7edcd-Abstract-Conference.html}.
\bibitem[{Joyce et~al.(2021)Joyce, {Smith-Doerr}, Alegria, Bell, Cruz, Hoffman, Noble and Shestakofsky}]{joyce2021SociologyArtificialIntelligence}
\bibinfo{author}{Joyce, K.}, \bibinfo{author}{{Smith-Doerr}, L.}, \bibinfo{author}{Alegria, S.}, \bibinfo{author}{Bell, S.}, \bibinfo{author}{Cruz, T.}, \bibinfo{author}{Hoffman, S.G.}, \bibinfo{author}{Noble, S.U.}, \bibinfo{author}{Shestakofsky, B.}, \bibinfo{year}{2021}.
\newblock \bibinfo{title}{Toward a {{Sociology}} of {{Artificial Intelligence}}: {{A Call}} for {{Research}} on {{Inequalities}} and {{Structural Change}}}.
\newblock \bibinfo{journal}{Socius} \bibinfo{volume}{7}, \bibinfo{pages}{2378023121999581}.
\newblock \DOIprefix\doi{10.1177/2378023121999581}.
\bibitem[{Jungwirth and Haluza(2023)}]{jungwirth2023forecasting}
\bibinfo{author}{Jungwirth, D.}, \bibinfo{author}{Haluza, D.}, \bibinfo{year}{2023}.
\newblock \bibinfo{title}{Forecasting geopolitical conflicts using gpt-3 ai: Reali-ty-check one year into the 2022 ukraine war} .
\bibitem[{Juren~Lin(2017)}]{juren2017}
\bibinfo{author}{Juren~Lin, Y.L.}, \bibinfo{year}{2017}.
\newblock \bibinfo{title}{Social science research methods}.
\newblock \bibinfo{publisher}{Shandong People's Publishing House}.
\bibitem[{Kalinin(2023)}]{kalinin2023geopolitical}
\bibinfo{author}{Kalinin, K.}, \bibinfo{year}{2023}.
\newblock \bibinfo{title}{Geopolitical forecasting analysis of the russia-ukraine war using the expert's survey, predictioneer's game and gpt-3}.
\newblock \bibinfo{journal}{Predictioneer's Game and GPT-3 (April 8, 2023)} .
\bibitem[{Karra et~al.(2023)Karra, Nguyen and Tulabandhula}]{karra2023EstimatingPersonalityWhiteBox}
\bibinfo{author}{Karra, S.R.}, \bibinfo{author}{Nguyen, S.T.}, \bibinfo{author}{Tulabandhula, T.}, \bibinfo{year}{2023}.
\newblock \bibinfo{title}{Estimating the {{Personality}} of {{White-Box Language Models}}}.
\newblock \DOIprefix\doi{10.48550/arXiv.2204.12000}, \href{http://arxiv.org/abs/2204.12000}{\tt arXiv:2204.12000}.
\bibitem[{Kieval(1997)}]{kieval1997pursuing}
\bibinfo{author}{Kieval, H.J.}, \bibinfo{year}{1997}.
\newblock \bibinfo{title}{Pursuing the golem of prague: Jewish culture and the invention of a tradition}.
\newblock \bibinfo{journal}{Modern Judaism} \bibinfo{volume}{17}, \bibinfo{pages}{1--23}.
\bibitem[{King(2023)}]{king2023GPT4AlignsNew}
\bibinfo{author}{King, M.}, \bibinfo{year}{2023}.
\newblock \bibinfo{title}{{{GPT-4}} aligns with the {{New Liberal Party}}, while other large language models refuse to answer political questions}.
\newblock \DOIprefix\doi{10.31224/2974}.
\bibitem[{Kjell et~al.(2023)Kjell, Kjell and Schwartz}]{kjell2023ai}
\bibinfo{author}{Kjell, O.}, \bibinfo{author}{Kjell, K.}, \bibinfo{author}{Schwartz, H.A.}, \bibinfo{year}{2023}.
\newblock \bibinfo{title}{Ai-based large language models are ready to transform psychological health assessment} .
\bibitem[{Kjell et~al.(2022)Kjell, Sikstr{\"o}m, Kjell and Schwartz}]{kjell2022natural}
\bibinfo{author}{Kjell, O.N.}, \bibinfo{author}{Sikstr{\"o}m, S.}, \bibinfo{author}{Kjell, K.}, \bibinfo{author}{Schwartz, H.A.}, \bibinfo{year}{2022}.
\newblock \bibinfo{title}{Natural language analyzed with ai-based transformers predict traditional subjective well-being measures approaching the theoretical upper limits in accuracy}.
\newblock \bibinfo{journal}{Scientific reports} \bibinfo{volume}{12}, \bibinfo{pages}{3918}.
\bibitem[{Klein and Kleinman(2002)}]{klein2002social}
\bibinfo{author}{Klein, H.K.}, \bibinfo{author}{Kleinman, D.L.}, \bibinfo{year}{2002}.
\newblock \bibinfo{title}{The social construction of technology: Structural considerations}.
\newblock \bibinfo{journal}{Science, Technology, \& Human Values} \bibinfo{volume}{27}, \bibinfo{pages}{28--52}.
\bibitem[{Kosinski(2023)}]{kosinski2023TheoryMindMay}
\bibinfo{author}{Kosinski, M.}, \bibinfo{year}{2023}.
\newblock \bibinfo{title}{Theory of {{Mind May Have Spontaneously Emerged}} in {{Large Language Models}}}.
\newblock \DOIprefix\doi{10.48550/arXiv.2302.02083}, \href{http://arxiv.org/abs/2302.02083}{\tt arXiv:2302.02083}.
\bibitem[{Kosoy et~al.(2022)Kosoy, Chan, Liu, Collins, Kaufmann, Huang, Hamrick, Canny, Ke and Gopnik}]{kosoy2022UnderstandingHowMachines}
\bibinfo{author}{Kosoy, E.}, \bibinfo{author}{Chan, D.M.}, \bibinfo{author}{Liu, A.}, \bibinfo{author}{Collins, J.}, \bibinfo{author}{Kaufmann, B.}, \bibinfo{author}{Huang, S.H.}, \bibinfo{author}{Hamrick, J.B.}, \bibinfo{author}{Canny, J.}, \bibinfo{author}{Ke, N.R.}, \bibinfo{author}{Gopnik, A.}, \bibinfo{year}{2022}.
\newblock \bibinfo{title}{Towards {{Understanding How Machines Can Learn Causal Overhypotheses}}}.
\newblock \DOIprefix\doi{10.48550/arXiv.2206.08353}, \href{http://arxiv.org/abs/2206.08353}{\tt arXiv:2206.08353}.
\bibitem[{Krenn and Zeilinger(2020)}]{krenn2020predicting}
\bibinfo{author}{Krenn, M.}, \bibinfo{author}{Zeilinger, A.}, \bibinfo{year}{2020}.
\newblock \bibinfo{title}{Predicting research trends with semantic and neural networks with an application in quantum physics}.
\newblock \bibinfo{journal}{Proceedings of the National Academy of Sciences} \bibinfo{volume}{117}, \bibinfo{pages}{1910--1916}.
\bibitem[{Krishna et~al.(2022)Krishna, Lee, Fei-Fei and Bernstein}]{krishna2022socially}
\bibinfo{author}{Krishna, R.}, \bibinfo{author}{Lee, D.}, \bibinfo{author}{Fei-Fei, L.}, \bibinfo{author}{Bernstein, M.S.}, \bibinfo{year}{2022}.
\newblock \bibinfo{title}{Socially situated artificial intelligence enables learning from human interaction}.
\newblock \bibinfo{journal}{Proceedings of the National Academy of Sciences} \bibinfo{volume}{119}, \bibinfo{pages}{e2115730119}.
\bibitem[{Krugman and Wells(2009)}]{krugman2009economics}
\bibinfo{author}{Krugman, P.R.}, \bibinfo{author}{Wells, R.}, \bibinfo{year}{2009}.
\newblock \bibinfo{title}{Economics}.
\newblock \bibinfo{publisher}{Macmillan}.
\bibitem[{Lamichhane(2023)}]{lamichhane2023evaluation}
\bibinfo{author}{Lamichhane, B.}, \bibinfo{year}{2023}.
\newblock \bibinfo{title}{Evaluation of chatgpt for nlp-based mental health applications}.
\newblock \bibinfo{journal}{arXiv preprint arXiv:2303.15727} .
\bibitem[{Leippold(2023)}]{leippold2023ThusSpokeGPT3}
\bibinfo{author}{Leippold, M.}, \bibinfo{year}{2023}.
\newblock \bibinfo{title}{Thus spoke {{GPT-3}}: {{Interviewing}} a large-language model on climate finance}.
\newblock \bibinfo{journal}{Finance Research Letters} \bibinfo{volume}{53}, \bibinfo{pages}{103617}.
\newblock \DOIprefix\doi{10.1016/j.frl.2022.103617}.
\bibitem[{Li et~al.(2022)Li, Knopman, Xu, Cohen and Pakhomov}]{li2022GPTDInducingDementiarelated}
\bibinfo{author}{Li, C.}, \bibinfo{author}{Knopman, D.}, \bibinfo{author}{Xu, W.}, \bibinfo{author}{Cohen, T.}, \bibinfo{author}{Pakhomov, S.}, \bibinfo{year}{2022}.
\newblock \bibinfo{title}{Gpt-d: Inducing dementia-related linguistic anomalies by deliberate degradation of artificial neural language models}, in: \bibinfo{booktitle}{Proceedings of the 60th Annual Meeting of the Association for Computational Linguistics (Volume 1: Long Papers)}, pp. \bibinfo{pages}{1866--1877}.
\bibitem[{Li et~al.(2023)Li, Li, Joty, Liu, Huang, Qiu and Bing}]{li2023DoesGPT3Demonstrate}
\bibinfo{author}{Li, X.}, \bibinfo{author}{Li, Y.}, \bibinfo{author}{Joty, S.}, \bibinfo{author}{Liu, L.}, \bibinfo{author}{Huang, F.}, \bibinfo{author}{Qiu, L.}, \bibinfo{author}{Bing, L.}, \bibinfo{year}{2023}.
\newblock \bibinfo{title}{Does {{GPT-3 Demonstrate Psychopathy}}? {{Evaluating Large Language Models}} from a {{Psychological Perspective}}}.
\newblock \DOIprefix\doi{10.48550/arXiv.2212.10529}, \href{http://arxiv.org/abs/2212.10529}{\tt arXiv:2212.10529}.
\bibitem[{Liu et~al.(2021)Liu, Jia, Wei, Xu, Wang and Vosoughi}]{liu2021MitigatingPoliticalBias}
\bibinfo{author}{Liu, R.}, \bibinfo{author}{Jia, C.}, \bibinfo{author}{Wei, J.}, \bibinfo{author}{Xu, G.}, \bibinfo{author}{Wang, L.}, \bibinfo{author}{Vosoughi, S.}, \bibinfo{year}{2021}.
\newblock \bibinfo{title}{Mitigating {{Political Bias}} in {{Language Models}} through {{Reinforced Calibration}}}.
\newblock \bibinfo{journal}{Proceedings of the AAAI Conference on Artificial Intelligence} \bibinfo{volume}{35}, \bibinfo{pages}{14857--14866}.
\newblock \DOIprefix\doi{10.1609/aaai.v35i17.17744}.
\bibitem[{Long et~al.(2022)Long, Jeff, Xu et~al.}]{ouyang2022InstructGPTTrainingLanguage}
\bibinfo{author}{Long, O.}, \bibinfo{author}{Jeff, W.}, \bibinfo{author}{Xu, J.}, et~al., \bibinfo{year}{2022}.
\newblock \bibinfo{title}{{{InstructGPT--Training}} language models to follow instructions with human feedback}.
\newblock \DOIprefix\doi{10.48550/arXiv.2203.02155}, \href{http://arxiv.org/abs/2203.02155}{\tt arXiv:2203.02155}.
\bibitem[{Lopez-Lira and Tang(2023)}]{Lopez_Lira_2023}
\bibinfo{author}{Lopez-Lira, A.}, \bibinfo{author}{Tang, Y.}, \bibinfo{year}{2023}.
\newblock \bibinfo{title}{Can {ChatGPT} forecast stock price movements? return predictability and large language models}.
\newblock \bibinfo{journal}{{SSRN} Electronic Journal} \URLprefix \url{https://doi.org/10.2139\%2Fssrn.4412788}, \DOIprefix\doi{10.2139/ssrn.4412788}.
\bibitem[{Lu et~al.(2022)Lu, Bartolo, Moore, Riedel and Stenetorp}]{lu2022FantasticallyOrderedPrompts}
\bibinfo{author}{Lu, Y.}, \bibinfo{author}{Bartolo, M.}, \bibinfo{author}{Moore, A.}, \bibinfo{author}{Riedel, S.}, \bibinfo{author}{Stenetorp, P.}, \bibinfo{year}{2022}.
\newblock \bibinfo{title}{Fantastically {{Ordered Prompts}} and {{Where}} to {{Find Them}}: {{Overcoming Few-Shot Prompt Order Sensitivity}}}, in: \bibinfo{booktitle}{Proceedings of the 60th {{Annual Meeting}} of the {{Association}} for {{Computational Linguistics}} ({{Volume}} 1: {{Long Papers}})}, \bibinfo{publisher}{{Association for Computational Linguistics}}, \bibinfo{address}{{Dublin, Ireland}}. pp. \bibinfo{pages}{8086--8098}.
\newblock \DOIprefix\doi{10.18653/v1/2022.acl-long.556}.
\bibitem[{Lucy and Bamman(2021)}]{lucy2021GenderRepresentationBias}
\bibinfo{author}{Lucy, L.}, \bibinfo{author}{Bamman, D.}, \bibinfo{year}{2021}.
\newblock \bibinfo{title}{Gender and {{Representation Bias}} in {{GPT-3 Generated Stories}}}, in: \bibinfo{booktitle}{Proceedings of the {{Third Workshop}} on {{Narrative Understanding}}}, \bibinfo{publisher}{{Association for Computational Linguistics}}, \bibinfo{address}{{Virtual}}. pp. \bibinfo{pages}{48--55}.
\newblock \DOIprefix\doi{10.18653/v1/2021.nuse-1.5}.
\bibitem[{Maia et~al.(2018)Maia, Handschuh, Freitas, Davis, McDermott, Zarrouk and Balahur}]{finqa}
\bibinfo{author}{Maia, M.}, \bibinfo{author}{Handschuh, S.}, \bibinfo{author}{Freitas, A.}, \bibinfo{author}{Davis, B.}, \bibinfo{author}{McDermott, R.}, \bibinfo{author}{Zarrouk, M.}, \bibinfo{author}{Balahur, A.}, \bibinfo{year}{2018}.
\newblock \bibinfo{title}{Www'18 open challenge: Financial opinion mining and question answering}, pp. \bibinfo{pages}{1941--1942}.
\newblock \DOIprefix\doi{10.1145/3184558.3192301}.
\bibitem[{Malo et~al.(2013)Malo, Sinha, Korhonen, Wallenius and Takala}]{Malo2013GoodDO}
\bibinfo{author}{Malo, P.}, \bibinfo{author}{Sinha, A.}, \bibinfo{author}{Korhonen, P.J.}, \bibinfo{author}{Wallenius, J.}, \bibinfo{author}{Takala, P.}, \bibinfo{year}{2013}.
\newblock \bibinfo{title}{Good debt or bad debt: Detecting semantic orientations in economic texts}.
\newblock \bibinfo{journal}{Journal of the Association for Information Science and Technology} \bibinfo{volume}{65}.
\bibitem[{Mauriello et~al.(2021)Mauriello, Lincoln, Hon et~al.}]{Mauriello2021SADAS}
\bibinfo{author}{Mauriello, M.L.}, \bibinfo{author}{Lincoln, E.T.}, \bibinfo{author}{Hon, G.}, et~al., \bibinfo{year}{2021}.
\newblock \bibinfo{title}{Sad: A stress annotated dataset for recognizing everyday stressors in sms-like conversational systems}.
\newblock \bibinfo{journal}{Extended Abstracts of the 2021 CHI Conference on Human Factors in Computing Systems} \URLprefix \url{https://api.semanticscholar.org/CorpusID:233361747}.
\bibitem[{McCorduck and Cfe(2004)}]{mccorduck2004machines}
\bibinfo{author}{McCorduck, P.}, \bibinfo{author}{Cfe, C.}, \bibinfo{year}{2004}.
\newblock \bibinfo{title}{Machines who think: A personal inquiry into the history and prospects of artificial intelligence}.
\newblock \bibinfo{publisher}{CRC Press}.
\bibitem[{McGee(2023a)}]{mcgee2023ChatGptBiased}
\bibinfo{author}{McGee, R.W.}, \bibinfo{year}{2023}a.
\newblock \bibinfo{title}{Is {{Chat Gpt Biased Against Conservatives}}? {{An Empirical Study}}}.
\newblock \DOIprefix\doi{10.2139/ssrn.4359405}.
\bibitem[{McGee(2023b)}]{mcgee2023using}
\bibinfo{author}{McGee, R.W.}, \bibinfo{year}{2023}b.
\newblock \bibinfo{title}{Using chatgpt to conduct literature searches: A case study}.
\newblock \bibinfo{journal}{Journal of Business Ethics} \bibinfo{volume}{95}, \bibinfo{pages}{165--178}.
\bibitem[{Mialon et~al.(2023)Mialon, Dess{\`\i}, Lomeli, Nalmpantis, Pasunuru, Raileanu, Rozi{\`e}re, Schick, Dwivedi-Yu, Celikyilmaz et~al.}]{mialon2023augmented}
\bibinfo{author}{Mialon, G.}, \bibinfo{author}{Dess{\`\i}, R.}, \bibinfo{author}{Lomeli, M.}, \bibinfo{author}{Nalmpantis, C.}, \bibinfo{author}{Pasunuru, R.}, \bibinfo{author}{Raileanu, R.}, \bibinfo{author}{Rozi{\`e}re, B.}, \bibinfo{author}{Schick, T.}, \bibinfo{author}{Dwivedi-Yu, J.}, \bibinfo{author}{Celikyilmaz, A.}, et~al., \bibinfo{year}{2023}.
\newblock \bibinfo{title}{Augmented language models: a survey}.
\newblock \bibinfo{journal}{arXiv preprint arXiv:2302.07842} .
\bibitem[{Miotto et~al.(2022)Miotto, Rossberg and Kleinberg}]{miotto2022WhoGPT3Exploration}
\bibinfo{author}{Miotto, M.}, \bibinfo{author}{Rossberg, N.}, \bibinfo{author}{Kleinberg, B.}, \bibinfo{year}{2022}.
\newblock \bibinfo{title}{Who is {{GPT-3}}? {{An}} exploration of personality, values and demographics}, in: \bibinfo{booktitle}{Proceedings of the {{Fifth Workshop}} on {{Natural Language Processing}} and {{Computational Social Science}} ({{NLP}}+{{CSS}})}, \bibinfo{publisher}{{Association for Computational Linguistics}}, \bibinfo{address}{{Abu Dhabi, UAE}}. pp. \bibinfo{pages}{218--227}.
\newblock \URLprefix \url{https://aclanthology.org/2022.nlpcss-1.24}.
\bibitem[{Misra(2022)}]{misra2022politifact}
\bibinfo{author}{Misra, R.}, \bibinfo{year}{2022}.
\newblock \bibinfo{title}{Politifact fact check dataset}.
\newblock \DOIprefix\doi{10.13140/RG.2.2.29923.22566}.
\bibitem[{Mohammad et~al.(2016)Mohammad, Kiritchenko, Sobhani et~al.}]{mohammad-etal-2016-semeval}
\bibinfo{author}{Mohammad, S.}, \bibinfo{author}{Kiritchenko, S.}, \bibinfo{author}{Sobhani, P.}, et~al., \bibinfo{year}{2016}.
\newblock \bibinfo{title}{{S}em{E}val-2016 task 6: Detecting stance in tweets}, in: \bibinfo{booktitle}{Proceedings of the 10th International Workshop on Semantic Evaluation ({S}em{E}val-2016)}, \bibinfo{publisher}{Association for Computational Linguistics}, \bibinfo{address}{San Diego, California}. pp. \bibinfo{pages}{31--41}.
\newblock \URLprefix \url{https://aclanthology.org/S16-1003}, \DOIprefix\doi{10.18653/v1/S16-1003}.
\bibitem[{Motoki et~al.(2023)Motoki, Neto and Rodrigues}]{motoki2023MoreHumanHuman}
\bibinfo{author}{Motoki, F.}, \bibinfo{author}{Neto, V.P.}, \bibinfo{author}{Rodrigues, V.}, \bibinfo{year}{2023}.
\newblock \bibinfo{title}{More human than human: Measuring chatgpt political bias}.
\newblock \bibinfo{journal}{Public Choice} , \bibinfo{pages}{1--21}.
\bibitem[{Niszczota and Abbas(2023)}]{niszczota2023GPTFinancialAdvisor}
\bibinfo{author}{Niszczota, P.}, \bibinfo{author}{Abbas, S.}, \bibinfo{year}{2023}.
\newblock \bibinfo{title}{Gpt as a financial advisor}.
\newblock \bibinfo{journal}{Available at SSRN 4384861} .
\bibitem[{OpenAI()}]{HowShouldAI}
\bibinfo{author}{OpenAI}, .
\newblock \bibinfo{title}{How should {{AI}} systems behave, and who should decide?}
\newblock \URLprefix \url{https://openai.com/blog/how-should-ai-systems-behave}.
\bibitem[{OpenAI(2023)}]{openai2023GPT4TechnicalReport}
\bibinfo{author}{OpenAI}, \bibinfo{year}{2023}.
\newblock \bibinfo{title}{{{GPT-4 Technical Report}}}.
\newblock \DOIprefix\doi{10.48550/arXiv.2303.08774}, \href{http://arxiv.org/abs/2303.08774}{\tt arXiv:2303.08774}.
\bibitem[{Park et~al.(2023a)Park, O'Brien, Cai, Morris, Liang and Bernstein}]{park2023GenerativeAgentsInteractivea}
\bibinfo{author}{Park, J.S.}, \bibinfo{author}{O'Brien, J.C.}, \bibinfo{author}{Cai, C.J.}, \bibinfo{author}{Morris, M.R.}, \bibinfo{author}{Liang, P.}, \bibinfo{author}{Bernstein, M.S.}, \bibinfo{year}{2023}a.
\newblock \bibinfo{title}{Generative {{Agents}}: {{Interactive Simulacra}} of {{Human Behavior}}}.
\newblock \DOIprefix\doi{10.48550/arXiv.2304.03442}, \href{http://arxiv.org/abs/2304.03442}{\tt arXiv:2304.03442}.
\bibitem[{Park et~al.(2022)Park, Popowski, Cai, Morris, Liang and Bernstein}]{park2022social}
\bibinfo{author}{Park, J.S.}, \bibinfo{author}{Popowski, L.}, \bibinfo{author}{Cai, C.}, \bibinfo{author}{Morris, M.R.}, \bibinfo{author}{Liang, P.}, \bibinfo{author}{Bernstein, M.S.}, \bibinfo{year}{2022}.
\newblock \bibinfo{title}{Social simulacra: Creating populated prototypes for social computing systems}, in: \bibinfo{booktitle}{Proceedings of the 35th Annual ACM Symposium on User Interface Software and Technology}, pp. \bibinfo{pages}{1--18}.
\bibitem[{Park et~al.(2023b)Park, Kaplan, Ren, Hsu, Li, Xu, Li and Li}]{park2023can}
\bibinfo{author}{Park, Y.J.}, \bibinfo{author}{Kaplan, D.}, \bibinfo{author}{Ren, Z.}, \bibinfo{author}{Hsu, C.W.}, \bibinfo{author}{Li, C.}, \bibinfo{author}{Xu, H.}, \bibinfo{author}{Li, S.}, \bibinfo{author}{Li, J.}, \bibinfo{year}{2023}b.
\newblock \bibinfo{title}{Can chatgpt be used to generate scientific hypotheses?}
\newblock \bibinfo{journal}{arXiv preprint arXiv:2304.12208} .
\bibitem[{Pellert et~al.(2022)Pellert, Lechner, Wagner, Rammstedt and Strohmaier}]{pellert2022AIPsychometricsUsing}
\bibinfo{author}{Pellert, M.}, \bibinfo{author}{Lechner, C.}, \bibinfo{author}{Wagner, C.}, \bibinfo{author}{Rammstedt, B.}, \bibinfo{author}{Strohmaier, M.}, \bibinfo{year}{2022}.
\newblock \bibinfo{title}{{{AI Psychometrics}}: {{Using}} psychometric inventories to obtain psychological profiles of large language models}.
\newblock \DOIprefix\doi{10.31234/osf.io/jv5dt}.
\bibitem[{Phelps and Russell(2023)}]{phelps2023InvestigatingEmergentGoalLike}
\bibinfo{author}{Phelps, S.}, \bibinfo{author}{Russell, Y.I.}, \bibinfo{year}{2023}.
\newblock \bibinfo{title}{Investigating {{Emergent Goal-Like Behaviour}} in {{Large Language Models Using Experimental Economics}}}.
\newblock \DOIprefix\doi{10.48550/arXiv.2305.07970}, \href{http://arxiv.org/abs/2305.07970}{\tt arXiv:2305.07970}.
\bibitem[{Pirina and {\c{C}}{\"o}ltekin(2018)}]{pirina-coltekin-2018-identifying}
\bibinfo{author}{Pirina, I.}, \bibinfo{author}{{\c{C}}{\"o}ltekin, {\c{C}}.}, \bibinfo{year}{2018}.
\newblock \bibinfo{title}{Identifying depression on {R}eddit: The effect of training data}, in: \bibinfo{booktitle}{Proceedings of the 2018 {EMNLP} Workshop {SMM}4{H}: The 3rd Social Media Mining for Health Applications Workshop {\&} Shared Task}, \bibinfo{publisher}{Association for Computational Linguistics}, \bibinfo{address}{Brussels, Belgium}. pp. \bibinfo{pages}{9--12}.
\newblock \URLprefix \url{https://aclanthology.org/W18-5903}, \DOIprefix\doi{10.18653/v1/W18-5903}.
\bibitem[{Pollin(1965)}]{pollin1965philosophical}
\bibinfo{author}{Pollin, B.R.}, \bibinfo{year}{1965}.
\newblock \bibinfo{title}{Philosophical and literary sources of frankenstein}.
\newblock \bibinfo{journal}{Comparative Literature} \bibinfo{volume}{17}, \bibinfo{pages}{97--108}.
\bibitem[{Prystawski et~al.(2022)Prystawski, Thibodeau and Goodman}]{prystawski2022PsychologicallyinformedChainofthoughtPrompts}
\bibinfo{author}{Prystawski, B.}, \bibinfo{author}{Thibodeau, P.}, \bibinfo{author}{Goodman, N.}, \bibinfo{year}{2022}.
\newblock \bibinfo{title}{Psychologically-informed chain-of-thought prompts for metaphor understanding in large language models}.
\newblock \DOIprefix\doi{10.48550/arXiv.2209.08141}, \href{http://arxiv.org/abs/2209.08141}{\tt arXiv:2209.08141}.
\bibitem[{Raffel et~al.(2020)Raffel, Shazeer, Roberts, Lee, Narang, Matena, Zhou, Li and Liu}]{10.5555/3455716.3455856}
\bibinfo{author}{Raffel, C.}, \bibinfo{author}{Shazeer, N.}, \bibinfo{author}{Roberts, A.}, \bibinfo{author}{Lee, K.}, \bibinfo{author}{Narang, S.}, \bibinfo{author}{Matena, M.}, \bibinfo{author}{Zhou, Y.}, \bibinfo{author}{Li, W.}, \bibinfo{author}{Liu, P.J.}, \bibinfo{year}{2020}.
\newblock \bibinfo{title}{Exploring the limits of transfer learning with a unified text-to-text transformer}.
\newblock \bibinfo{journal}{The Journal of Machine Learning Research} \bibinfo{volume}{21}, \bibinfo{pages}{5485--5551}.
\bibitem[{Rao et~al.(2023)Rao, Leung and Miao}]{rao2023can}
\bibinfo{author}{Rao, H.}, \bibinfo{author}{Leung, C.}, \bibinfo{author}{Miao, C.}, \bibinfo{year}{2023}.
\newblock \bibinfo{title}{Can chatgpt assess human personalities? a general evaluation framework}.
\newblock \bibinfo{journal}{arXiv preprint arXiv:2303.01248} .
\bibitem[{Rathje et~al.(2023)Rathje, Mirea, Sucholutsky, Marjieh, Robertson and Van~Bavel}]{rathje2023gpt}
\bibinfo{author}{Rathje, S.}, \bibinfo{author}{Mirea, D.M.}, \bibinfo{author}{Sucholutsky, I.}, \bibinfo{author}{Marjieh, R.}, \bibinfo{author}{Robertson, C.}, \bibinfo{author}{Van~Bavel, J.J.}, \bibinfo{year}{2023}.
\newblock \bibinfo{title}{Gpt is an effective tool for multilingual psychological text analysis} .
\bibitem[{Rivas and Zhao(2023)}]{rivas2023MarketingChatGPTNavigating}
\bibinfo{author}{Rivas, P.}, \bibinfo{author}{Zhao, L.}, \bibinfo{year}{2023}.
\newblock \bibinfo{title}{Marketing with {{ChatGPT}}: {{Navigating}} the {{Ethical Terrain}} of {{GPT-Based Chatbot Technology}}}.
\newblock \bibinfo{journal}{AI} \bibinfo{volume}{4}, \bibinfo{pages}{375--384}.
\newblock \DOIprefix\doi{10.3390/ai4020019}.
\bibitem[{Rodríguez-Ibánez et~al.(2023)Rodríguez-Ibánez, Casánez-Ventura, Castejón-Mateos and Cuenca-Jiménez}]{RODRIGUEZIBANEZ2023119862}
\bibinfo{author}{Rodríguez-Ibánez, M.}, \bibinfo{author}{Casánez-Ventura, A.}, \bibinfo{author}{Castejón-Mateos, F.}, \bibinfo{author}{Cuenca-Jiménez, P.M.}, \bibinfo{year}{2023}.
\newblock \bibinfo{title}{A review on sentiment analysis from social media platforms}.
\newblock \bibinfo{journal}{Expert Systems with Applications} \bibinfo{volume}{223}, \bibinfo{pages}{119862}.
\newblock \URLprefix \url{https://www.sciencedirect.com/science/article/pii/S0957417423003639}, \DOIprefix\doi{https://doi.org/10.1016/j.eswa.2023.119862}.
\bibitem[{Rubin and Babbie(2016)}]{rubin2016empowerment}
\bibinfo{author}{Rubin, A.}, \bibinfo{author}{Babbie, E.R.}, \bibinfo{year}{2016}.
\newblock \bibinfo{title}{Empowerment series: Research methods for social work}.
\newblock \bibinfo{publisher}{Cengage Learning}.
\bibitem[{Russell(2010)}]{russell2010artificial}
\bibinfo{author}{Russell, S.J.}, \bibinfo{year}{2010}.
\newblock \bibinfo{title}{Artificial intelligence a modern approach}.
\newblock \bibinfo{publisher}{Pearson Education, Inc.}
\bibitem[{Rutinowski et~al.(2023)Rutinowski, Franke, Endendyk, Dormuth and Pauly}]{rutinowski2023SelfPerceptionPoliticalBiases}
\bibinfo{author}{Rutinowski, J.}, \bibinfo{author}{Franke, S.}, \bibinfo{author}{Endendyk, J.}, \bibinfo{author}{Dormuth, I.}, \bibinfo{author}{Pauly, M.}, \bibinfo{year}{2023}.
\newblock \bibinfo{title}{The {{Self-Perception}} and {{Political Biases}} of {{ChatGPT}}}.
\newblock \DOIprefix\doi{10.48550/arXiv.2304.07333}, \href{http://arxiv.org/abs/2304.07333}{\tt arXiv:2304.07333}.
\bibitem[{Salehi et~al.(2022)Salehi, Hassan, Lammerse, Sabet, Riiser, R{\o}ed, Johnson, Thambawita, Hicks, Powell et~al.}]{salehi2022synthesizing}
\bibinfo{author}{Salehi, P.}, \bibinfo{author}{Hassan, S.Z.}, \bibinfo{author}{Lammerse, M.}, \bibinfo{author}{Sabet, S.S.}, \bibinfo{author}{Riiser, I.}, \bibinfo{author}{R{\o}ed, R.K.}, \bibinfo{author}{Johnson, M.S.}, \bibinfo{author}{Thambawita, V.}, \bibinfo{author}{Hicks, S.A.}, \bibinfo{author}{Powell, M.}, et~al., \bibinfo{year}{2022}.
\newblock \bibinfo{title}{Synthesizing a talking child avatar to train interviewers working with maltreated children}.
\newblock \bibinfo{journal}{Big Data and Cognitive Computing} \bibinfo{volume}{6}, \bibinfo{pages}{62}.
\bibitem[{Santurkar et~al.(2023)Santurkar, Durmus, Ladhak, Lee, Liang and Hashimoto}]{santurkar2023WhoseOpinionsLanguage}
\bibinfo{author}{Santurkar, S.}, \bibinfo{author}{Durmus, E.}, \bibinfo{author}{Ladhak, F.}, \bibinfo{author}{Lee, C.}, \bibinfo{author}{Liang, P.}, \bibinfo{author}{Hashimoto, T.}, \bibinfo{year}{2023}.
\newblock \bibinfo{title}{Whose {{Opinions Do Language Models Reflect}}?}
\newblock \DOIprefix\doi{10.48550/arXiv.2303.17548}, \href{http://arxiv.org/abs/2303.17548}{\tt arXiv:2303.17548}.
\bibitem[{Sap et~al.(2022)Sap, Le~Bras, Fried and Choi}]{sap2022NeuralTheoryofMindLimits}
\bibinfo{author}{Sap, M.}, \bibinfo{author}{Le~Bras, R.}, \bibinfo{author}{Fried, D.}, \bibinfo{author}{Choi, Y.}, \bibinfo{year}{2022}.
\newblock \bibinfo{title}{Neural {{Theory-of-Mind}}? {{On}} the {{Limits}} of {{Social Intelligence}} in {{Large LMs}}}, in: \bibinfo{booktitle}{Proceedings of the 2022 {{Conference}} on {{Empirical Methods}} in {{Natural Language Processing}}}, \bibinfo{publisher}{{Association for Computational Linguistics}}, \bibinfo{address}{{Abu Dhabi, United Arab Emirates}}. pp. \bibinfo{pages}{3762--3780}.
\newblock \URLprefix \url{https://aclanthology.org/2022.emnlp-main.248}.
\bibitem[{Scao et~al.(2023)Scao, Fan, Akiki, Pavlick, Ili{\'c}, Hesslow, Castagn{\'e}, Luccioni, Yvon, Gall{\'e} et~al.}]{workshop2023BLOOM176BParameterOpenAccess}
\bibinfo{author}{Scao, T.L.}, \bibinfo{author}{Fan, A.}, \bibinfo{author}{Akiki, C.}, \bibinfo{author}{Pavlick, E.}, \bibinfo{author}{Ili{\'c}, S.}, \bibinfo{author}{Hesslow, D.}, \bibinfo{author}{Castagn{\'e}, R.}, \bibinfo{author}{Luccioni, A.S.}, \bibinfo{author}{Yvon, F.}, \bibinfo{author}{Gall{\'e}, M.}, et~al., \bibinfo{year}{2023}.
\newblock \bibinfo{title}{{{BLOOM}}: {{A 176B-Parameter Open-Access Multilingual Language Model}}}.
\newblock \DOIprefix\doi{10.48550/arXiv.2211.05100}, \href{http://arxiv.org/abs/2211.05100}{\tt arXiv:2211.05100}.
\bibitem[{Shanahan(2022)}]{shanahan2022talking}
\bibinfo{author}{Shanahan, M.}, \bibinfo{year}{2022}.
\newblock \bibinfo{title}{Talking about large language models}.
\newblock \bibinfo{journal}{arXiv preprint arXiv:2212.03551} .
\bibitem[{Sinha and Khandait(2021)}]{sinha_impact_2021}
\bibinfo{author}{Sinha, A.}, \bibinfo{author}{Khandait, T.}, \bibinfo{year}{2021}.
\newblock \bibinfo{title}{Impact of {News} on the {Commodity} {Market}: {Dataset} and {Results}}, in: \bibinfo{editor}{Arai, K.} (Ed.), \bibinfo{booktitle}{Advances in {Information} and {Communication}}, \bibinfo{publisher}{Springer International Publishing}, \bibinfo{address}{Cham}. pp. \bibinfo{pages}{589--601}.
\bibitem[{Son et~al.(2023)Son, Jung, Hahm, Na and Jin}]{son2023ClassificationFinancialReasoning}
\bibinfo{author}{Son, G.}, \bibinfo{author}{Jung, H.}, \bibinfo{author}{Hahm, M.}, \bibinfo{author}{Na, K.}, \bibinfo{author}{Jin, S.}, \bibinfo{year}{2023}.
\newblock \bibinfo{title}{Beyond {{Classification}}: {{Financial Reasoning}} in {{State-of-the-Art Language Models}}}.
\newblock \DOIprefix\doi{10.48550/arXiv.2305.01505}, \href{http://arxiv.org/abs/2305.01505}{\tt arXiv:2305.01505}.
\bibitem[{Soun et~al.(2022)Soun, Yoo, Cho et~al.}]{10020720}
\bibinfo{author}{Soun, Y.}, \bibinfo{author}{Yoo, J.}, \bibinfo{author}{Cho, M.}, et~al., \bibinfo{year}{2022}.
\newblock \bibinfo{title}{Accurate stock movement prediction with self-supervised learning from sparse noisy tweets}, in: \bibinfo{booktitle}{2022 IEEE International Conference on Big Data (Big Data)}, pp. \bibinfo{pages}{1691--1700}.
\newblock \DOIprefix\doi{10.1109/BigData55660.2022.10020720}.
\bibitem[{Stevenson et~al.(2022)Stevenson, Smal, Baas, Grasman and {van der Maas}}]{stevenson2022PuttingGPT3Creativity}
\bibinfo{author}{Stevenson, C.}, \bibinfo{author}{Smal, I.}, \bibinfo{author}{Baas, M.}, \bibinfo{author}{Grasman, R.}, \bibinfo{author}{{van der Maas}, H.}, \bibinfo{year}{2022}.
\newblock \bibinfo{title}{Putting {{GPT-3}}'s {{Creativity}} to the ({{Alternative Uses}}) {{Test}}}.
\newblock \DOIprefix\doi{10.48550/arXiv.2206.08932}, \href{http://arxiv.org/abs/2206.08932}{\tt arXiv:2206.08932}.
\bibitem[{Tang et~al.(2023)Tang, Peng, Wang, Ding, Durrett and Rousseau}]{tang2023less}
\bibinfo{author}{Tang, L.}, \bibinfo{author}{Peng, Y.}, \bibinfo{author}{Wang, Y.}, \bibinfo{author}{Ding, Y.}, \bibinfo{author}{Durrett, G.}, \bibinfo{author}{Rousseau, J.F.}, \bibinfo{year}{2023}.
\newblock \bibinfo{title}{Less likely brainstorming: Using language models to generate alternative hypotheses}, in: \bibinfo{booktitle}{Proceedings of the 61th Annual Meeting of the Association for Computational Linguistics (Volume 1: Long Papers)}, \bibinfo{publisher}{Association for Computational Linguistics}.
\bibitem[{Terwiesch(2023)}]{terwiesch2023WouldChatGPT3}
\bibinfo{author}{Terwiesch, C.}, \bibinfo{year}{2023}.
\newblock \bibinfo{title}{Would chat gpt3 get a wharton mba}.
\newblock \bibinfo{journal}{A prediction based on its performance in the operations management course. Wharton: Mack Institute for Innovation Management/University of Pennsylvania/School Wharton} .
\bibitem[{Tiku(2023)}]{tiku2023GoogleEngineerWho}
\bibinfo{author}{Tiku, N.}, \bibinfo{year}{2023}.
\newblock \bibinfo{title}{The {{Google}} engineer who thinks the company's {{AI}} has come to life}.
\newblock \bibinfo{journal}{Washington Post} \URLprefix \url{https://www.washingtonpost.com/technology/2022/06/11/google-ai-lamda-blake-lemoine/}.
\bibitem[{Touvron et~al.(2023)Touvron, Lavril, Izacard, Martinet, Lachaux, Lacroix, Rozi{\`e}re, Goyal, Hambro, Azhar, Rodriguez, Joulin, Grave and Lample}]{touvron2023LLaMAOpenEfficient}
\bibinfo{author}{Touvron, H.}, \bibinfo{author}{Lavril, T.}, \bibinfo{author}{Izacard, G.}, \bibinfo{author}{Martinet, X.}, \bibinfo{author}{Lachaux, M.A.}, \bibinfo{author}{Lacroix, T.}, \bibinfo{author}{Rozi{\`e}re, B.}, \bibinfo{author}{Goyal, N.}, \bibinfo{author}{Hambro, E.}, \bibinfo{author}{Azhar, F.}, \bibinfo{author}{Rodriguez, A.}, \bibinfo{author}{Joulin, A.}, \bibinfo{author}{Grave, E.}, \bibinfo{author}{Lample, G.}, \bibinfo{year}{2023}.
\newblock \bibinfo{title}{{{LLaMA}}: {{Open}} and {{Efficient Foundation Language Models}}}.
\newblock \DOIprefix\doi{10.48550/arXiv.2302.13971}, \href{http://arxiv.org/abs/2302.13971}{\tt arXiv:2302.13971}.
\bibitem[{Trochim and Donnelly(2001)}]{trochim2001research}
\bibinfo{author}{Trochim, W.M.}, \bibinfo{author}{Donnelly, J.P.}, \bibinfo{year}{2001}.
\newblock \bibinfo{title}{Research methods knowledge base}. volume~\bibinfo{volume}{2}.
\newblock \bibinfo{publisher}{Atomic Dog Pub. Macmillan Publishing Company, New York}.
\bibitem[{Trott et~al.(2023)Trott, Jones, Chang, Michaelov and Bergen}]{trott2022LargeLanguageModels}
\bibinfo{author}{Trott, S.}, \bibinfo{author}{Jones, C.}, \bibinfo{author}{Chang, T.}, \bibinfo{author}{Michaelov, J.}, \bibinfo{author}{Bergen, B.}, \bibinfo{year}{2023}.
\newblock \bibinfo{title}{Do large language models know what humans know?}
\newblock \bibinfo{journal}{Cognitive Science} \bibinfo{volume}{47}, \bibinfo{pages}{e13309}.
\bibitem[{Tu et~al.(2023)Tu, Li, Yu, Wang, Hou and Li}]{tu2023ChatLogRecordingAnalyzing}
\bibinfo{author}{Tu, S.}, \bibinfo{author}{Li, C.}, \bibinfo{author}{Yu, J.}, \bibinfo{author}{Wang, X.}, \bibinfo{author}{Hou, L.}, \bibinfo{author}{Li, J.}, \bibinfo{year}{2023}.
\newblock \bibinfo{title}{{{ChatLog}}: {{Recording}} and {{Analyzing ChatGPT Across Time}}}.
\newblock \DOIprefix\doi{10.48550/arXiv.2304.14106}, \href{http://arxiv.org/abs/2304.14106}{\tt arXiv:2304.14106}.
\bibitem[{Turcan and McKeown(2019)}]{Turcan2019DreadditAR}
\bibinfo{author}{Turcan, E.}, \bibinfo{author}{McKeown, K.}, \bibinfo{year}{2019}.
\newblock \bibinfo{title}{Dreaddit: A reddit dataset for stress analysis in social media}, in: \bibinfo{booktitle}{Conference on Empirical Methods in Natural Language Processing}.
\newblock \URLprefix \url{https://api.semanticscholar.org/CorpusID:207870937}.
\bibitem[{Törnberg(2023)}]{tornberg2023chatgpt4}
\bibinfo{author}{Törnberg, P.}, \bibinfo{year}{2023}.
\newblock \bibinfo{title}{Chatgpt-4 outperforms experts and crowd workers in annotating political twitter messages with zero-shot learning}.
\newblock \href{http://arxiv.org/abs/2304.06588}{\tt arXiv:2304.06588}.
\bibitem[{Ullman(2023)}]{ullman2023LargeLanguageModels}
\bibinfo{author}{Ullman, T.}, \bibinfo{year}{2023}.
\newblock \bibinfo{title}{Large {{Language Models Fail}} on {{Trivial Alterations}} to {{Theory-of-Mind Tasks}}}.
\newblock \DOIprefix\doi{10.48550/arXiv.2302.08399}, \href{http://arxiv.org/abs/2302.08399}{\tt arXiv:2302.08399}.
\bibitem[{Uludag(2023)}]{uludag2023chatgpt}
\bibinfo{author}{Uludag, K.}, \bibinfo{year}{2023}.
\newblock \bibinfo{title}{Chatgpt can distinguish paranoid thoughts in patients with schizophrenia}.
\newblock \bibinfo{journal}{Available at SSRN 4391941} .
\bibitem[{Wang et~al.(2023a)Wang, Ma, Feng, Zhang, Yang, Zhang, Chen, Tang, Chen, Lin et~al.}]{wang2023survey}
\bibinfo{author}{Wang, L.}, \bibinfo{author}{Ma, C.}, \bibinfo{author}{Feng, X.}, \bibinfo{author}{Zhang, Z.}, \bibinfo{author}{Yang, H.}, \bibinfo{author}{Zhang, J.}, \bibinfo{author}{Chen, Z.}, \bibinfo{author}{Tang, J.}, \bibinfo{author}{Chen, X.}, \bibinfo{author}{Lin, Y.}, et~al., \bibinfo{year}{2023}a.
\newblock \bibinfo{title}{A survey on large language model based autonomous agents}.
\newblock \bibinfo{journal}{arXiv preprint arXiv:2308.11432} .
\bibitem[{Wang et~al.(2023b)Wang, Scells, Koopman and Zuccon}]{wang2023can}
\bibinfo{author}{Wang, S.}, \bibinfo{author}{Scells, H.}, \bibinfo{author}{Koopman, B.}, \bibinfo{author}{Zuccon, G.}, \bibinfo{year}{2023}b.
\newblock \bibinfo{title}{Can chatgpt write a good boolean query for systematic review literature search?}
\newblock \bibinfo{journal}{arXiv preprint arXiv:2302.03495} .
\bibitem[{Webb et~al.(2023)Webb, Holyoak and Lu}]{webb2023EmergentAnalogicalReasoning}
\bibinfo{author}{Webb, T.}, \bibinfo{author}{Holyoak, K.J.}, \bibinfo{author}{Lu, H.}, \bibinfo{year}{2023}.
\newblock \bibinfo{title}{Emergent analogical reasoning in large language models}.
\newblock \bibinfo{journal}{Nature Human Behaviour} , \bibinfo{pages}{1--16}.
\bibitem[{Wei et~al.(2022)Wei, Wang, Schuurmans, Bosma, Xia, Chi, Le, Zhou et~al.}]{wei2022chain}
\bibinfo{author}{Wei, J.}, \bibinfo{author}{Wang, X.}, \bibinfo{author}{Schuurmans, D.}, \bibinfo{author}{Bosma, M.}, \bibinfo{author}{Xia, F.}, \bibinfo{author}{Chi, E.}, \bibinfo{author}{Le, Q.V.}, \bibinfo{author}{Zhou, D.}, et~al., \bibinfo{year}{2022}.
\newblock \bibinfo{title}{Chain-of-thought prompting elicits reasoning in large language models}.
\newblock \bibinfo{journal}{Advances in Neural Information Processing Systems} \bibinfo{volume}{35}, \bibinfo{pages}{24824--24837}.
\bibitem[{Willer and Walker(2007)}]{willer2007building}
\bibinfo{author}{Willer, D.}, \bibinfo{author}{Walker, H.A.}, \bibinfo{year}{2007}.
\newblock \bibinfo{title}{Building experiments: Testing social theory}.
\newblock \bibinfo{publisher}{Stanford University Press}.
\bibitem[{Wilson et~al.(2018)Wilson, Wilkins, Holt, Choi, Konecki, Lin, Koire, Chen, Kim, Wang et~al.}]{wilson2018automated}
\bibinfo{author}{Wilson, S.J.}, \bibinfo{author}{Wilkins, A.D.}, \bibinfo{author}{Holt, M.V.}, \bibinfo{author}{Choi, B.K.}, \bibinfo{author}{Konecki, D.}, \bibinfo{author}{Lin, C.H.}, \bibinfo{author}{Koire, A.}, \bibinfo{author}{Chen, Y.}, \bibinfo{author}{Kim, S.Y.}, \bibinfo{author}{Wang, Y.}, et~al., \bibinfo{year}{2018}.
\newblock \bibinfo{title}{Automated literature mining and hypothesis generation through a network of medical subject headings}.
\newblock \bibinfo{journal}{bioRxiv} , \bibinfo{pages}{403667}.
\bibitem[{Wood et~al.(2023)Wood, Achhpilia, Adams, Aghazadeh, Akinyele, Akpan, Allee, Allen, Almer, Ames et~al.}]{wood2023ChatGPTArtificialIntelligence}
\bibinfo{author}{Wood, D.A.}, \bibinfo{author}{Achhpilia, M.P.}, \bibinfo{author}{Adams, M.T.}, \bibinfo{author}{Aghazadeh, S.}, \bibinfo{author}{Akinyele, K.}, \bibinfo{author}{Akpan, M.}, \bibinfo{author}{Allee, K.D.}, \bibinfo{author}{Allen, A.M.}, \bibinfo{author}{Almer, E.D.}, \bibinfo{author}{Ames, D.}, et~al., \bibinfo{year}{2023}.
\newblock \bibinfo{title}{The {{ChatGPT Artificial Intelligence Chatbot}}: {{How Well Does It Answer Accounting Assessment Questions}}?}
\newblock \bibinfo{journal}{Issues in Accounting Education} \bibinfo{volume}{38}.
\newblock \URLprefix \url{https://www.researchgate.net/publication/370211135_The_ChatGPT_Artificial_Intelligence_Chatbot_How_Well_Does_It_Answer_Accounting_Assessment_Questions}.
\bibitem[{Woolgar(1985)}]{woolgar1985not}
\bibinfo{author}{Woolgar, S.}, \bibinfo{year}{1985}.
\newblock \bibinfo{title}{Why not a sociology of machines? the case of sociology and artificial intelligence}.
\newblock \bibinfo{journal}{Sociology} \bibinfo{volume}{19}, \bibinfo{pages}{557--572}.
\bibitem[{Wright et~al.(2010)Wright, Marsden et~al.}]{wright2010survey}
\bibinfo{author}{Wright, J.D.}, \bibinfo{author}{Marsden, P.V.}, et~al., \bibinfo{year}{2010}.
\newblock \bibinfo{title}{Survey research and social science: History, current practice, and future prospects}.
\newblock \bibinfo{journal}{Handbook of survey research} , \bibinfo{pages}{3--26}.
\bibitem[{Wu et~al.(2018)Wu, Zhang, Shen et~al.}]{Wu2018HybridDS}
\bibinfo{author}{Wu, H.}, \bibinfo{author}{Zhang, W.}, \bibinfo{author}{Shen, W.}, et~al., \bibinfo{year}{2018}.
\newblock \bibinfo{title}{Hybrid deep sequential modeling for social text-driven stock prediction}.
\newblock \bibinfo{journal}{Proceedings of the 27th ACM International Conference on Information and Knowledge Management} \URLprefix \url{https://api.semanticscholar.org/CorpusID:53038910}.
\bibitem[{Wu et~al.(2023a)Wu, Nagler, Tucker and Messing}]{wu2023large}
\bibinfo{author}{Wu, P.Y.}, \bibinfo{author}{Nagler, J.}, \bibinfo{author}{Tucker, J.A.}, \bibinfo{author}{Messing, S.}, \bibinfo{year}{2023}a.
\newblock \bibinfo{title}{Large language models can be used to scale the ideologies of politicians in a zero-shot learning setting}.
\newblock \href{http://arxiv.org/abs/2303.12057}{\tt arXiv:2303.12057}.
\bibitem[{Wu et~al.(2023b)Wu, Irsoy, Lu, Dabravolski, Dredze, Gehrmann, Kambadur, Rosenberg and Mann}]{wuBloombergGPTLargeLanguage2023}
\bibinfo{author}{Wu, S.}, \bibinfo{author}{Irsoy, O.}, \bibinfo{author}{Lu, S.}, \bibinfo{author}{Dabravolski, V.}, \bibinfo{author}{Dredze, M.}, \bibinfo{author}{Gehrmann, S.}, \bibinfo{author}{Kambadur, P.}, \bibinfo{author}{Rosenberg, D.}, \bibinfo{author}{Mann, G.}, \bibinfo{year}{2023}b.
\newblock \bibinfo{title}{{{BloombergGPT}}: {{A Large Language Model}} for {{Finance}}}.
\newblock \URLprefix \url{http://arxiv.org/abs/2303.17564}, \DOIprefix\doi{10.48550/arXiv.2303.17564}, \href{http://arxiv.org/abs/2303.17564}{\tt arXiv:2303.17564}.
\bibitem[{Xie et~al.(2023a)Xie, Han, Lai, Peng and Huang}]{xie2023wall}
\bibinfo{author}{Xie, Q.}, \bibinfo{author}{Han, W.}, \bibinfo{author}{Lai, Y.}, \bibinfo{author}{Peng, M.}, \bibinfo{author}{Huang, J.}, \bibinfo{year}{2023}a.
\newblock \bibinfo{title}{The wall street neophyte: A zero-shot analysis of chatgpt over multimodal stock movement prediction challenges}.
\newblock \href{http://arxiv.org/abs/2304.05351}{\tt arXiv:2304.05351}.
\bibitem[{Xie et~al.(2023b)Xie, Han, Zhang et~al.}]{Xie2023PIXIUAL}
\bibinfo{author}{Xie, Q.}, \bibinfo{author}{Han, W.}, \bibinfo{author}{Zhang, X.}, et~al., \bibinfo{year}{2023}b.
\newblock \bibinfo{title}{Pixiu: A large language model, instruction data and evaluation benchmark for finance}.
\newblock \bibinfo{journal}{ArXiv} \bibinfo{volume}{abs/2306.05443}.
\newblock \URLprefix \url{https://api.semanticscholar.org/CorpusID:259129602}.
\bibitem[{Xu and Cohen(2018)}]{xu-cohen-2018-stock}
\bibinfo{author}{Xu, Y.}, \bibinfo{author}{Cohen, S.B.}, \bibinfo{year}{2018}.
\newblock \bibinfo{title}{Stock movement prediction from tweets and historical prices}, in: \bibinfo{booktitle}{Proceedings of the 56th Annual Meeting of the Association for Computational Linguistics (Volume 1: Long Papers)}, \bibinfo{publisher}{Association for Computational Linguistics}, \bibinfo{address}{Melbourne, Australia}. pp. \bibinfo{pages}{1970--1979}.
\newblock \URLprefix \url{https://aclanthology.org/P18-1183}, \DOIprefix\doi{10.18653/v1/P18-1183}.
\bibitem[{Yang et~al.(2023a)Yang, Ji, Zhang, Xie and Ananiadou}]{yang2023evaluations}
\bibinfo{author}{Yang, K.}, \bibinfo{author}{Ji, S.}, \bibinfo{author}{Zhang, T.}, \bibinfo{author}{Xie, Q.}, \bibinfo{author}{Ananiadou, S.}, \bibinfo{year}{2023}a.
\newblock \bibinfo{title}{On the evaluations of chatgpt and emotion-enhanced prompting for mental health analysis}.
\newblock \bibinfo{journal}{arXiv preprint arXiv:2304.03347} .
\bibitem[{Yang et~al.(2023b)Yang, Ji, Zhang et~al.}]{yang2023interpretable}
\bibinfo{author}{Yang, K.}, \bibinfo{author}{Ji, S.}, \bibinfo{author}{Zhang, T.}, et~al., \bibinfo{year}{2023}b.
\newblock \bibinfo{title}{Towards interpretable mental health analysis with large language models}.
\newblock \href{http://arxiv.org/abs/2304.03347}{\tt arXiv:2304.03347}.
\bibitem[{Yuan(2013)}]{fang2013}
\bibinfo{author}{Yuan, F.}, \bibinfo{year}{2013}.
\newblock \bibinfo{title}{Tutorial on Social Research methods}.
\newblock \bibinfo{publisher}{Peking University Press}.
\bibitem[{Zhang et~al.(2023)Zhang, Ding and Jing}]{zhang2023stance}
\bibinfo{author}{Zhang, B.}, \bibinfo{author}{Ding, D.}, \bibinfo{author}{Jing, L.}, \bibinfo{year}{2023}.
\newblock \bibinfo{title}{How would stance detection techniques evolve after the launch of chatgpt?}
\newblock \href{http://arxiv.org/abs/2212.14548}{\tt arXiv:2212.14548}.
\bibitem[{Zhang et~al.(2022)Zhang, Roller, Goyal, Artetxe, Chen, Chen, Dewan, Diab, Li, Lin, Mihaylov, Ott, Shleifer, Shuster, Simig, Koura, Sridhar, Wang and Zettlemoyer}]{zhang2022OPTOpenPretrained}
\bibinfo{author}{Zhang, S.}, \bibinfo{author}{Roller, S.}, \bibinfo{author}{Goyal, N.}, \bibinfo{author}{Artetxe, M.}, \bibinfo{author}{Chen, M.}, \bibinfo{author}{Chen, S.}, \bibinfo{author}{Dewan, C.}, \bibinfo{author}{Diab, M.}, \bibinfo{author}{Li, X.}, \bibinfo{author}{Lin, X.V.}, \bibinfo{author}{Mihaylov, T.}, \bibinfo{author}{Ott, M.}, \bibinfo{author}{Shleifer, S.}, \bibinfo{author}{Shuster, K.}, \bibinfo{author}{Simig, D.}, \bibinfo{author}{Koura, P.S.}, \bibinfo{author}{Sridhar, A.}, \bibinfo{author}{Wang, T.}, \bibinfo{author}{Zettlemoyer, L.}, \bibinfo{year}{2022}.
\newblock \bibinfo{title}{{{OPT}}: {{Open Pre-trained Transformer Language Models}}}.
\newblock \DOIprefix\doi{10.48550/arXiv.2205.01068}, \href{http://arxiv.org/abs/2205.01068}{\tt arXiv:2205.01068}.
\bibitem[{Zhao et~al.(2023)Zhao, Zhou, Li, Tang, Wang, Hou, Min, Zhang, Zhang, Dong et~al.}]{zhao2023survey}
\bibinfo{author}{Zhao, W.X.}, \bibinfo{author}{Zhou, K.}, \bibinfo{author}{Li, J.}, \bibinfo{author}{Tang, T.}, \bibinfo{author}{Wang, X.}, \bibinfo{author}{Hou, Y.}, \bibinfo{author}{Min, Y.}, \bibinfo{author}{Zhang, B.}, \bibinfo{author}{Zhang, J.}, \bibinfo{author}{Dong, Z.}, et~al., \bibinfo{year}{2023}.
\newblock \bibinfo{title}{A survey of large language models}.
\newblock \bibinfo{journal}{arXiv preprint arXiv:2303.18223} .
\bibitem[{Zhao et~al.(2021)Zhao, Wallace, Feng, Klein and Singh}]{zhao2021CalibrateUseImproving}
\bibinfo{author}{Zhao, Z.}, \bibinfo{author}{Wallace, E.}, \bibinfo{author}{Feng, S.}, \bibinfo{author}{Klein, D.}, \bibinfo{author}{Singh, S.}, \bibinfo{year}{2021}.
\newblock \bibinfo{title}{Calibrate {{Before Use}}: {{Improving Few-shot Performance}} of {{Language Models}}}, in: \bibinfo{booktitle}{Proceedings of the 38th {{International Conference}} on {{Machine Learning}}}, \bibinfo{publisher}{{PMLR}}. pp. \bibinfo{pages}{12697--12706}.
\newblock \URLprefix \url{https://proceedings.mlr.press/v139/zhao21c.html}.
\bibitem[{Zhou et~al.(2023)Zhou, Jiang, Li, Wu, Wang, Qiu, Zhang, Chen, Wu, Wang et~al.}]{zhou2023agents}
\bibinfo{author}{Zhou, W.}, \bibinfo{author}{Jiang, Y.E.}, \bibinfo{author}{Li, L.}, \bibinfo{author}{Wu, J.}, \bibinfo{author}{Wang, T.}, \bibinfo{author}{Qiu, S.}, \bibinfo{author}{Zhang, J.}, \bibinfo{author}{Chen, J.}, \bibinfo{author}{Wu, R.}, \bibinfo{author}{Wang, S.}, et~al., \bibinfo{year}{2023}.
\newblock \bibinfo{title}{Agents: An open-source framework for autonomous language agents}.
\newblock \bibinfo{journal}{arXiv preprint arXiv:2309.07870} .
\bibitem[{Ziems et~al.(2023)Ziems, Held, Shaikh, Chen, Zhang and Yang}]{ziems2023can}
\bibinfo{author}{Ziems, C.}, \bibinfo{author}{Held, W.}, \bibinfo{author}{Shaikh, O.}, \bibinfo{author}{Chen, J.}, \bibinfo{author}{Zhang, Z.}, \bibinfo{author}{Yang, D.}, \bibinfo{year}{2023}.
\newblock \bibinfo{title}{Can large language models transform computational social science?}
\newblock \bibinfo{journal}{arXiv preprint arXiv:2305.03514} .
\bibitem[{Zimbardo et~al.(1971)Zimbardo, Haney, Banks and Jaffe}]{zimbardo1971stanford}
\bibinfo{author}{Zimbardo, P.G.}, \bibinfo{author}{Haney, C.}, \bibinfo{author}{Banks, W.C.}, \bibinfo{author}{Jaffe, D.}, \bibinfo{year}{1971}.
\newblock \bibinfo{title}{The Stanford prison experiment}.
\newblock \bibinfo{publisher}{Zimbardo, Incorporated}.

\end{thebibliography}

\end{document}